\documentclass{article}

\usepackage[preprint]{neurips_2025}

\usepackage{enumitem}
\usepackage[hidelinks]{hyperref}
\usepackage{url}
\usepackage{booktabs}
\usepackage{amsfonts}
\usepackage{graphicx}
\usepackage{fontawesome}
\usepackage{sidecap}
\usepackage[small]{caption}
\usepackage{subcaption}
\usepackage{amsmath}
\allowdisplaybreaks
\usepackage{amsthm}

\newtheorem{definition}{Definition}
\usepackage{multicol}

\usepackage{float}
\usepackage{xcolor}
\usepackage{colortbl}
\usepackage{amssymb}
\usepackage{adjustbox}
\usepackage{pifont}
\usepackage[noabbrev,capitalize]{cleveref}
\usepackage{wrapfig}

\usepackage[compact]{titlesec}
\titlespacing{\section}{0pt}{1ex}{0.5ex}
\titlespacing{\subsection}{0pt}{0.5ex}{0ex}
\titlespacing{\subsubsection}{0pt}{0.5ex}{0ex} 
\usepackage{setspace}
\AtBeginDocument{%
  \addtolength\abovedisplayskip{-0.25\baselineskip}%
  \addtolength\belowdisplayskip{-0.25\baselineskip}%
  \addtolength\abovedisplayshortskip{-0.25\baselineskip}%
  \addtolength\belowdisplayshortskip{-0.25\baselineskip}%
}

\usepackage{xspace}
\newcommand{\circone}{\ding{172}\xspace}
\newcommand{\circtwo}{\ding{173}\xspace}
\newcommand{\circthree}{\ding{174}\xspace}
\newcommand{\circfour}{\ding{175}\xspace}

\newcommand{\ie}{\textit{i}.\textit{e}.,\xspace}
\newcommand{\eg}{\textit{e}.\textit{g}.,\xspace}

\usepackage{tabularx}
\newcolumntype{C}{>{\centering\arraybackslash}X}
\newcolumntype{R}{>{\raggedleft\arraybackslash}X}
\newcolumntype{S}{>{\raggedleft\arraybackslash\hsize=.5\hsize}X}

\usepackage{arydshln}
\usepackage{multirow}

\usepackage{makecell}

\newcommand{\optparens}[1]{\if\relax\detokenize{#1}\relax\else(#1)\fi}

\crefname{equation}{equation}{equations}
\crefname{section}{section}{sections}
\crefname{footnote}{footnote}{footnotes}   
\crefname{line}{line}{lines}   
\crefname{assumption}{assumption}{assumptions}
\crefname{lstlisting}{listing}{listings}
\Crefname{lstlisting}{Listing}{Listings}
\crefname{appendix}{Appendix}{Appendices}
\Crefname{appendix}{Appendix}{Appendices}
\usepackage{bbm}
\usepackage{appendix}

\usepackage{soul}
\definecolor{aigold}{RGB}{244,210, 1} 
\definecolor{aigreen}{RGB}{245, 255, 249}

\definecolor{humanpurple}{RGB}{235, 222, 240}

\definecolor{commentgray}{RGB}{86, 101, 115}

\definecolor{light-blue}{rgb}{0.6,0.6,1}

\definecolor{aired}{RGB}{255,180,181}

\usepackage{listings}
\usepackage{parcolumns}
\lstdefinestyle{datalogstyle}{
	basicstyle={\codefont\small},  
	xleftmargin={6pt},
        xrightmargin={6pt},
        breakindent=0pt,
	frame=tb,
	stepnumber=1,
	firstnumber=1,
	numberfirstline=true,
	tabsize=2,
	showtabs=false,
	showspaces=false,
	showstringspaces=false,
	extendedchars=true,
	breaklines=true,
	columns=fullflexible,
	keepspaces=true,
	escapeinside={@}{@},
	firstnumber=last,
	captionpos=b,
	commentstyle=\color{black!65},
	numberstyle=\tiny\color{black!65},
	stringstyle=\color{codepurple},
	breakatwhitespace=false, 
	keepspaces=true,                 
	numbersep=5pt,                  
	showspaces=false,                
	showstringspaces=false,
	showtabs=false,
	aboveskip={0.8\baselineskip},
	belowskip={0.2\baselineskip},
	backgroundcolor=\color{aigreen},
}
\definecolor{rebuttal}{RGB}{229,255,204}
\sethlcolor{rebuttal}

\newcommand{\codefont}{\fontfamily{lmtt}\selectfont}

\usepackage[textwidth=2.7cm]{todonotes}

\newcommand{\cutforspace}[1]{}

\newcommand\myshade{85}
\colorlet{myurlcolor}{blue}
\hypersetup{
  citecolor  = black,
  urlcolor   = myurlcolor!\myshade!black,
  colorlinks = true,
}

\title{\textsc{QuitoBench}: A High-\underline{Qu}al\underline{it}y \underline{O}pen Time Series Forecasting \underline{Bench}mark}

\vspace{-2cm}
\author{
    Siqiao Xue$^{*\ddagger}$, Zhaoyang Zhu$^*$, Wei Zhang, 
    Rongyao Cai$^\ddagger$, Rui Wang,\\
    \textbf{Yixiang Mu$^{\ddagger}$, Fan Zhou$^{\ddagger}$, Jianguo Li, Peng Di, Hang Yu$^\dagger$} \\[0.3em]
    Ant Group \\[0.6em]
    \textbf{
        \href{https://hq-bench.github.io/quito/}
             {\textcolor{blue!60!black}{\faGlobe\enspace{Website}}}
        \quad
        \href{https://github.com/alipay/quito}
             {\textcolor{blue!60!black}{\faGithub\enspace{Code}}}
        \quad
        \href{https://huggingface.co/datasets/hq-bench/quitobench}
             {\textcolor{blue!60!black}{\faDatabase\enspace Data}}
    }
}

\begin{document}

\maketitle
{
  \renewcommand{\thefootnote}{\fnsymbol{footnote}}
  \footnotetext[1]{$^*$Equal contribution.\quad $^\dagger$Corresponding author.\quad $^\ddagger$Work done at Alipay.}
}

\begin{abstract}
Time series forecasting is critical across finance, healthcare, and cloud computing, yet progress is constrained by a fundamental bottleneck: the scarcity of large-scale, high-quality benchmarks. To address this gap, we introduce \textsc{QuitoBench}, a regime-balanced benchmark for time series forecasting with coverage across eight trend$\times$seasonality$\times$forecastability (TSF) regimes, designed to capture forecasting-relevant properties rather than application-defined domain labels. The benchmark is built upon \textsc{Quito}, a billion-scale time series corpus of application traffic from Alipay spanning nine business domains. Benchmarking 10 models from deep learning, foundation models, and statistical baselines across 232,200 evaluation instances, we report four key findings: \circone a context-length crossover where deep learning models lead at short context ($L=96$) but foundation models dominate at long context ($L \ge 576$); \circtwo forecastability is the dominant difficulty driver, producing a $3.64 \times$ MAE gap across regimes; \circthree deep learning models match or surpass foundation models at $59 \times$ fewer parameters; and \circfour scaling the amount of training data provides substantially greater benefit than scaling model size for both model families. These findings are validated by strong cross-benchmark and cross-metric consistency. Our open-source release enables reproducible, regime-aware evaluation for time series forecasting research.
\end{abstract}

\section{Introduction}
\label{section:intro}

\begin{wrapfigure}{r}{0.42\textwidth}
    \centering
\vspace{-1.2\baselineskip}
\includegraphics[width=\linewidth]{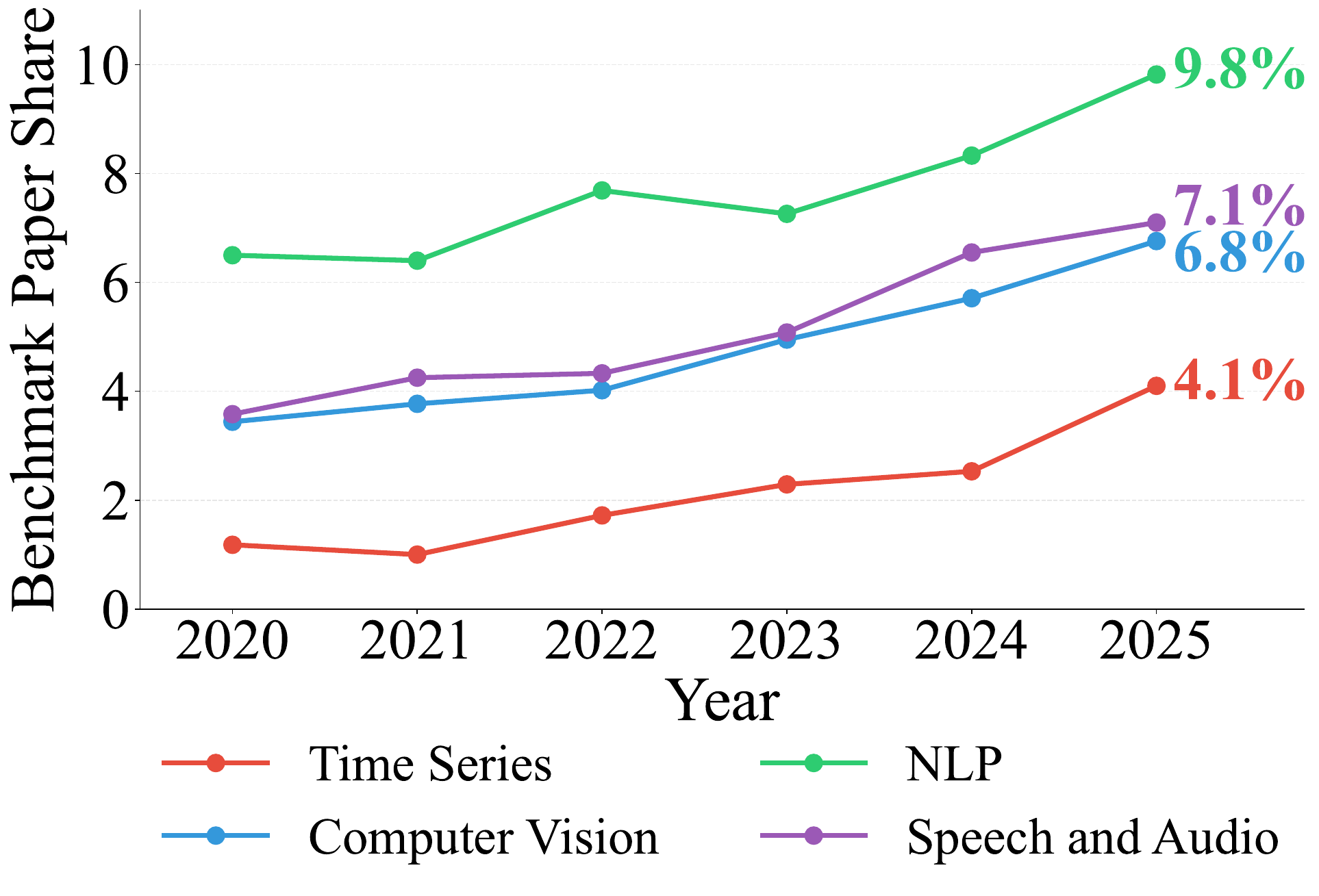}
\caption{Benchmark contribution rate across domains. Time series has the lowest share (4.2\%) vs.\ NLP (9.9\%), speech (7.0\%), and vision (6.8\%). See \cref{app:arxiv}.}
    \label{fig:benchmark_trend}
\vspace{-1.0\baselineskip}
\end{wrapfigure}

Time series forecasting drives high-stakes decisions in finance~\citep{cao2023fints}, healthcare~\citep{xue2024easytpp}, and cloud operations~\citep{liu2022pyraformer}.
Recent years have seen an explosion of foundation models for time series~\citep{jin2023large,ansari2025Chronos,das2023TimesFM}, yet \citet{meyer2025timeseries} warn that the field is heading towards an evaluation crisis analogous to that of large language models: as pre-training corpora grow, benchmark integrity erodes.
We identify three interconnected challenges that undermine the reliability of current forecasting evaluations.

\textbf{Challenge 1: No unified benchmark ecosystem.}
While counterpart fields converge on canonical benchmarks (ImageNet~\citep{deng2009imagenet} and COCO~\citep{lin2014coco} for vision, GLUE~\citep{wang2018glue} for NLP, LibriSpeech~\citep{panayotov2015librispeech} for speech), time series forecasting lacks a comparable standard.
Surveying arXiv papers from 2020--2025, we find that time series has the lowest rate of benchmark-dedicated publications among all four domains (\cref{fig:benchmark_trend}); while vision continues to produce specialised benchmarks for emerging tasks~\citep{gao2026lookbench}, time series practitioners are left to evaluate models on ad-hoc, often incomparable dataset assemblies.

\begin{figure}[h]
    \centering
    \begin{subfigure}[t]{0.32\textwidth}
        \centering
        \includegraphics[width=\linewidth]{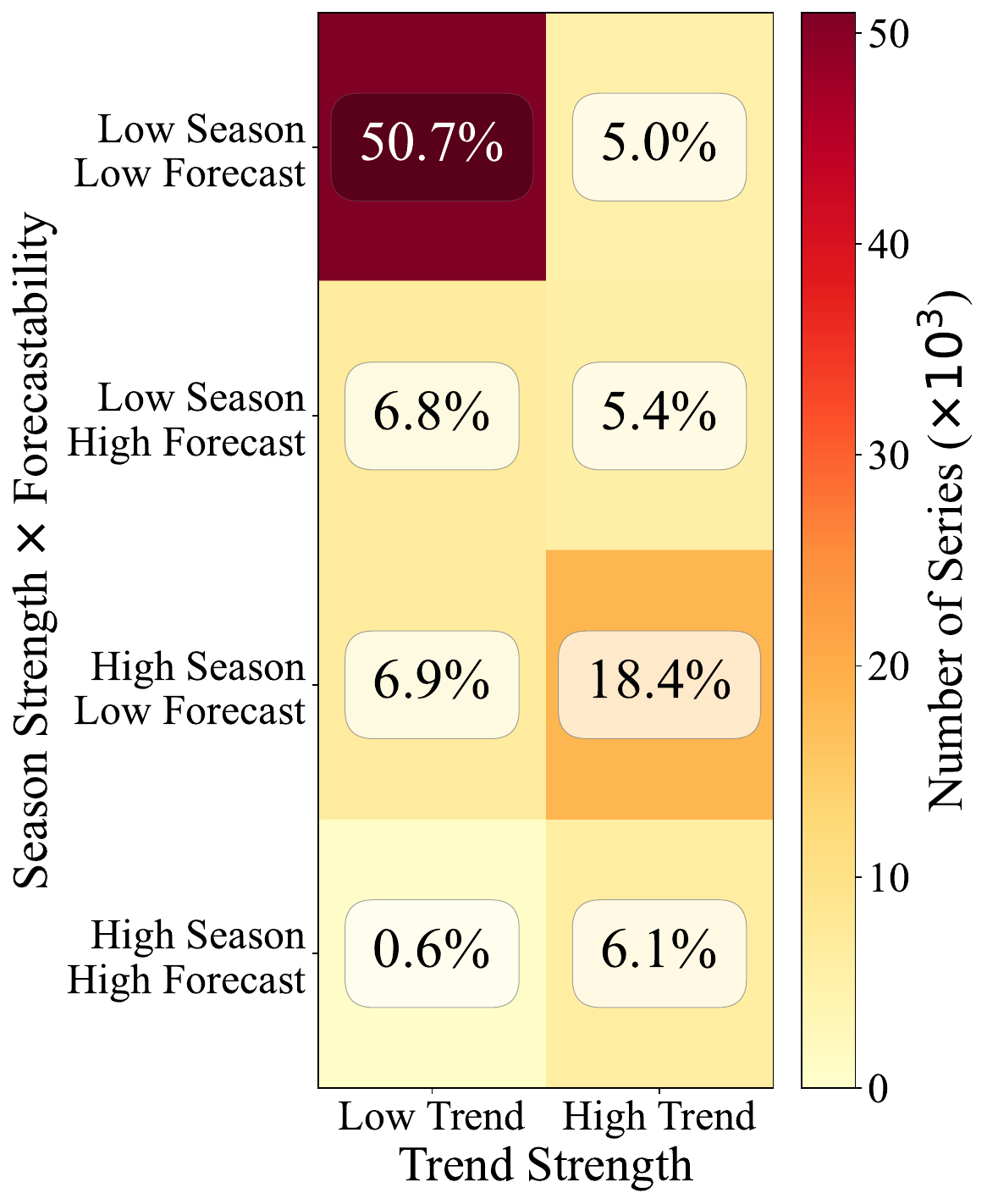}
        \caption{GIFT-Eval}
    \label{fig:gifteval_grid}
    \end{subfigure}
\hfill
    \begin{subfigure}[t]{0.32\textwidth}
    \centering
        \includegraphics[width=\linewidth]{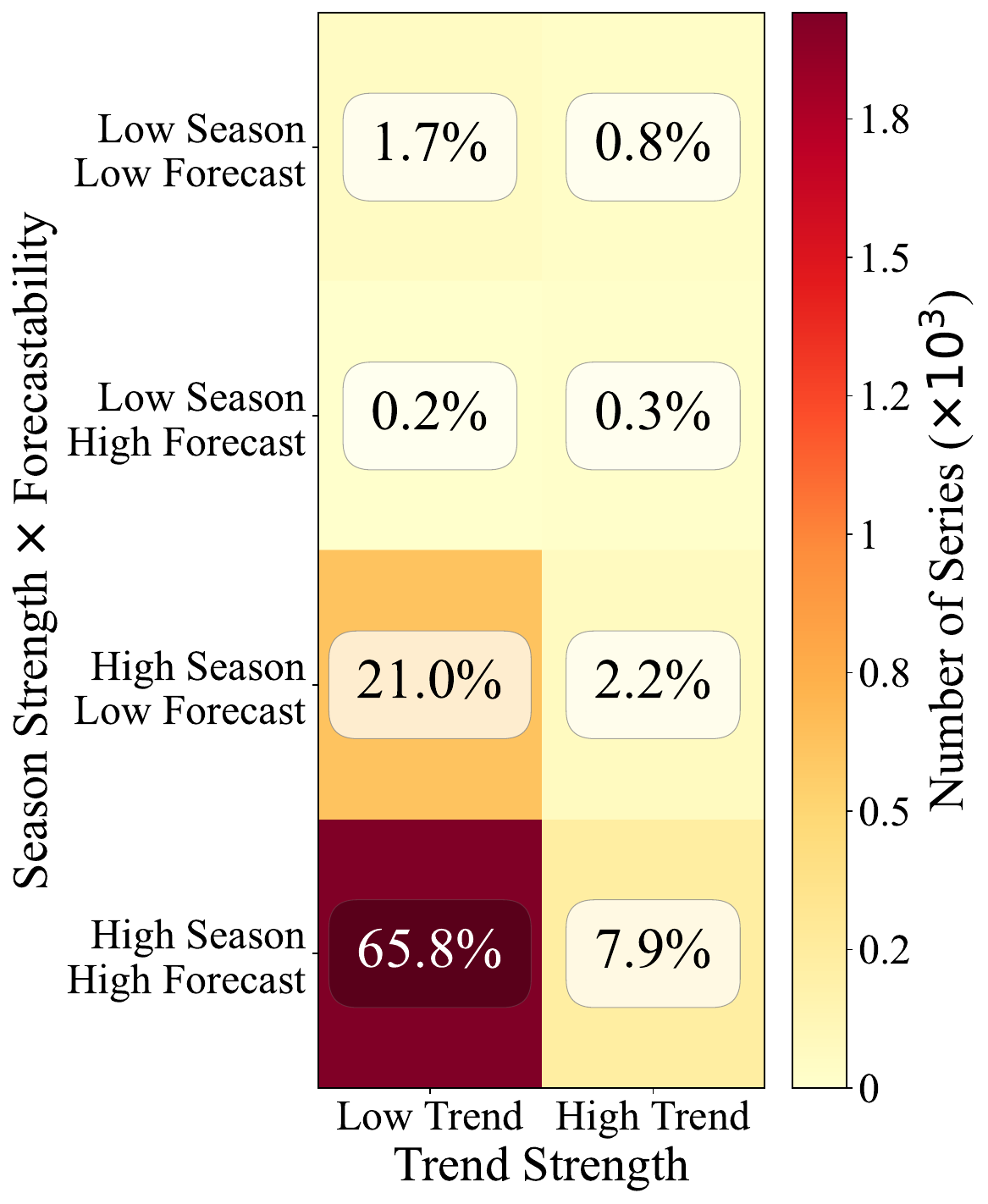}
        \caption{Timer}
    \label{fig:timer_grid}
    \end{subfigure}
\hfill
    \begin{subfigure}[t]{0.32\textwidth}
    \centering
        \includegraphics[width=\linewidth]{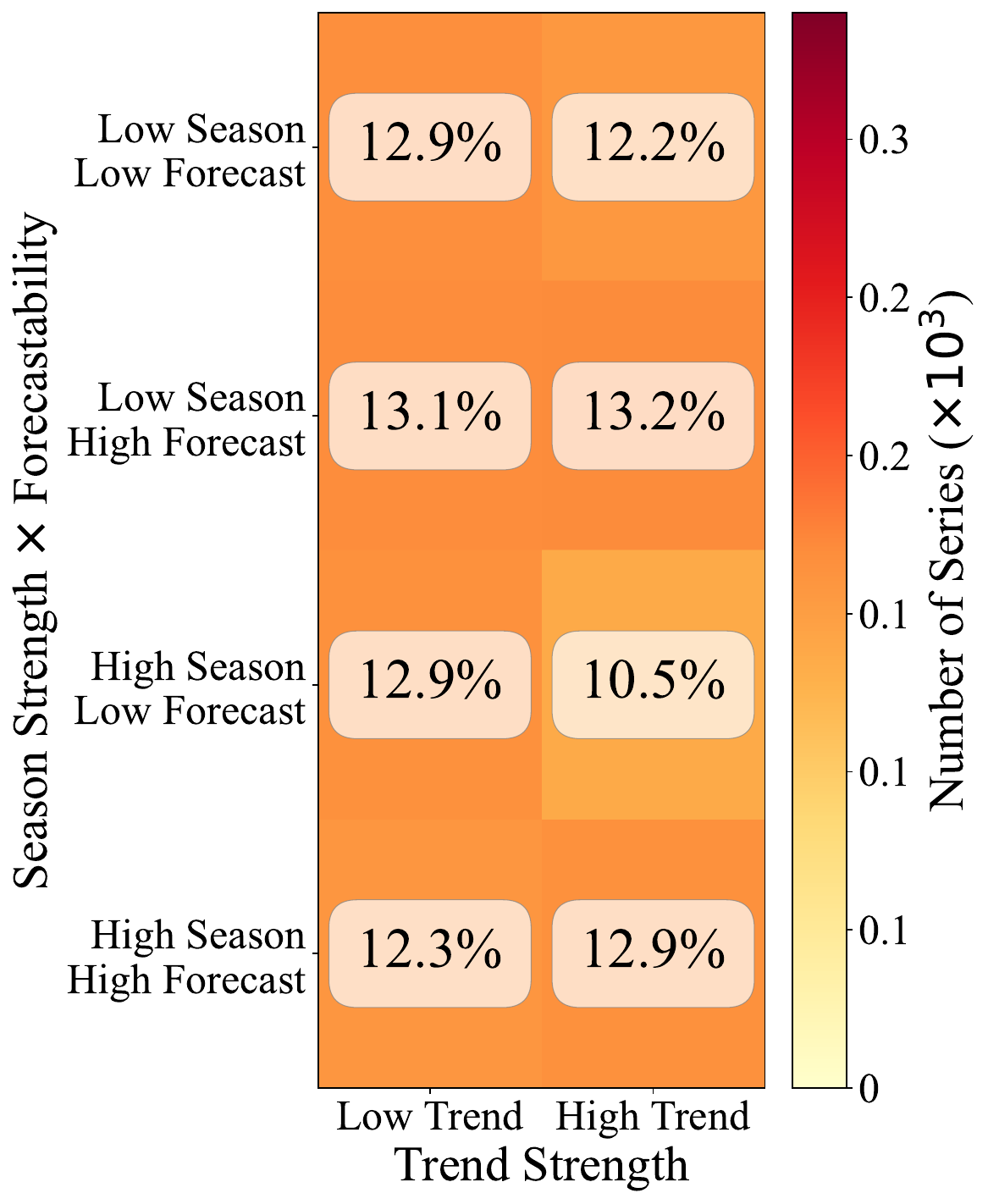}
        \caption{\textsc{Quito}}
    \label{fig:quito_grid}
    \end{subfigure}
    \caption{
        8-grid TSF regime classification across three benchmarks. \subref{fig:gifteval_grid}.~GIFT-Eval is highly imbalanced, with 50.7\% of series in a single low-structure regime. \subref{fig:timer_grid}.~Timer concentrates 65.8\% in the high-seasonality, high-forecastability regime. \subref{fig:quito_grid}.~\textsc{Quito} distributes series near-uniformly (${\sim}12\%$ per regime).
    }
    \label{fig:combined_grid}
\end{figure}

\textbf{Challenge 2: Flawed existing benchmarks.}
The few large-scale benchmarks that do exist (notably GIFT-Eval~\citep{aksu2024gifteval} and Timer~\citep{liu2024timer}\footnote{By \emph{Timer benchmark} we refer to the test sets (ETT~\citep{zhou2021informer}, ECL, Traffic, Weather~\citep{wu2021autoformer}, and PEMS~\citep{chen2001freeway}) adopted in the Timer paper---collectively the most widely used evaluation suite in time series forecasting research.
}) exhibit four structural weaknesses:
\circone~Coarse categorization: existing benchmarks group series by application domain (\eg electricity, traffic, weather), yet there is no systematic justification for why these domain labels should predict forecasting difficulty. A domain such as ``traffic'' encompasses an uncountable variety of temporal dynamics---from smooth commuter flows to bursty event-driven spikes---so two traffic series can differ far more in predictability than a traffic and an electricity series that share similar structure. Intrinsic statistical properties such as trend strength, seasonality, and forecastability are more principled descriptors of what makes a series easy or hard to forecast, yet no current benchmark stratifies evaluation along these axes.
\circtwo~Distributional skew: when series are categorized by these intrinsic properties, GIFT-Eval concentrates 50.7\% of series in a single TSF regime and Timer concentrates 76.2\% (\cref{fig:gifteval_grid,fig:timer_grid}), so aggregate metrics are dominated by the most prevalent data type.
\circthree~Information leakage: \citet{meyer2025timeseries} identify two leakage channels: direct overlap from multi-purpose reuse of public datasets across training and evaluation pipelines, and indirect leakage from temporally correlated series sharing common causal drivers. Assembling heterogeneous public sources with unclear provenance leaves both channels open.
\circfour~Short-series bias: 50\% of GIFT-Eval series contain fewer than 200 time points (\cref{fig:gifteval_cumul}), precluding long-context evaluation.

\noindent
\begin{minipage}[t]{0.42\textwidth}
    \centering
    \begingroup
    \setlength{\tabcolsep}{3pt}        
    \renewcommand{\arraystretch}{0.95} 
    \begin{small}
    \begin{sc}
    \begin{tabular}{@{}l l l@{}}       
        \toprule
        Subset     & Quito-Min   & Quito-Hour\\
        \midrule
        \# Series  & 22,522      & 12,544 \\
        \addlinespace[0.3em]
        \# Tokens  & 0.7 Billion & 1.0 Billion \\
        \addlinespace[0.3em]
        Frequency  & 10 minute   & 1 hour \\
        \addlinespace[0.3em]
        \multirow{2}{*}{Start time} & 2023-07-10 & 2021-11-18 \\
                                    & 00:00:00   & 04:00:00   \\
                                   \addlinespace[0.3em]
        \multirow{2}{*}{End time}   & 2023-08-19 & 2023-08-19 \\
                                    & 23:50:00   & 23:00:00   \\
                                   \addlinespace[0.3em]
        Length     & 5,904       & 15,356 \\
        \addlinespace[0.3em]
        \# Variates& 5           & 5 \\
        \bottomrule
    \end{tabular}
    \captionof{table}{Key statistics of two subsets of \textsc{Quito} (1.6B tokens combined).}
    \label{tab:stat_quito}
    \end{sc}
    \end{small}
    \endgroup
    \end{minipage}
\hfill
    \begin{minipage}[t]{0.55\textwidth}
    \centering
    \begingroup
    \setlength{\tabcolsep}{3pt}
    \renewcommand{\arraystretch}{0.95}
    \begin{small}
    \begin{sc}
    \begin{tabular}{@{}l l ll@{}}  
        \toprule
        & \multirow{2}{*}{Overall} & \multicolumn{2}{c@{}}{Curated from} \\
        \cmidrule(lr){3-4}
        &        & Quito-Min & Quito-Hour \\
         \midrule
        \# Series & 1,290 & 773 & 517 \\
        \midrule
        \textsc{Train set} &  &  &  \\
        \quad \# Tokens & 7,726,550 & 1,603,202 & 6,123,348 \\
        \quad    Length & -         &   2,074   &  11,844 \\
        \midrule
        \textsc{Valid set} &  &  &  \\
        \quad \# Tokens & 1,930,734 & 400,414   & 1,530,320 \\
        \quad    Length & -         &    518    &   2,960 \\
        \midrule
        \textsc{Test set} &  &  &  \\
        \quad \# Tokens & 2,845,560 & 2,560,176 & 285,384 \\
        \quad    Length & -         &  3,312    &    552 \\
        \bottomrule
    \end{tabular}
    \captionof{table}{Key statistics of \textsc{QuitoBench}.}
    \label{tab:stat_quito_bench}
    \end{sc}
    \end{small}
    \endgroup
\end{minipage}

\vspace{0.2cm}
\textbf{Challenge 3: No practical model selection guidance.}
With over 20 time series foundation models published in the last two years alone~\citep{meyer2025timeseries}, practitioners face a pressing question: when is a 200\,M-parameter foundation model worth deploying over a 1M-parameter deep learning model trained from scratch?
Current benchmarks, compromised by the issues above, lack the scale, balance, and evaluation rigour to answer this reliably across the axes that matter: context length, forecast horizon, forecasting mode, and intrinsic data characteristics.

To address all three challenges, we present \textsc{Quito}, a billion-scale, single-provenance time series dataset of application traffic from Alipay's production platform, covering nine business verticals from finance and e-commerce to infrastructure and IoT (1.6\,B tokens, two granularities, uniformly long series; \cref{tab:stat_quito,fig:vertical_dist}), and \textsc{QuitoBench} (\cref{tab:stat_quito_bench}), the first forecasting benchmark that explicitly balances evaluation series across all eight trend$\times$seasonality$\times$forecastability (TSF) regime cells (\cref{fig:quito_grid}).
The design directly resolves each flaw above:
\textbf{(D1)}\label{item:d1} Characteristic-based categorization: series are categorized by their intrinsic statistical properties (trend, seasonality, and forecastability) rather than coarse domain labels, directly exposing the drivers of forecast difficulty (see \cref{app:tsf_justification} for an empirical justification).
\textbf{(D2)}\label{item:d2} Regime-balanced curation: the heavily skewed distributions of GIFT-Eval and Timer are replaced with near-uniform coverage across all eight TSF regimes, so aggregate metrics reflect model capability rather than data prevalence.
\textbf{(D3)}\label{item:d3} Leakage-free evaluation: because every series originates from a single proprietary operational environment with no overlap with any public pre-training corpus, both direct and indirect information leakage are eliminated by construction.
\textbf{(D4)}\label{item:d4} Uniformly long series: all series span 5,900--15,300 time steps, enabling rigorous evaluation at context lengths up to 1,024\@, far beyond the reach of short-series benchmarks. Beyond the benchmark itself, \textsc{Quito} serves as a high-quality training resource: \textbf{(D5)}\label{item:d5} its sanitised, single-provenance series support both training from scratch and fine-tuning foundation models, enabling controlled data-scaling studies.

Leveraging this design, we evaluate ten models spanning three foundation models (30--200\,M parameters) and five deep learning architectures (0.3--5\,M parameters) across 18 configurations (3 context lengths $\times$ 3 horizons $\times$ 2 modes), generating ${\sim}1.6{\times}10^7$ predictions per model through dense rolling-window evaluation (\cref{fig:rolling_windows}).
This scale of evaluation yields four key findings that would be obscured by conventional benchmarks:
\circone~Context-length crossover (enabled by \hyperref[item:d4]{D4}): deep learning models lead at short context ($L = 96$), yet foundation models overtake them at $L\ge576$, making history length the primary factor in model selection.
\circtwo~Regime specialization (enabled by \hyperref[item:d1]{D1}, \hyperref[item:d2]{D2}): foundation models dominate 6 of 8 TSF regimes while deep learning retains advantages in the remaining low-seasonality regimes, with forecastability emerging as the dominant difficulty axis.
\circthree~Parameter efficiency (enabled by \hyperref[item:d3]{D3}, \hyperref[item:d2]{D2}): deep learning models match or surpass foundation models at $59{\times}$ fewer parameters, and degrade more gracefully as forecast horizon increases.
\circfour~Data scaling (enabled by \hyperref[item:d5]{D5}): for both model families, increasing the amount of training data yields substantially larger gains than increasing model size.

These results translate into clear model-selection guidance. 
For short-context forecasting (\(L=96\)) or resource-constrained deployment, compact deep learning models such as CrossFormer (\( \sim \)1M parameters) are the strongest default, offering the best accuracy--efficiency trade-off. When longer histories are available (\(L \geq 576\)), especially for strongly seasonal series, foundation models such as Chronos-2 (\( \sim \)100M parameters) become the preferred choice. 
Regime also matters: foundation models are generally more effective in highly seasonal, more forecastable settings, whereas deep learning models remain competitive in low-seasonality regimes and degrade more gracefully as the forecast horizon increases. Across both model families, scaling training data is consistently more beneficial than scaling parameter count. 
More broadly, our results challenge the assumption that larger pre-trained models are uniformly superior for time series forecasting: task-specific architectures can match or exceed foundation-model performance at up to \(59\times\) fewer parameters. Because \textsc{QuitoBench} is designed to avoid the information leakage that affects prior evaluations, it enables more reliable comparison across model classes. We release the dataset, code, and evaluation framework to support reproducible and regime-aware future research.


\vspace{-0.15cm}

\section{\textsc{Quito} Description}
\label{section:bench}

\begin{figure}[t]
    \centering
    \includegraphics[width=\linewidth]{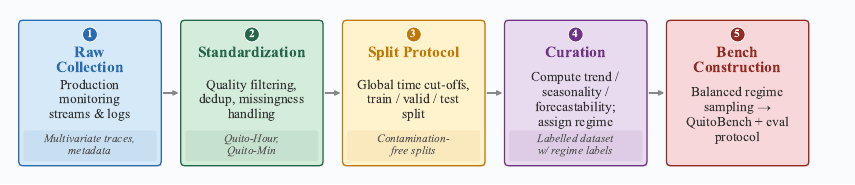}
    \caption{
        Overall pipeline of \textsc{Quito} and \textsc{QuitoBench}.
        It contains five key stages: (1) Raw collection, (2) sanitization and standardization, (3) leakage-free temporal splitting, (4) trend/seasonality/forecastability computation and regime labeling, and (5) balanced \textsc{QuitoBench} construction and evaluation.
    }
    \label{fig:pipeline}
\end{figure}
\vspace{-0.15cm}

\subsection{Overall Pipeline}
\label{sec:pipeline}

\Cref{fig:pipeline} summarizes the end-to-end workflow that converts raw application traffic telemetry into the released \textsc{Quito} datasets and the derived benchmark \textsc{QuitoBench}. 

\vspace{2pt}\noindent\textbf{Data sources and series formation.}
Our data are collected from the production platform of Alipay, one of the world's largest digital payment and lifestyle platforms.
Each series records the traffic workload of a distinct application service; collectively, these services span nine business verticals, including finance, commerce, advertising, platform infrastructure, risk and compliance, which reflect the breadth of a full-scale digital economy rather than a single narrow domain (\cref{fig:vertical_dist}).
The \emph{workload} of each application is driven by the traffic from several request subtypes (\eg remote procedure calls and message subscriptions)~\citep{xue_meta_2022}, represented at each time step as a multi-dimensional vector.
Each workload trace is assigned a unique identifier (\texttt{item\_id}) and contains five variates, each corresponding to a distinct traffic subtype; due to commercial sensitivity, we anonymize subtype semantics and denote the channels as $\texttt{index}_i$ for $i=1,\dots,5$.
We construct two complementary subsets at different temporal granularities (minute-level and hour-level), which are constructed from two disjoint pools of raw workload traces (\ie no overlap in \texttt{item\_id}s).
The two subsets have different start dates (2023-07-10 for \textsc{Quito-Min} and 2021-11-18 for \textsc{Quito-Hour}): high-frequency 10-minute telemetry is subject to a shorter retention window in the production system, whereas hourly aggregates are archived long-term.
Both subsets share the same end date (2023-08-19), which also serves as the global cutoff for \textsc{QuitoBench}, ensuring leakage-free splits across both granularities.
We then align the five channels on a shared timeline and aggregate them into standardized minute- and hour-level multivariate series; see \cref{app:quito_preprocessing} for details.
Overall, this yields approximately 500{,}000 workload series.

\vspace{2pt}\noindent\textbf{Sanitization and standardization.}
After filtering, we retain 14,244 minute-level and 16,746 hour-level series, released as \textsc{Quito-Min} and \textsc{Quito-Hour}. See \cref{app:quito_dedup} for details.

\vspace{2pt}\noindent\textbf{\textsc{TSF} diagnostics and regime labeling.}
We characterize each series along three fundamental axes of time series behavior and use them to define a discrete regime taxonomy.

\begin{definition}[\textsc{TSF} Regime]\label{def:tsf}
    Let $\{x_t\}$ be a time series. Its \emph{TSF profile} is the triple $(T, S, F) \in [0,1]^3$, where 
    \circone $T$ (\emph{trend strength}) measures long-range drift, computed as the fraction of variance explained by the trend component of an STL decomposition~\citep{cleveland1990stl}; \circtwo $S$ (\emph{seasonality strength}) measures periodic structure, computed analogously from the seasonal component~\citep{wen2022robust};
    \circthree $F$ (\emph{forecastability}) measures signal regularity, defined as $F = 1 - H$ where $H$ is the normalized spectral entropy~\citep{welch1967use}.
    
    Given a threshold $\tau$ (default $\tau{=}0.4$), each diagnostic is binarized into \textsc{high} ($>\tau$) or \textsc{low} ($\le\tau$), yielding $2^3{=}8$ \emph{\textsc{TSF} regime cells} denoted \textsc{trend\_seasonality\_forecastability} (\eg \textsc{high\_high\_high}).
    For multivariate series, $T$, $S$, and $F$ are averaged across all variate channels before binarization.
\end{definition}

Concretely, $T$ and $S$ are obtained via STL (Seasonal-Trend decomposition using LOESS) with robust fitting to resist the outliers common in operational traffic data.
STL separates each univariate channel into additive trend ($\tau_t$), seasonal ($s_t$), and residual ($r_t$) components with the seasonal period set to one daily cycle ($p{=}144$ for 10-minute series, $p{=}24$ for hourly series).
Following~\citet{wen2022robust}, $T$ and $S$ measure the fraction of variance explained by trend and seasonality relative to the residual: a value near~1 means the component dominates, while a value near~0 means it is negligible relative to noise.
Forecastability $F{=}1{-}H$, where $H$ is the normalized spectral entropy estimated via Welch's method~\citep{welch1967use}: $F{=}1$ corresponds to a signal whose energy is concentrated in a few frequencies (highly predictable), while $F{=}0$ corresponds to white noise with a flat spectrum.
A binary split keeps the taxonomy tractable: finer quantisation (\eg quintiles) causes a combinatorial explosion of regime cells and data-sparse groups.
These regime labels allow us to analyze model behaviour across diverse operational dynamics and later serve as the basis for constructing a balanced benchmark (\cref{sec:benchmark_construction}); see \cref{app:tsf_diagnostics} for exact formulations and sensitivity analysis.

\subsection{Benchmark Construction: \textsc{QuitoBench}}
\label{sec:benchmark_construction}

\vspace{2pt}\noindent\textbf{Leakage-free temporal splitting.}
We apply a global temporal cutoff at 2023-07-28 00:00:00, ensuring that all series share the same train/validation/test boundary and that no future information leaks into training.
Data before the cutoff is divided into train (80\%) and validation (20\%) splits; data from the cutoff onward forms the test set.
The resulting test lengths are 3,312 time points per series for \textsc{Quito-Min} and 552 for \textsc{Quito-Hour}; see \cref{app:quitobench_testset} for full split statistics.

\vspace{2pt}\noindent\textbf{Balanced benchmark construction.}
A key design goal of \textsc{QuitoBench} is to prevent prevalence-driven conclusions where headline metrics are dominated by the most common (and often easiest) regime.
Natural time-series collections are highly skewed: shown in \cref{fig:gifteval_grid,fig:timer_grid}, GIFT-Eval places 50.7\% of its series in a single low/low/low cell (stationary, noisy, and unpredictable), while Timer is dominated by the low/high/high regime at 65.8\% (TSF diagnostics for these benchmarks are computed following \cref{app:tsf_external}).
In such settings, a model that specializes in the majority regime can rank near the top on aggregate metrics while failing silently on remaining regimes.

To remedy this, we apply stratified sampling over the eight \textsc{TSF} regime cells.
After assigning each candidate series its regime label using training data (stage~4 of the pipeline), we partition the test pool by cell and draw up to a fixed quota of approximately 162 series per cell.
This yields 1,290 test series (773 from \textsc{Quito-Min} and 517 from \textsc{Quito-Hour}) distributed near-uniformly across all eight cells ($\sim\!160$ series/cell; 10.5\%--13.2\%; see \cref{fig:quito_grid}).
The balanced design supports two complementary aggregation modes:
\emph{micro-averaged} mean rank averages over all 1,290 individual series, reflecting overall expected performance;
\emph{macro-averaged} mean rank first aggregates within each of the eight cells and then averages across cells, weighting all regime behaviors equally regardless of cell size.
Together these two views reduce the confounding effect of regime prevalence and enable fine-grained, per-regime diagnostic reporting, revealing where models succeed or fail across distinct operational dynamics.

\section{Experiments}

\subsection{Experimental Setup}
\label{sec:exp_setup}

\vspace{2pt}\noindent\textbf{Benchmarked models.}
We evaluate ten models spanning three families. 
    \circone Deep learning models: Crossformer~\citep{zhang2023crossformer}, DLinear~\citep{zeng2023dlinear}, iTransformer~\citep{liu2024itransformer}, PatchTST~\citep{nie2023patchtst}, and TSMixer~\citep{chen2023tsmixer}. 
    \circtwo Foundation models: Chronos-2~\citep{ansari2025Chronos}, TimesFM-2.5~\citep{das2023TimesFM}, and TiRex~\citep{25tirex}.
    \circthree Statistical baselines: Exponential Smoothing (ES) and Seasonal Na\"ive (SNaive). 
    
The deep learning models cover diverse architectures, from linear projections to patch-based and cross-dimensional attention transformers, while the foundation models represent leading zero-shot pre-trained approaches; see \cref{tab:model_sizes} for the full breakdown.

\vspace{2pt}\noindent\textbf{Evaluation configurations.}
Each model is evaluated under $3 \times 3 \times 2 = 18$ task configurations: three context lengths $L \in \{96, 576, 1024\}$, three forecast horizons $H \in \{48, 288, 512\}$, and two forecasting modes.
In multivariate (MV) mode, all five variate channels are provided jointly as input and predicted simultaneously; in univariate (UV) mode, each channel is processed independently.
Across 1,290 test series, 18 configurations, and 10 models, this spans 232,200 aggregated (series, configuration, model) evaluation instances.

\vspace{2pt}\noindent\textbf{Rolling window evaluation.}
Unlike benchmarks such as \textsc{gift-eval}, which use non-overlapping windows (stride $H$, capped at 20), \textsc{QuitoBench} employs \emph{dense rolling windows} with unit stride: for each $(L,H)$ pair the window slides one step at a time, producing $W(H) = T_{\text{test}} - H + 1$ windows per series (up to 1,489 at $H{=}48$).
This yields ${\sim}1.6{\times}10^7$ predictions per model, which is one to two orders of magnitude more than sparse schemes and substantially stabilises per-series MAE estimates.
See \cref{app:rolling_windows} and \cref{fig:rolling_windows} for details.

\vspace{2pt}\noindent\textbf{Training protocol.}
Deep learning models follow a three-stage pipeline per configuration: hyperparameter tuning on the validation split, training from scratch with the selected hyperparameters, and evaluation of the best checkpoint on the test split.
To reduce variance, each model is run under three random seeds and the mean MAE is reported.
Foundation models are applied zero-shot without any gradient updates on \textsc{QuitoBench} data; see \cref{app:training_details} for full details.

\vspace{2pt}\noindent\textbf{Metrics.}
We report MAE as the primary metric and convert per-series MAE to rank scores (1--10) to aggregate fairly across heterogeneous scales; the mean rank is reported over all series and configurations (See \cref{app:metrics} for details). We also report MSE in \cref{tab:quito_group_mse} and \cref{tab:quito_group_mse_rank}. All metrics show remarkable consistency (See \cref{app:quito_mse} for details).

\subsection{Results and Analysis}
\label{sec:analysis}

\vspace{2pt}\noindent\textbf{Main results.} 
\Cref{tab:main_results} summarizes mean MAE and mean rank for all ten models across 232,200 evaluation instances.
CrossFormer achieves the best overall mean rank of 2.86 and the lowest MAE of 0.279, ranking first under both MV and UV modes.
Chronos-2 is the only foundation model in the top two with rank 3.36, driven by stronger multivariate performance; the remaining foundation models TimesFM-2.5 and TiRex place mid-table, while DLinear lags at 7.26, consistent with its simpler linear design.
At the category level, deep learning models average MAE 0.312 versus 0.319 for foundation models, a gap that is not statistically significant (Cohen's $d{=}{-}0.067$ for CrossFormer vs.\ Chronos-2); see \cref{app:stat_tests} for all statistical tests.
Statistical baselines ES and SNaive rank last by a wide margin, confirming the benchmark is discriminative.

\begin{figure}[h]
    \centering
    \begin{subfigure}[t]{0.48\linewidth}
        \centering
        \includegraphics[width=\linewidth]{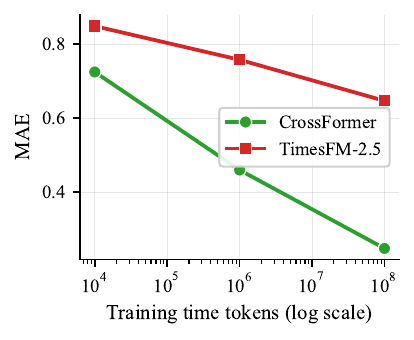}
        \caption{Data scaling: MAE vs.\ training time tokens.}
        \label{fig:data_scaling}
    \end{subfigure}
\hfill
    \begin{subfigure}[t]{0.48\linewidth}
        \centering
        \includegraphics[width=\linewidth]{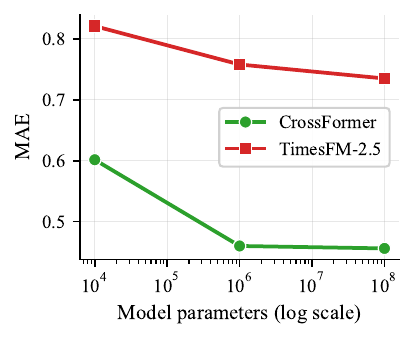}
        \caption{Model scaling: MAE vs.\ model parameters.}
        \label{fig:model_scaling}
    \end{subfigure}
    \caption{Scaling behavior on \textsc{Quito} for CrossFormer (deep learning) and TimesFM-2.5 (foundation model). More data yields far larger gains than more parameters for both models.}
    \label{fig:scaling}
\end{figure}

While foundation models are evaluated zero-shot, we fine-tuned TimesFM-2.5 on \textsc{Quito} training data, yet its MAE remains higher than CrossFormer's (\cref{fig:scaling}); see details in Analysis I.

\begin{table}[h]
    \centering
    \small
    \begin{sc}
    \begin{tabular}{llrrrrrr}
        \toprule
        &  & \multicolumn{3}{c}{Mean rank} & \multicolumn{3}{c}{Mean MAE} \\
        \cmidrule(lr){3-5}\cmidrule(lr){6-8}
        Model & Category & MV & UV & Overall & MV & UV & Overall \\
        \midrule
        \rowcolor{blue!8}CrossFormer  & Deep learning  & \textbf{3.05} & \textbf{2.67} & \textbf{2.86} & \textbf{0.282} & \textbf{0.275} & \textbf{0.279} \\
        Chronos-2    & Foundation     & 3.21 & 3.51 & 3.36 & 0.310 & 0.317 & 0.314 \\
        TimesFM-2.5  & Foundation     & 4.21 & 4.21 & 4.21 & 0.319 & 0.319 & 0.319 \\
        PatchTST     & Deep learning  & 4.37 & 4.34 & 4.35 & 0.299 & 0.298 & 0.299 \\
        TiRex        & Foundation     & 4.36 & 4.36 & 4.36 & 0.322 & 0.322 & 0.322 \\
        iTransformer & Deep learning  & 4.56 & 4.78 & 4.67 & 0.299 & 0.302 & 0.301 \\
        TSMixer      & Deep learning  & 5.58 & 5.43 & 5.51 & 0.313 & 0.309 & 0.311 \\
        DLinear      & Deep learning  & 7.24 & 7.29 & 7.26 & 0.368 & 0.371 & 0.369 \\
        ES           & Baseline       & 9.18 & 9.18 & 9.18 & 0.695 & 0.695 & 0.695 \\
        SNaive       & Baseline       & 9.25 & 9.25 & 9.25 & 0.675 & 0.675 & 0.675 \\
        \bottomrule
    \end{tabular}
    \end{sc}
    \vspace{4pt}
    \caption{Overall performance of all ten models sorted by overall mean rank (lower is better). Mean rank and mean MAE are each broken down by MV/UV mode and averaged overall.}
    \label{tab:main_results}
\end{table}
\vspace{-0.2cm}

\vspace{2pt}\noindent\textbf{Analysis I: Scaling laws for data and model size.}
A natural question is whether time series forecasting exhibits scaling laws analogous to those observed in language modeling?
Due to computational constraints, we select one representative model from each family---CrossFormer (deep learning, 1\,M parameters) and TimesFM-2.5 (foundation model, 200\,M parameters)---and train both on \textsc{Quito} at three logarithmically spaced data and model scales; see \cref{fig:scaling}.

\emph{Data scaling.}
Increasing the training data budget from 10\,K to 100\,M tokens substantially improves performance. CrossFormer's MAE decreases from 0.725 to 0.248, a 66\% reduction that follows a near-linear trend on the log--log plot. Over the same range, TimesFM-2.5 improves from 0.849 to 0.647, a 24\% reduction with a shallower slope (\cref{fig:data_scaling}), suggesting that task-specific architectures extract more value per additional training token.

\emph{Model scaling.}
Varying model size from 10\,K to 100\,M parameters, CrossFormer's MAE drops sharply from 0.602 to 0.456 but plateaus beyond 1\,M parameters (\cref{fig:model_scaling}); TimesFM-2.5 shows a similar plateau (0.821 to 0.735).
Taken together, these scaling curves reveal a clear asymmetry: \emph{more data} is far more valuable than \emph{more parameters} for both model families.

\vspace{-0.15cm}

\vspace{2pt}\noindent\textbf{Analysis II: Effects of context length.}
The ranking between the two model categories reverses as context length $L$ increases, as shown in \cref{tab:context_analysis}.
At $L{=}96$, deep learning models outperform foundation models by 24.6\% in MAE; the advantage flips at $L{=}576$ and widens further at $L{=}1024$, where foundation models lead by 22.0\%.
The asymmetry is stark: foundation models improve 43--50\% when moving from short to long context, whereas deep learning models gain only 7--12\%, as their task-specific parameterisation is already near-optimal. This crossover suggests a functional division between the two families.
Deep learning models are stronger short-context specialists, fitting local dependencies effectively when history is limited.
Foundation models benefit much more from long context, likely because pre-training equips them to exploit recurring motifs, delayed dependencies, and stable seasonal structure once sufficient history is available.
Hence, context length is not merely an evaluation setting but a primary axis of model selection.

\begin{table}[h]
    \small
    \noindent
    \begin{minipage}[t]{0.34\textwidth}
    \centering
    \begin{sc}
    \begin{tabular}{lrr}
        \toprule
        Ctx & Foundation & Deep \\
        $L$ & Models & Learning \\
        \midrule
        \rowcolor{blue!8} 96  & 0.455 & \textbf{0.343} \\
        \rowcolor{red!5} 576  & \textbf{0.256} & 0.293 \\
        \rowcolor{red!8} 1024 & \textbf{0.245} & 0.299 \\
        \bottomrule
    \end{tabular}
    \end{sc}
    \captionof{table}{Mean MAE by context length (Ctx $L$). Bold = winner. \colorbox{blue!8}{Blue}: deep learning leads; \colorbox{red!8}{Red}: foundation model leads.}
    \label{tab:context_analysis}
    \end{minipage}%
\hfill
    \begin{minipage}[t]{0.62\textwidth}
    \centering
    \begin{sc}
    \begin{tabular}{lrr}
        \toprule
        \textsc{tsf} & Cross- & \\
        Regime & Former & Chronos-2 \\
        \midrule
        \textsc{high\_high\_high} & 0.165 & \textbf{0.163} \\
        \textsc{high\_high\_low}  & 0.356 & \textbf{0.353} \\
        \rowcolor{blue!8}\textsc{high\_low\_high} & \textbf{0.180} & 0.349 \\
        \textsc{high\_low\_low}   & \textbf{0.600} & 0.628 \\
        \textsc{low\_high\_high}  & 0.199 & \textbf{0.197} \\
        \textsc{low\_high\_low}   & 0.239 & \textbf{0.235} \\
        \rowcolor{blue!8}\textsc{low\_low\_high}  & \textbf{0.154} & 0.207 \\
        \textsc{low\_low\_low}    & \textbf{0.370} & 0.397 \\
        \bottomrule
    \end{tabular}
    \end{sc}
    \vspace{4pt}
    \captionof{table}{Mean MAE per \textsc{tsf} regime for CrossFormer (deep learning) and Chronos-2 (foundation). Bold = winner. Highlighted: deep learning advantage $>$10\%.}
    \label{tab:group_analysis}
    \end{minipage}
\end{table}
\vspace{-0.15cm}

\vspace{2pt}\noindent\textbf{Analysis III: Effects of forecast horizon.}
MAE increases monotonically with forecast horizon for all models, but degradation rates differ between the two categories; see \cref{tab:horizon_analysis}.
Deep learning models degrade 15--34\% from $H{=}48$ to $H{=}512$, while foundation models degrade 31--37\%, indicating that task-specific architectures are generally more horizon-robust.
DLinear is the most robust at only 14.8\% degradation, though at the cost of a higher baseline MAE; CrossFormer maintains the best absolute MAE at every horizon. This pattern suggests that foundation models derive more of their advantage from short-range structure, whereas deep learning models trained directly on the task remain more stable as uncertainty accumulates over longer horizons.
DLinear highlights this trade-off clearly: it is less accurate overall, but its simple inductive bias makes it especially resistant to horizon growth.
In practice, forecast horizon should be treated as a model-selection constraint, not merely a reporting axis.

\begin{table}[h]
    \centering
    \small
    \begin{sc}
    \begin{tabular}{lrrrrr}
        \toprule
        Model & $H$=48 & $H$=288 & $H$=512 & $\Delta$(48→288) & $\Delta$(48→512) \\
        \midrule
        \rowcolor{blue!8}CrossFormer  & \textbf{0.237} & \textbf{0.283} & \textbf{0.317} & $+$19.3\% & $+$33.9\% \\
        PatchTST     & 0.252 & 0.300 & 0.344 & $+$19.0\% & $+$36.7\% \\
        iTransformer & 0.260 & 0.306 & 0.335 & $+$17.6\% & $+$28.5\% \\
        TSMixer      & 0.273 & 0.316 & 0.345 & $+$15.8\% & $+$26.3\% \\
        \rowcolor{green!8}DLinear & 0.345 & 0.367 & 0.396 & \textbf{$+$6.5\%} & \textbf{$+$14.8\%} \\
        \midrule
        Chronos-2    & 0.262 & 0.321 & 0.358 & $+$22.8\% & $+$37.0\% \\
        TimesFM-2.5  & 0.271 & 0.329 & 0.358 & $+$21.3\% & $+$32.1\% \\
        TiRex        & 0.276 & 0.331 & 0.361 & $+$19.9\% & $+$30.7\% \\
        \bottomrule
    \end{tabular}
    \end{sc}
    \vspace{4pt}
    \caption{Mean MAE at each forecast horizon and percentage degradation relative to $H = 48$. \colorbox{blue!8}{Blue}: best MAE (CrossFormer). \colorbox{green!8}{Green}: smallest degradation (DLinear, $+$14.8\%). Bold = best in column.}
    \label{tab:horizon_analysis}
\end{table}
\vspace{-0.15cm}

\vspace{2pt}\noindent\textbf{Analysis IV: TSF regime analysis: forecastability, specialization, and pathological regimes.}
\label{sec:analysis_forecastability}

\begin{figure}[h]
    \centering
    \begin{minipage}[c]{0.55\textwidth}
        \centering
        \includegraphics[width=\linewidth]{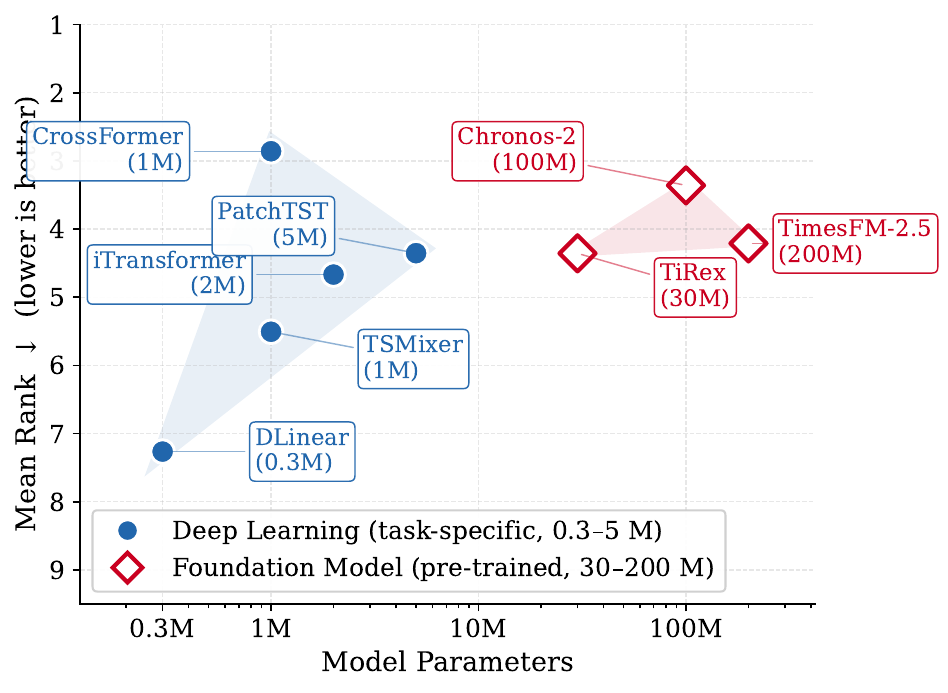}
        \caption{Efficiency frontier: mean rank vs.\ model scale. Deep learning models (blue, 0.3--5\,M) match or beat foundation models (red, 30--200\,M) at $58{\times}$ fewer parameters.}
        \label{fig:efficiency_main}
    \end{minipage}%
\hfill
    \begin{minipage}[c]{0.42\textwidth}
        \centering
        \small
        \begin{sc}
        \begin{tabular}{clr}
            \toprule
            Rank & TSF Regime & Mean \\
            &  &  MAE \\
            \midrule
            \rowcolor{green!8} 1 (Easiest) & \textsc{high\_high\_high} & \textbf{0.205} \\
            \rowcolor{green!5} 2 & \textsc{low\_low\_high} & 0.220 \\
            3 & \textsc{low\_high\_high} & 0.299 \\
            4 & \textsc{low\_high\_low} & 0.359 \\
            5 & \textsc{high\_low\_high} & 0.376 \\
            6 & \textsc{low\_low\_low} & 0.456 \\
            \rowcolor{red!5} 7 & \textsc{high\_high\_low} & 0.478 \\
            \rowcolor{red!8} 8 (Hardest) & \textsc{high\_low\_low} & \textbf{0.749} \\
            \bottomrule
        \end{tabular}
        \end{sc}
        \vspace{2pt}
        \captionof{table}{TSF regime difficulty ranking by mean MAE. 3.64$\times$ gap between easiest and hardest. Full table in \cref{tab:forecastability_analysis}.}
        \label{tab:regime_difficulty}
    \end{minipage}
\end{figure}

\emph{Forecastability as the dominant difficulty driver.}
Among the three TSF dimensions, forecastability yields the largest error separation: mean MAE increases from 0.278 on high-forecastability series to 0.505 on low-forecastability series (1.81$\times$), whereas trend and seasonality produce substantially smaller gaps.
Consistently, the easiest regime, \textsc{high\_high\_high}, is 3.64$\times$ easier than the hardest, \textsc{high\_low\_low} (\cref{tab:regime_difficulty}).
This is expected from the diagnostic definitions: trend and seasonality measure the strength of individual STL components relative to the residual, whereas forecastability captures overall temporal regularity and is therefore a more direct proxy for intrinsic difficulty than coarse domain labels.
Foundation models are more robust under low forecastability, degrading by $\sim$ 1.8$\times$ versus $\sim$ 2.3$\times$ for deep learning models, suggesting better noise tolerance from pre-training.

\emph{Model specialization across regimes.}
At the regime level, foundation models win six of eight TSF regimes, specifically those with high seasonality or high forecastability, where recurring structure aligns well with patterns acquired during pre-training; see \cref{tab:group_analysis}.
Deep learning models dominate the remaining two low-seasonality regimes: CrossFormer leads \textsc{high\_low\_high} by 38.4\% and \textsc{low\_low\_high} by 17.7\%.
This split suggests a difference in inductive bias: foundation models are strongest when the signal contains reusable global structure, while task-specific architectures are better at exploiting localized, cross-variate dependencies when periodic structure is weak.

\emph{Pathological \textsc{high\_low\_low} regime.}
The \textsc{high\_low\_low} regime (high trend, low seasonality, low forecastability) is a stress test: it is 56.7\% harder than the next-hardest regime and causes statistical baselines to fail catastrophically (MAE $>$ 1.0).
Even CrossFormer (MAE 0.600) performs 2.91$\times$ worse than on high-forecastability series, indicating a ceiling that current architectures cannot overcome without explicit non-stationary modeling; see \cref{app:tsf_regime_analysis} for detailed rankings.

\emph{TSF regime sensitivity.}
All regime-level conclusions depend on the threshold used to binarize trend, seasonality, and forecastability into \textsc{high}/\textsc{low} labels.
A sensitivity sweep over $\tau_{\mathrm{TSF}}\in[0.3,0.5]$ (\cref{app:threshold_sensitivity}) shows that the forecastability gap ($1.86$--$2.01\times$), CrossFormer's overall lead, and foundation models' regime advantage remain stable across thresholds.
This indicates that the conclusions are not an artifact of a particular discretization choice, but reflect persistent differences in model behavior across intrinsic series characteristics.

\vspace{-0.15cm}

\vspace{2pt}\noindent\textbf{Analysis V: Parameter efficiency of deep learning vs.\ foundation models.}
CrossFormer with 1\,M parameters and MAE 0.279 outperforms Chronos-2 with 100\,M parameters and MAE 0.314, using $100{\times}$ fewer parameters; see \cref{fig:efficiency_main}.
This is not an outlier: on average, deep learning models at 1.9\,M parameters match foundation models at 110\,M in accuracy (mean MAE 0.312 versus 0.319) despite a $58{\times}$ parameter gap.
Deep learning models are therefore far more parameter-efficient, challenging the assumption that large pre-trained models are uniformly superior for forecasting tasks.

\vspace{2pt}\noindent\textbf{Analysis VI: Ranking robustness: cross-metric and cross-benchmark.}
A natural concern is whether the observed rankings are artifacts of a particular choice of metric or dataset.

\begin{figure}[t]
    \centering
    \begin{subfigure}[t]{0.48\linewidth}
        \centering
        \includegraphics[width=\linewidth]{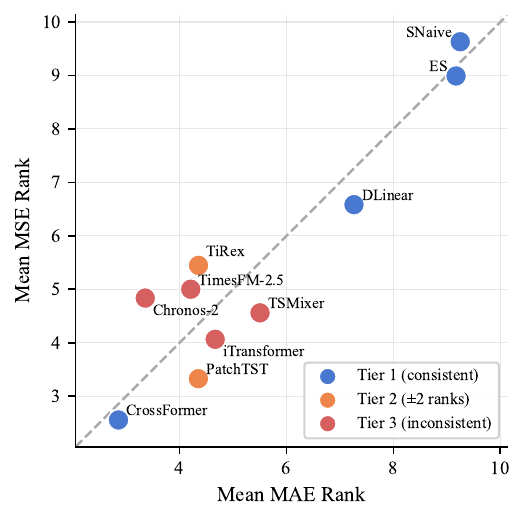}
        \caption{MAE vs.\ MSE rankings ($\rho{=}0.733$).}
        \label{fig:mse_mae_consistency}
    \end{subfigure}
\hfill
    \begin{subfigure}[t]{0.48\linewidth}
        \centering
        \includegraphics[width=\linewidth]{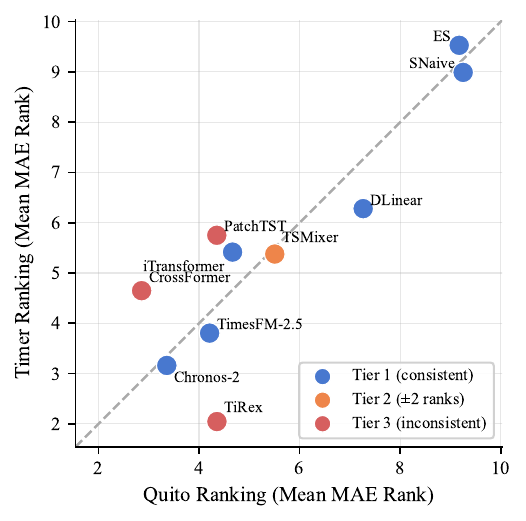}
        \caption{Quito vs.\ Timer MAE rankings ($\rho{=}0.865$).}
        \label{fig:quito_timer_consistency}
    \end{subfigure}
    \caption{Ranking robustness. (a)~Cross-metric: MAE and MSE produce highly consistent model orderings. (b)~Cross-benchmark: model rankings on \textsc{QuitoBench} vs.\ Timer are strongly correlated. Color indicates ordinal-shift tier. CrossFormer ranks first in all cases.}
    \label{fig:consistency}
\end{figure}

\emph{Cross-metric consistency.}
Model rankings under MAE and MSE are highly correlated: Spearman $\rho=0.733$ at the aggregate level, rising to a mean of 0.847 per configuration (\cref{fig:mse_mae_consistency}).
CrossFormer retains the top rank under both metrics.
A detailed breakdown is in \cref{app:mse_mae_consistency}.

\emph{Cross-benchmark consistency.}
\textsc{QuitoBench} vs.\ Timer rankings yield Spearman $\rho=0.865$ ($\rho=0.891$ for deep learning models), with CrossFormer ranking first in both benchmarks (\cref{fig:quito_timer_consistency}; see \cref{app:cross_benchmark_consistency} for a tier-based breakdown).
This agreement should not be read as redundancy: Timer validates the external relevance of the aggregate ranking, whereas \textsc{QuitoBench} contributes properties that Timer does not provide: contamination-resistant single provenance, near-uniform coverage of all TSF regimes, uniformly long series for long-context evaluation, and regime-level diagnosis of where different model families succeed or fail.
Despite Timer aggregating 11 public datasets from 5 domains, \textsc{Quito} achieves $1.24$--$1.57{\times}$ higher TSF diversity and $1.67{\times}$ higher regime entropy (\cref{app:tsf_provenance}), showing that single provenance does not limit statistical diversity.

To conclude, our findings are neither metric-dependent nor benchmark-specific.

\section{Conclusion}
\label{sec:conclusion}

We present \textsc{Quito} and \textsc{QuitoBench}, a billion-scale application-traffic corpus and regime-balanced benchmark that enable rigorous, contamination-free evaluation of time series forecasting models. Due to space limits, the discussion of related work is in \cref{app:related_work}.
All data, code, and pipelines are openly released.

\clearpage

\bibliographystyle{icml2020_url}
\bibliography{references}

@string{cvpr="Proceedings of the IEEE Conference on Computer Vision and Pattern Recognition (CVPR)"}

@string{eccv="Proceedings of the European Conference on Computer Vision (ECCV)"}

@string{aaai="Proceedings of the AAAI Conference on Artificial Intelligence"}

@string{icml="Proceedings of the International Conference on Machine Learning (ICML)"}

@string{neurips="Advances in Neural Information Processing Systems (NeurIPS)"}

@string{icassp="Proceedings of the International Conference on Acoustics, Speech and Signal Processing (ICASSP)"}

@string{iclr="Proceedings of the International Conference on Learning Representations (ICLR)"}

@string{kdd="Proceedings of the ACM SIGKDD International Conference on Knowledge Discovery and Data Mining"}

@string{sosp="Proceedings of the ACM Symposium on Operating Systems Principles (SOSP)"}

@string{eurosys="Proceedings of the European Conference on Computer Systems (EuroSys)"}

@article{arima,
  title     = {Distribution of Residual Autocorrelations in Autoregressive-Integrated Moving Average Time Series Models},
  author    = {Box, G. E. P. and Pierce, David A.},
  journal   = {Journal of the American Statistical Association},
  volume    = {65},
  number    = {332},
  pages     = {1509--1526},
  year      = {1970},
  publisher = {[American Statistical Association, Taylor \& Francis, Ltd.]},
  url       = {http://www.jstor.org/stable/2284333}
}

@article{cleveland1990stl,
  title     = {{STL}: A Seasonal-Trend Decomposition Procedure Based on Loess},
  author    = {Cleveland, Robert B. and Cleveland, William S. and McRae, Jean E. and Terpenning, Irma},
  journal   = {Journal of Official Statistics},
  volume    = {6},
  number    = {1},
  pages     = {3--73},
  year      = {1990},
  publisher = {Statistics Sweden}
}

@article{wang2006characteristic,
  title     = {Characteristic-Based Clustering for Time Series Data},
  author    = {Wang, Xiaozhe and Smith, Kate and Hyndman, Rob},
  journal   = {Data Mining and Knowledge Discovery},
  volume    = {13},
  number    = {3},
  pages     = {335--364},
  year      = {2006},
  publisher = {Springer},
  doi       = {10.1007/s10618-005-0039-x},
  url       = {https://doi.org/10.1007/s10618-005-0039-x}
}

@article{welch1967use,
  title     = {The Use of Fast {Fourier} Transform for the Estimation of Power Spectra:
               A Method Based on Time Averaging Over Short, Modified Periodograms},
  author    = {Welch, Peter D.},
  journal   = {IEEE Transactions on Audio and Electroacoustics},
  volume    = {15},
  number    = {2},
  pages     = {70--73},
  year      = {1967},
  publisher = {IEEE},
  doi       = {10.1109/TAU.1967.1161901},
  url       = {https://doi.org/10.1109/TAU.1967.1161901}
}

@inproceedings{flunkert2017deepar,
  title     = {DeepAR: Probabilistic Forecasting with Autoregressive Recurrent Networks},
  author    = {Flunkert, Valentin and Salinas, David and Gasthaus, Jan},
  booktitle = iclr,
  year      = {2017},
  url       = {https://arxiv.org/abs/1704.04110}
}

@inproceedings{zhou2021informer,
  title     = {Informer: Beyond Efficient Transformer for Long Sequence Time-Series Forecasting},
  author    = {Zhou, Haoyi and Zhang, Shanghang and Peng, Jieqi and Zhang, Shuai and Li, Jianxin and Xiong, Hui and Zhang, Wancai},
  booktitle = aaai,
  pages     = {11106--11115},
  year      = {2021},
  url       = {https://arxiv.org/abs/2012.07436}
}

@inproceedings{wu2021autoformer,
  title     = {Autoformer: Decomposition Transformers with Auto-Correlation for Long-Term Series Forecasting},
  author    = {Wu, Haixu and Xu, Jiehui and Wang, Jianmin and Long, Mingsheng},
  booktitle = neurips,
  year      = {2021},
  url       = {https://arxiv.org/abs/2106.13008}
}

@inproceedings{liu2022pyraformer,
  title     = {Pyraformer: Low-Complexity Pyramidal Attention for Long-Range Time Series Modeling and Forecasting},
  author    = {Liu, Shizhan and Yu, Hang and Liao, Cong and Li, Jianguo and Lin, Weiyao and Liu, Alex X and Dustdar, Schahram},
  booktitle = iclr,
  year      = {2022},
  url       = {https://openreview.net/forum?id=Lk2XzEjR3L}
}

@inproceedings{nie2023patchtst,
  title     = {A Time Series is Worth 64 Words: Long-term Forecasting with Transformers},
  author    = {Nie, Yuqi and Nguyen, Nam H. and Sinthong, Phanwadee and Kalagnanam, Jayant},
  booktitle = iclr,
  year      = {2023},
  url       = {https://openreview.net/forum?id=Jbdc0vTOcol}
}

@inproceedings{zhang2023crossformer,
  title     = {Crossformer: Transformer Utilizing Cross-Dimension Dependency for Multivariate Time Series Forecasting},
  author    = {Zhang, Yunhao and Yan, Junchi},
  booktitle = iclr,
  year      = {2023},
  url       = {https://openreview.net/forum?id=vSVLM2j9eie}
}

@inproceedings{zeng2023dlinear,
  title     = {Are Transformers Effective for Time Series Forecasting?},
  author    = {Zeng, Ailing and Chen, Ming and Zhang, Lei and Xu, Qiang},
  booktitle = aaai,
  year      = {2023},
  url       = {https://arxiv.org/abs/2205.13504}
}

@article{chen2023tsmixer,
  title   = {{TSMixer}: An All-MLP Architecture for Time Series Forecasting},
  author  = {Chen, Si-An and Li, Chun-Liang and Yoder, Nate and Arik, Sercan O. and Pfister, Tomas},
  journal = {arXiv preprint arXiv:2303.06053},
  year    = {2023},
  url     = {https://arxiv.org/abs/2303.06053}
}

@inproceedings{liu2024itransformer,
  title     = {iTransformer: Inverted Transformers Are Effective for Time Series Forecasting},
  author    = {Liu, Yong and Hu, Tengge and Zhang, Haoran and Wu, Haixu and Wang, Shiyu and Ma, Lintao and Long, Mingsheng},
  booktitle = iclr,
  year      = {2024},
  url       = {https://openreview.net/forum?id=JePfAI8fah}
}

@article{das2023timesfm,
  title   = {{TimesFM}: A Decoder-only Foundation Model for Time Series Forecasting},
  author  = {Das, Abhimanyu and Kong, Weiqing and Leach, Andrew and Sen, Rajat and Kalagnanam, Jayant},
  journal = {arXiv preprint arXiv:2310.10688},
  year    = {2023},
  url     = {https://arxiv.org/abs/2310.10688}
}

@article{ansari2024chronos,
  title   = {Chronos: Learning the Language of Time Series},
  author  = {Ansari, Aman and Stella, Lorenzo and Turcotte, Mathieu and Salinas, David and Schmid, Philipp and Bohlke-Schneider, Michael and Mercier, David and Candela, Joaquin},
  journal = {arXiv preprint arXiv:2403.07815},
  year    = {2024},
  url     = {https://arxiv.org/abs/2403.07815}
}

@article{woo2024moirai,
  title   = {Moirai: A Large-scale Universal Time Series Forecasting Model},
  author  = {Woo, Gerald and Liu, Cheng and Lim, Brian and Bohlke-Schneider, Michael and Salinas, David},
  journal = {arXiv preprint arXiv:2402.02592},
  year    = {2024},
  url     = {https://arxiv.org/abs/2402.02592}
}

@inproceedings{liu2024timer,
  title     = {Timer: Generative Pre-trained Transformers Are Large Time Series Models},
  author    = {Liu, Yong and Zhang, Yuxuan and Li, Zhe and Chen, Wei and Xiong, Hui},
  booktitle = icml,
  year      = {2024},
  url       = {https://arxiv.org/abs/2402.02368}
}

@inproceedings{25tirex,
  title     = {{TiRex}: Zero-Shot Forecasting Across Long and Short Horizons with Enhanced In-Context Learning},
  author    = {Auer, Andreas and Podest, Patrick and Klotz, Daniel and B{\"o}ck, Sebastian and Klambauer, G{\"u}nter and Hochreiter, Sepp},
  booktitle = neurips,
  year      = {2025},
  url       = {https://arxiv.org/abs/2505.23719}
}

@article{jin2023large,
  title   = {Large Models for Time Series and Spatio-Temporal Data: A Survey and Outlook},
  author  = {Jin, Ming and Wen, Qingsong and Liang, Yuxuan and Zhang, Chaoli and Xue, Siqiao and Wang, Xue and Zhang, James and Wang, Yi and Chen, Haifeng and Li, Xiaoli and others},
  journal = {arXiv preprint arXiv:2310.10196},
  year    = {2023},
  url     = {https://arxiv.org/abs/2310.10196}
}

@article{aksu2024gifteval,
  title   = {{GIFT-Eval}: A Benchmark For General Time Series Forecasting Model Evaluation},
  author  = {Aksu, Taha and Woo, Gerald and Liu, Juncheng and Liu, Xu and Liu, Chenghao and Savarese, Silvio and Xiong, Caiming and Sahoo, Doyen},
  journal = {arXiv preprint arXiv:2410.10393},
  year    = {2024},
  url     = {https://arxiv.org/abs/2410.10393}
}

@article{meyer2025timeseries,
  title   = {Rethinking Evaluation in the Era of Time Series Foundation Models: {(Un)known} Information Leakage Challenges},
  author  = {Meyer, Marcel and Kaltenpoth, Sascha and Zalipski, Kevin and M{\"u}ller, Oliver},
  journal = {arXiv preprint arXiv:2510.13654},
  year    = {2025},
  url     = {https://arxiv.org/abs/2510.13654}
}

@article{freiesleben2025benchmark,
  title   = {The Benchmarking Epistemology: Construct Validity for Evaluating Machine Learning Models},
  author  = {Freiesleben, Timo and Zezulka, Sebastian},
  journal = {arXiv preprint arXiv:2510.23191},
  year    = {2025},
  url     = {https://arxiv.org/abs/2510.23191}
}

@inproceedings{deng2009imagenet,
  title     = {{ImageNet}: A Large-Scale Hierarchical Image Database},
  author    = {Deng, Jia and Dong, Wei and Socher, Richard and Li, Li-Jia and Li, Kai and Fei-Fei, Li},
  booktitle = cvpr,
  year      = {2009},
  url       = {https://doi.org/10.1109/CVPR.2009.5206848}
}

@inproceedings{lin2014coco,
  title     = {Microsoft {COCO}: Common Objects in Context},
  author    = {Lin, Tsung-Yi and Maire, Michael and Belongie, Serge and others},
  booktitle = eccv,
  year      = {2014},
  url       = {https://arxiv.org/abs/1405.0312}
}

@inproceedings{wang2018glue,
  title     = {{GLUE}: A Multi-Task Benchmark and Analysis Platform for Natural Language Understanding},
  author    = {Wang, Alex and Singh, A. and Michael, J. and Hill, F. and Levy, O. and Bowman, S.},
  booktitle = iclr,
  year      = {2019},
  url       = {https://arxiv.org/abs/1804.07461}
}

@inproceedings{panayotov2015librispeech,
  title     = {Librispeech: An {ASR} Corpus Based on Public Domain Audio Books},
  author    = {Panayotov, Vassil and Chen, Guoguo and Povey, Daniel and Khudanpur, Sanjeev},
  booktitle = icassp,
  year      = {2015},
  url       = {https://doi.org/10.1109/ICASSP.2015.7178964}
}

@inproceedings{cortez2017resource,
  title     = {Resource Central: Understanding and Predicting Workloads for Improved Resource Management in Large Cloud Platforms},
  author    = {Cortez, Eli and Bonde, Anand and Muzio, Alexandre and Russinovich, Mark and Fontoura, Marcus and Bianchini, Ricardo},
  booktitle = sosp,
  pages     = {153--167},
  year      = {2017},
  url       = {https://doi.org/10.1145/3132747.3132772},
  doi       = {10.1145/3132747.3132772}
}

@inproceedings{tirmazi2020borg,
  title     = {Borg: The Next Generation},
  author    = {Tirmazi, Muhammad and Barker, Adam and Deng, Nan and Haque, Md E. and Qin, Zhijing Genie and Hand, Steven and Harchol-Balter, Mor and Wilkes, John},
  booktitle = eurosys,
  pages     = {1--14},
  year      = {2020},
  url       = {https://doi.org/10.1145/3342195.3387517},
  doi       = {10.1145/3342195.3387517}
}

@inproceedings{luo2021characterizing,
  title     = {Characterizing Microservice Dependency and Performance: {Alibaba} Trace Analysis},
  author    = {Luo, Shutian and Xu, Huanle and Lu, Chengzhi and Ye, Kejiang and Xu, Guoyao and Zhang, Liping and Ding, Yu and He, Jian and Xu, Chengzhong},
  booktitle = {Proceedings of the ACM Symposium on Cloud Computing (SoCC)},
  pages     = {412--426},
  year      = {2021},
  url       = {https://doi.org/10.1145/3472883.3487003},
  doi       = {10.1145/3472883.3487003}
}

@techreport{chen2001freeway,
  title       = {Freeway Performance Measurement System: Mining Loop Detector Data},
  author      = {Chen, Chao and Petty, Karl and Skabardonis, Alexander and Varaiya, Pravin and Jia, Zhanfeng},
  institution = {University of California, Berkeley},
  year        = {2001}
}

@inproceedings{cao2023fints,
  title     = {Large Scale Financial Time Series Forecasting with Multi-faceted Model},
  author    = {Cao, Defu and Zheng, Yixiang and Hassanzadeh, Parisa and Lamba, Simran and Liu, Xiaomo and Liu, Yan},
  booktitle = {Proceedings of the Fourth ACM International Conference on AI in Finance (ICAIF)},
  pages     = {472--480},
  year      = {2023},
  doi       = {10.1145/3604237.3626868},
  url       = {https://doi.org/10.1145/3604237.3626868}
}

@inproceedings{paszke2019pytorch,
  title     = {{PyTorch}: An Imperative Style, High-Performance Deep Learning Library},
  author    = {Paszke, Adam and Gross, Sam and Massa, Francisco and Lerer, Adam and Bradbury, James and Chanan, Gregory and Killeen, Trevor and Lin, Zeming and Gimelshein, Natalia and Antiga, Luca and others},
  booktitle = neurips,
  volume    = {32},
  year      = {2019},
  url       = {https://arxiv.org/abs/1912.01703}
}

@inproceedings{wen2022robust,
  title     = {Robust Time Series Analysis and Applications: An Industrial Perspective},
  author    = {Wen, Qingsong and Yang, Linxiao and Zhou, Tian and Sun, Liang},
  booktitle = kdd,
  pages     = {4836--4837},
  year      = {2022},
  url       = {https://doi.org/10.1145/3534678.3542599}
}

@inproceedings{xue_meta_2022,
  title     = {A Meta Reinforcement Learning Approach for Predictive Autoscaling in the Cloud},
  author    = {Xue, Siqiao and Qu, Chao and Shi, Xiaoming and Liao, Cong and Zhu, Shiyi and Tan, Xiaoyu and Ma, Lintao and Wang, Shiyu and Wang, Shijun and Hu, Yun and Lei, Lei and Zheng, Yangfei and Li, Jianguo and Zhang, James},
  booktitle = kdd,
  pages     = {4290--4299},
  publisher = {ACM},
  year      = {2022},
  doi       = {10.1145/3534678.3539063},
  url       = {https://doi.org/10.1145/3534678.3539063}
}

@article{xue2023weaverbird,
  title   = {{WeaverBird}: Empowering Financial Decision-Making with Large Language Model, Knowledge Base, and Search Engine},
  author  = {Xue, Siqiao and Zhou, Fan and Xu, Yi and Jin, Ming and Wen, Qingsong and Hao, Hongyan and Dai, Qingyang and Jiang, Caigao and Zhao, Hongyu and Xie, Shuo and He, Jianshan and Zhang, James and Mei, Hongyuan},
  journal = {arXiv preprint arXiv:2308.05361},
  year    = {2023},
  url     = {https://arxiv.org/abs/2308.05361}
}

@inproceedings{xue2024easytpp,
  title     = {{EasyTPP}: Towards Open Benchmarking Temporal Point Processes},
  author    = {Xue, Siqiao and Shi, Xiaoming and Chu, Zhixuan and Wang, Yan and Hao, Hongyan and Zhou, Fan and Jiang, Caigao and Pan, Chen and Zhang, James Y. and Wen, Qingsong and Zhou, Jun and Mei, Hongyuan},
  booktitle = iclr,
  year      = {2024},
  url       = {https://arxiv.org/abs/2307.08097}
}

@article{xue2024famma,
  title   = {{FAMMA}: A Benchmark for Financial Domain Multilingual Multimodal Question Answering},
  author  = {Xue, Siqiao and Li, Xiaojing and Zhou, Fan and Dai, Qingyang and Chu, Zhixuan and Mei, Hongyuan},
  journal = {arXiv preprint arXiv:2410.04526},
  year    = {2024},
  url     = {https://arxiv.org/abs/2410.04526}
}

@article{gao2026lookbench,
  title   = {{LookBench}: A Live and Holistic Open Benchmark for Fashion Image Retrieval},
  author  = {Gensmo.ai and Gao, Chao and Xue, Siqiao and Peng, Yimin and Fu, Jiwen and Gu, Tingyi and Li, Shanshan and Zhou, Fan},
  journal = {arXiv preprint arXiv:2601.14706},
  year    = {2026},
  url     = {https://arxiv.org/abs/2601.14706}
}

@article{ansari2025chronos,
  title={Chronos-2: From univariate to universal forecasting},
  author={Ansari, Abdul Fatir and Shchur, Oleksandr and K{\"u}ken, Jaris and Auer, Andreas and Han, Boran and Mercado, Pedro and Rangapuram, Syama Sundar and Shen, Huibin and Stella, Lorenzo and Zhang, Xiyuan and others},
  journal={arXiv preprint arXiv:2510.15821},
  year={2025}
}

@inproceedings{pan2023deep,
      title={Deep Optimal Timing Strategies for Time Series}, 
      author={Chen Pan and Fan Zhou and Xuanwei Hu and Xinxin Zhu and Wenxin Ning and Zi Zhuang and Siqiao Xue and James Zhang and Yunhua Hu},
      year={2023},
      booktitle = {ICDM}
}

@inproceedings{Zhou_2024,
   title={GMP-AR: Granularity Message Passing and Adaptive Reconciliation for Temporal Hierarchy Forecasting},
   author={Zhou, Fan and Pan, Chen and Ma, Lintao and Liu, Yu and Xue, Siqiao and Zhang, James and Zhou, Jun and Mei, Hongyuan and Lin, Weitao and Zhuang, Zi and Ning, Wenxin and Hu, Yunhua},
   booktitle=AAAI,
   year={2024}}

\clearpage
\appendix
\appendixpage

\crefalias{section}{appendix}
\crefalias{subsection}{appendix}

\section{Extended Related Work}
\label{app:related_work}

\vspace{2pt}\noindent\textbf{Time series forecasting models.}
Time series forecasting has been extensively studied, with methods broadly categorized into statistical models, deep learning models, and, more recently, foundation models.
Classical statistical methods, such as ARIMA~\citep{arima}, model future values based on explicit assumptions about temporal structure and data distributions.
Deep learning models~\citep{pan2023deep,Zhou_2024} relax these assumptions and capture nonlinear dynamics, with representative approaches including DeepAR~\citep{flunkert2017deepar} and DLinear~\citep{zeng2023dlinear}, as well as transformer-based models such as PatchTST~\citep{nie2023patchtst} and Crossformer~\citep{zhang2023crossformer}.
More recently, foundation models for time series forecasting~\citep{jin2023large} have emerged, inspired by the success of large-scale pretraining across multiple domains.
Representative examples include TimesFM-2.5~\citep{das2023TimesFM}, Moirai~\citep{woo2024moirai}, and Chronos-2~\citep{ansari2025Chronos}.
Despite their promising performance, the development and evaluation of these models are constrained by the limited availability of high-quality, large-scale, and standardized benchmark datasets.

\vspace{2pt}\noindent\textbf{Time series forecasting benchmarks.}
Several large-scale benchmarks have been proposed for time series forecasting evaluation, including LOTSA~\citep{woo2024moirai}, Timer~\citep{liu2024timer}, and GIFT-Eval~\citep{aksu2024gifteval}.
While they aggregate diverse public time series spanning multiple domains, their reliance on heterogeneous open sources introduces evaluation risks: \citet{meyer2025timeseries} demonstrate that both direct train-test overlaps and indirect leakage from temporally correlated series systematically inflate performance estimates, and \citet{freiesleben2025benchmark} catalogue further methodological pitfalls in current benchmarking practice.
Our benchmark addresses these concerns by construction: it is built from a single proprietary industrial source with no overlap with any public pre-training corpus, and it is carefully balanced across intrinsic data characteristics, enabling leakage-free, rigorous evaluation across diverse forecasting scenarios.

\section{Broader Impact, Limitations, and Future Directions}
\label{app:impact}

\vspace{2pt}\noindent\textbf{Broader Impact.}
By releasing industrial-scale, multi-vertical application traffic data and a balanced benchmark under an open license (CC-BY 4.0), we enable researchers without access to proprietary data to develop and evaluate models on realistic, diverse workloads. For practitioners, our TSF-regime framework enables proactive identification of high-risk forecasting scenarios: the \textsc{high\_low\_low} regime, while representing only 12\% of our benchmark, shows 3$\times$ higher error rates that warrant specialized modeling approaches.

\vspace{2pt}\noindent\textbf{Limitations.}
All data originates from a single organisation (Alipay).
Although the corpus spans nine functionally distinct business verticals and achieves higher TSF diversity than multi-provenance benchmarks (\cref{app:tsf_provenance}), certain workload patterns specific to other cloud platforms (\eg batch-heavy HPC clusters, CDN edge caches) may be under-represented.
Our benchmark focuses on point forecasting of application traffic; results may differ for probabilistic forecasting, anomaly detection, or other time series tasks.
The TSF regime threshold (0.4) was chosen for balanced coverage but may not be optimal for all use cases (see \cref{app:threshold_sensitivity} for a sensitivity analysis).
Foundation model access requires sufficient computational resources for inference.

\vspace{2pt}\noindent\textbf{Future Directions.}
We plan to extend \textsc{QuitoBench} to additional prediction tasks (probabilistic forecasting, anomaly detection), incorporate new foundation models as they emerge, and develop forecastability-aware model selection guidance tools.
As large language models are increasingly applied to financial and operational decision-making~\citep{xue2023weaverbird}, evaluating their integration with time series forecasting pipelines is another promising direction.
A direct TSF comparison with public CloudOps traces (\eg Azure~\citep{cortez2017resource}, Google Borg~\citep{tirmazi2020borg}) would further validate that regime-level findings are provenance-agnostic; we leave this as future work.

\section{Analysis on GIFT-Eval}
\label{app:gift_eval}

\begin{minipage}{0.48\textwidth}
    \centering
    \includegraphics[width=\linewidth]{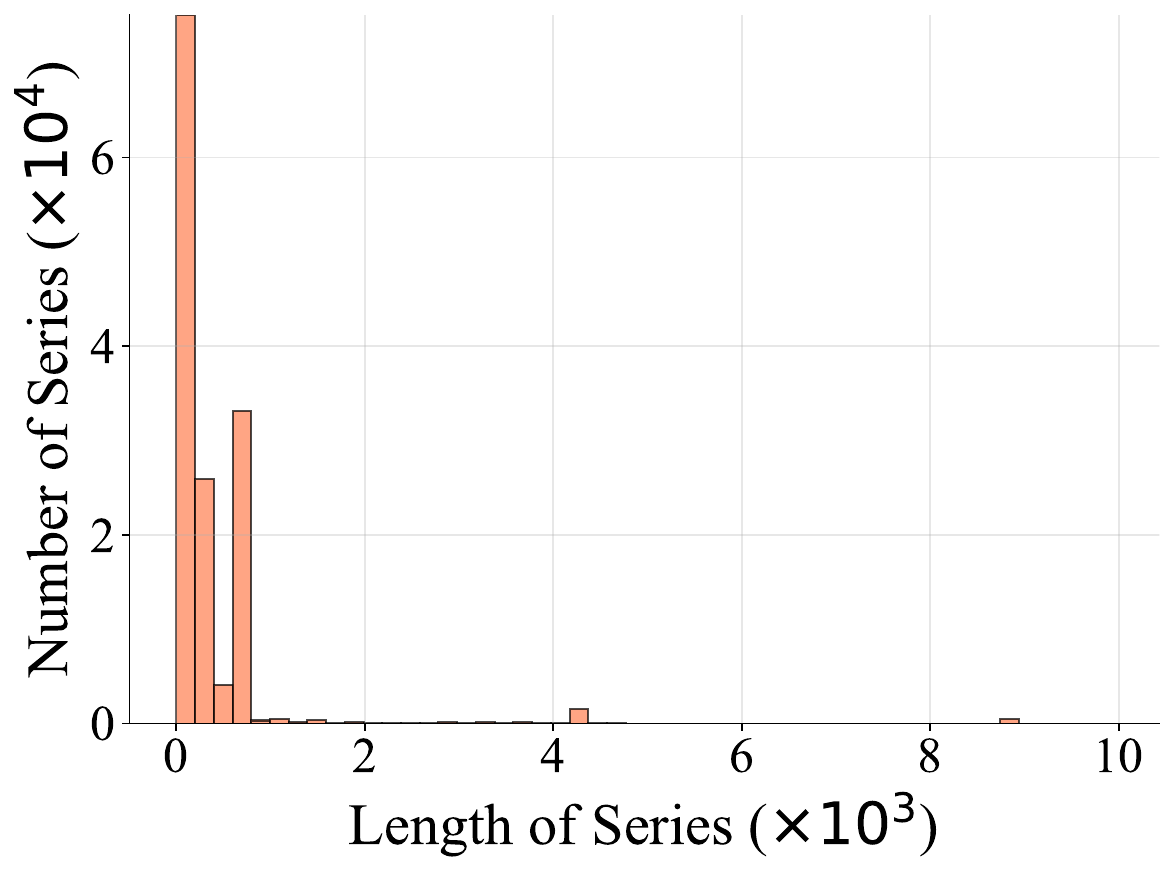}%
    \captionof{figure}{Distribution of series lengths in GIFT-Eval benchmark. The histogram shows a highly unbalanced distribution with most series having lengths below 1,000, while a small fraction extends beyond 10,000.}
    \label{fig:gifteval_histogram}
\end{minipage}
\hfill
\begin{minipage}{0.48\textwidth}
    \centering
    \includegraphics[width=\linewidth]{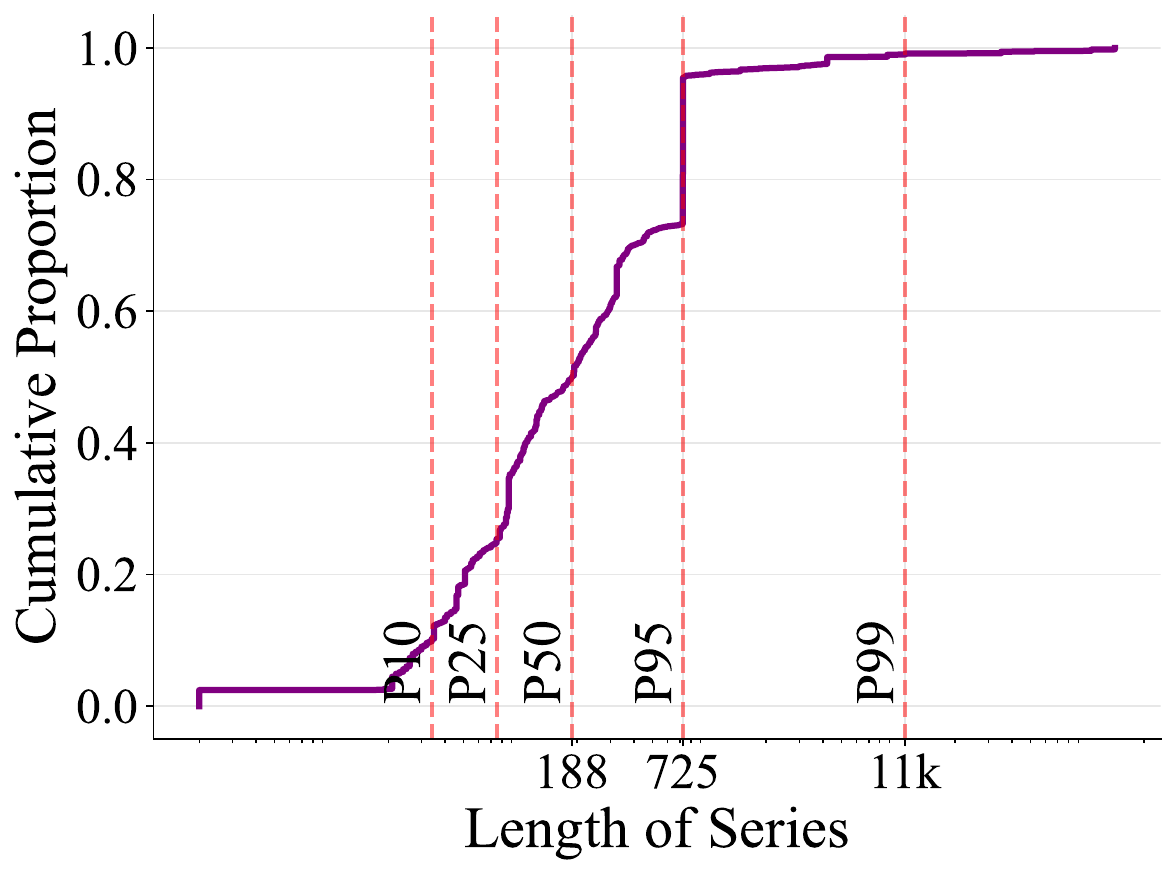}%
    \captionof{figure}{Cumulative distribution of series lengths in GIFT-Eval benchmark, with percentile markers indicating that 50\% of series are below 200, 95\% are below 1,000, while 99\% are below 10,000.}
    \label{fig:gifteval_cumul}
\end{minipage}

\section{Dataset Details}
\label{app:dataset}

\subsection{Dataset Statistics and Format}
\label{app:dataset_stats}

We release two complementary datasets: \textsc{Quito} (the full source corpus) and \textsc{QuitoBench} (the evaluation benchmark curated from \textsc{Quito}), both distributed as Apache Parquet files.

\vspace{2pt}\noindent\textbf{Data format.}
Both datasets use a long (tidy) row format: each row corresponds to one timestamp of one series, with all series stored in a single file.
The schema is as follows:

\begin{center}
\small
\begin{sc}
\begin{tabular}{@{}ll@{}}
\toprule
Column & Description \\
\midrule
item\_id       & unique series identifier \\
date\_time     & UTC timestamp at the series resolution \\
ind\_1\,--\,ind\_5 & five anonymized variate channels (float; NaN for missing) \\
cluster        & \textsc{tsf} regime label (8-class; benchmark files only) \\
\bottomrule
\end{tabular}
\end{sc}
\captionof{table}{Parquet schema shared by \textsc{Quito} and \textsc{QuitoBench}. The cluster column is present only in the benchmark (evaluation) files.}
\label{tab:schema}
\end{center}

\noindent To reconstruct a single multivariate series, filter rows by \texttt{item\_id} and sort by \texttt{date\_time}; no silent imputation is applied.

\vspace{2pt}\noindent\textbf{\textsc{Quito} full corpus.}
\textsc{Quito} is released as two Parquet files (open\_hour\_data\_pretrain and open\_min\_data\_pretrain), whose key statistics are summarized in \cref{tab:stat_quito}.
\textsc{Quito-Min} contains 22,522 series at 10-minute resolution (5,904 time points each; 0.7\,B total tokens), spanning 2023-07-10 to 2023-08-19.
\textsc{Quito-Hour} contains 12,544 multivariate series at 1-hour resolution (15,356 time points each; 1.0\,B total tokens), spanning 2021-11-18 to 2023-08-19.
The two subsets are drawn from disjoint pools of applications with no overlap in item identifiers.
The differing start dates reflect the production system's tiered retention policy: hourly aggregates are archived long-term, while high-frequency 10-minute telemetry is only retained for a shorter rolling window.

\vspace{2pt}\noindent\textbf{\textsc{QuitoBench} evaluation files.}
\textsc{QuitoBench} is released as two Parquet files (open\_hour\_data\_test and open\_min\_data\_test).
These files contain the complete series for the 1,290 sampled benchmark items, covering the period from each subset's start time through 2023-08-19, the shared end date of the \textsc{Quito} corpus.
\textsc{QuitoBench-Min} contains 773 series (5,904 time points each, spanning 2023-07-10 to 2023-08-19) and \textsc{QuitoBench-Hour} contains 517 series (15,356 time points each, spanning 2021-11-18 to 2023-08-19).
Users can reconstruct the train/valid/test splits using the global cutoff defined in \cref{app:quitobench_testset}.
The schema is identical to the full corpus; the cluster column carries the \textsc{tsf} regime label used for stratified evaluation.
Key statistics are summarized in \cref{tab:stat_quito_bench}.

\subsection{Series Formulation Details}
\label{app:quito_preprocessing}

\vspace{2pt}\noindent\textbf{Minute- and hour-level collections.}
\textsc{Quito-Min} and \textsc{Quito-Hour} are constructed from two disjoint pools of raw workload series (\ie there is no intersection of applications between the minute and hour releases). 
Both subsets share the same end date (2023-08-19), which also serves as the global cutoff for \textsc{QuitoBench}, ensuring leakage-free train/valid/test splits across both granularities.

\vspace{2pt}\noindent\textbf{Time aggregation and resolution.}
The underlying telemetry is collected at a 1\,s sampling frequency and continuously aggregated by the production pipeline into coarser resolutions for storage.
For our release, the target resolution is $r\!\in\!\{60,\,3600\}$ seconds (10-minute / 1-hour): we partition time into fixed, non-overlapping bins anchored at the chosen start time and aggregate each variate by max pooling within each bin.
Concretely, for an item and variate, the aggregated value at bin $t$ is
\[
x^{(r)}_{t} \;=\; \max_{s \in [t_0 + tr,\; t_0 + (t+1)r)} x_s,
\]
where $t_0$ is the start timestamp of the selected window.
Max-based aggregation preserves bursty workload peaks while producing standardized multivariate time series at minute and hour resolutions for downstream modeling and evaluation.

\vspace{2pt}\noindent\textbf{Time-series format and anonymized variates.}
Each workload trace is assigned a unique identifier \texttt{item\_id}. 
At each time step, the workload is represented as a 5-dimensional vector whose channels correspond to different traffic subtypes; due to commercial sensitivity, we mask subtype semantics and denote the five variates as $\texttt{index}_i$ for $i=1,\dots,5$.
Accordingly, each record can be viewed as (\texttt{item\_id}, timestamp, $\texttt{index}_1,\dots,\texttt{index}_5$), with missing values represented explicitly when applicable.

\subsection{Business Vertical Distribution}
\label{app:vertical_dist}

\begin{figure}[h]
    \centering
    \includegraphics[width=\linewidth]{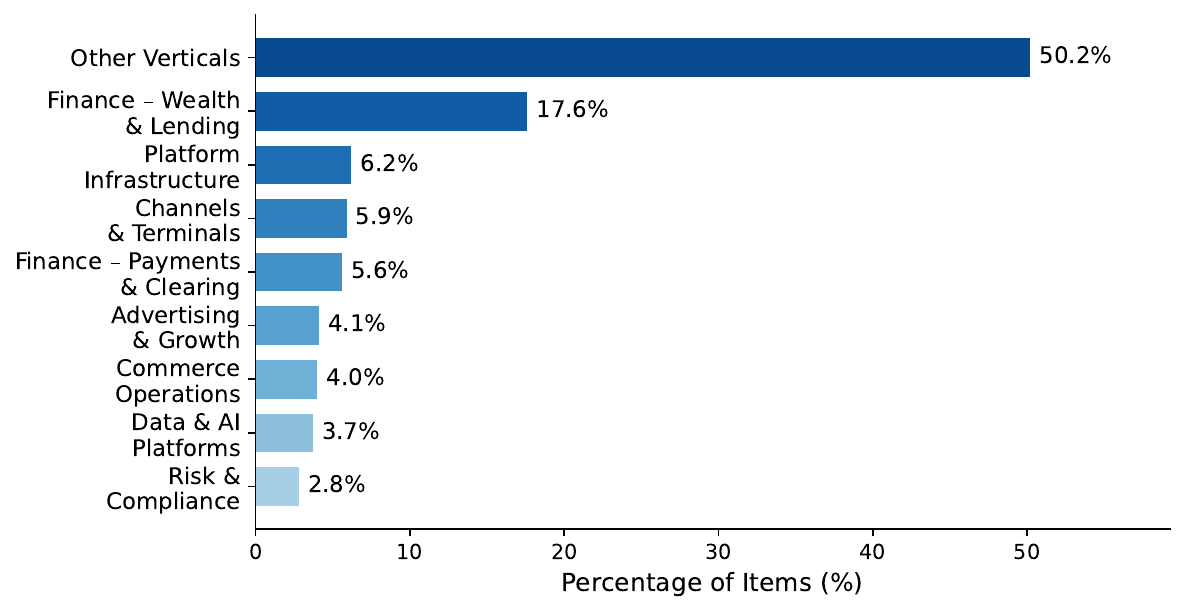}
    \caption{Distribution of \textsc{Quito} items across nine coarse-grained business verticals. Each item corresponds to the traffic workload of a distinct application service, spanning finance, commerce, advertising, infrastructure, and more, reflecting the diversity of real-world application traffic.}
    \label{fig:vertical_dist}
\end{figure}

Each \texttt{item\_id} in \textsc{Quito} corresponds to the traffic workload of a distinct application service running on Alipay's production platform.
Alipay is one of the world's largest digital payment and lifestyle platforms; its backend infrastructure hosts hundreds of thousands of microservices spanning virtually every facet of a modern digital economy.
\Cref{fig:vertical_dist} shows the distribution of items across nine coarse-grained business verticals\footnote{a. The data set does not contain any Personal Identifiable Information (PII);
b. The data set is only used for academic research, it does not represent any real business situation.}:
\begin{itemize}[nosep,leftmargin=*]
    \item Finance: wealth management, lending, payments, and clearing services ($\sim$23\% of the corpus), exhibiting strong weekly and monthly seasonality driven by periodic cycles and promotional campaigns.
    \item Commerce operations: services supporting merchant and order workflows, with pronounced intra-day and holiday-driven traffic spikes.
    \item Advertising and growth: recommendation and user-acquisition pipelines characterized by highly bursty, campaign-driven workloads.
    \item Platform infrastructure ($\sim$6\%): middleware and DevOps services with more stationary patterns and occasional step-changes during deployments or scaling events.
    \item Data and AI platforms: batch and streaming analytics producing a mix of periodic peaks and steady baselines.
    \item Risk and compliance: services with real-time latency requirements and event-driven traffic patterns.
    \item Channels and terminals: app backends and device interfaces reflecting end-user activity patterns.
    \item Other verticals: a diverse tail including public services, lifestyle, logistics, and content platforms.
\end{itemize}
This breadth is central to our claim that \textsc{Quito}, despite being single-provenance, is \emph{not} single-domain.
The traffic patterns of a compliance service, an advertising pipeline, and a financial gateway differ as fundamentally as electricity, weather, and transportation series in traditional benchmarks, yet they coexist within a single, contamination-free data environment.
The diversity is also reflected in the TSF regime distribution: \textsc{Quito} series populate all eight trend$\times$seasonality$\times$forecastability regimes with near-uniform coverage (\cref{fig:quito_grid}), confirming that the underlying dynamics are heterogeneous rather than clustered in a narrow region of the forecasting difficulty space.
Quantitatively, \textsc{Quito} achieves $1.24$--$1.57{\times}$ higher TSF standard deviation and $1.67{\times}$ higher regime entropy than the Timer benchmark, which aggregates 11 public datasets from 5 distinct domains (\cref{app:tsf_provenance}).
This demonstrates that single-provenance data from a large-scale, multi-vertical platform can be \emph{more} diverse in TSF space than conventional multi-provenance benchmarks.

\subsection{Standardization Details}
\label{app:quito_dedup}

\vspace{2pt}\noindent\textbf{Deduplication.}
Large-scale production telemetry may contain redundant copies of the same underlying workload trace due to mirrored exporters, repeated exports, or multiple ingestion pipelines.
We therefore apply a two-stage deduplication procedure after standardization.
First, we remove \emph{exact} duplicates by computing a deterministic hash of each standardized multivariate series (concatenating the five channels after alignment) and dropping repeated hashes.
Second, to identify \emph{near}-duplicates, we compute compact fingerprints for each series by (i) z-normalizing each channel, (ii) downsampling to a fixed length, and (iii) quantizing the resulting values.
We use these fingerprints to retrieve candidate near-duplicate pairs efficiently, and then verify each candidate by measuring similarity over the overlapping time interval.
Two series are considered duplicates if their channel-wise Pearson correlations exceed 0.99 (with an optional $\pm 1$-bin lag to tolerate minor timestamp shifts); among duplicates we keep a single canonical series and record provenance links to the removed copies.

\subsection{TSF Diagnostic Details}
\label{app:tsf_diagnostics}

We compute three scalar diagnostics for each time series to characterize its dynamic regime: \emph{trend strength} ($T$), \emph{seasonality strength} ($S$), and \emph{forecastability} ($F$).
Each diagnostic is bounded in $[0,1]$, where larger values indicate stronger presence of the respective property.

\vspace{2pt}\noindent\textbf{Trend and Seasonality Strength via STL.}
We decompose each univariate series $\{x_t\}$ using STL (Seasonal-Trend decomposition via LOESS)~\citep{cleveland1990stl} with robust fitting, which produces three additive components:
\[
x_t = \tau_t + s_t + r_t,
\]
where $\tau_t$ is the trend, $s_t$ the seasonal component, and $r_t$ the residual.
Following~\citet{wang2006characteristic}, we define:
\[
S = \max\!\left(0,\;1 - \frac{\mathrm{Var}(r)}{\mathrm{Var}(s + r)}\right), \qquad
T = \max\!\left(0,\;1 - \frac{\mathrm{Var}(r)}{\mathrm{Var}(\tau + r)}\right).
\]
A value near~1 means the trend (or seasonal) component dominates over the residual; a value near~0 means the component is negligible relative to noise.
The seasonal period $p$ is set according to the series resolution: $p = 144$ for \textsc{Quito-Min} (one daily cycle per $144\times 10$-minute observations) and $p = 24$ for \textsc{Quito-Hour} (one daily cycle per 24 hourly observations).

\vspace{2pt}\noindent\textbf{Forecastability via Spectral Entropy.}
We measure forecastability as the complement of normalized spectral entropy estimated via Welch's method~\citep{welch1967use}.
Concretely, for a mean-subtracted series $\tilde{x}_t$, we compute the power spectral density $P_k$ using Welch's overlapping-segment periodogram with a Hann window and segment length $\min(|x|, 1024)$:
\[
H = -\sum_k \hat{p}_k \log \hat{p}_k\;\Big/\;\log K, \qquad \hat{p}_k = \frac{P_k}{\sum_j P_j},
\]
where $K$ is the number of frequency bins and $H\in[0,1]$ is the normalized entropy.
Forecastability is then defined as
\[
F = 1 - H.
\]
$F=1$ corresponds to a perfectly structured (deterministic) series, while $F=0$ corresponds to white noise.

\vspace{2pt}\noindent\textbf{Multivariate aggregation.}
Each \textsc{Quito} series is multivariate with five variates (channels).
We compute $T$, $S$, and $F$ independently for each variate channel, then obtain per-series diagnostics by averaging across the five channels:
\[
T_i = \frac{1}{5}\sum_{j=1}^{5} T_{i,j}, \quad
S_i = \frac{1}{5}\sum_{j=1}^{5} S_{i,j}, \quad
F_i = \frac{1}{5}\sum_{j=1}^{5} F_{i,j}.
\]

\vspace{2pt}\noindent\textbf{Regime labeling.}
Each of the three diagnostics is binarized using a fixed threshold of $0.4$: a series is labeled \textsc{high} if the value exceeds $0.4$ and \textsc{low} otherwise.
The three binary labels are combined into one of eight \textsc{tsf} regime cells, denoted \textsc{trend}$\times$\textsc{season}$\times$\textsc{forecast} (\eg \textsc{low\_high\_high}).
This threshold was chosen to yield balanced cell occupancy on training data and is consistent with the labeling used for the GIFT-Eval and Timer benchmarks in our analysis.
A sensitivity analysis over $\tau_{\mathrm{TSF}}\in[0.3,0.5]$ (\cref{app:threshold_sensitivity}) confirms that all qualitative findings are robust to this choice.

\begin{figure}[h]
\centering
\includegraphics[width=\linewidth]{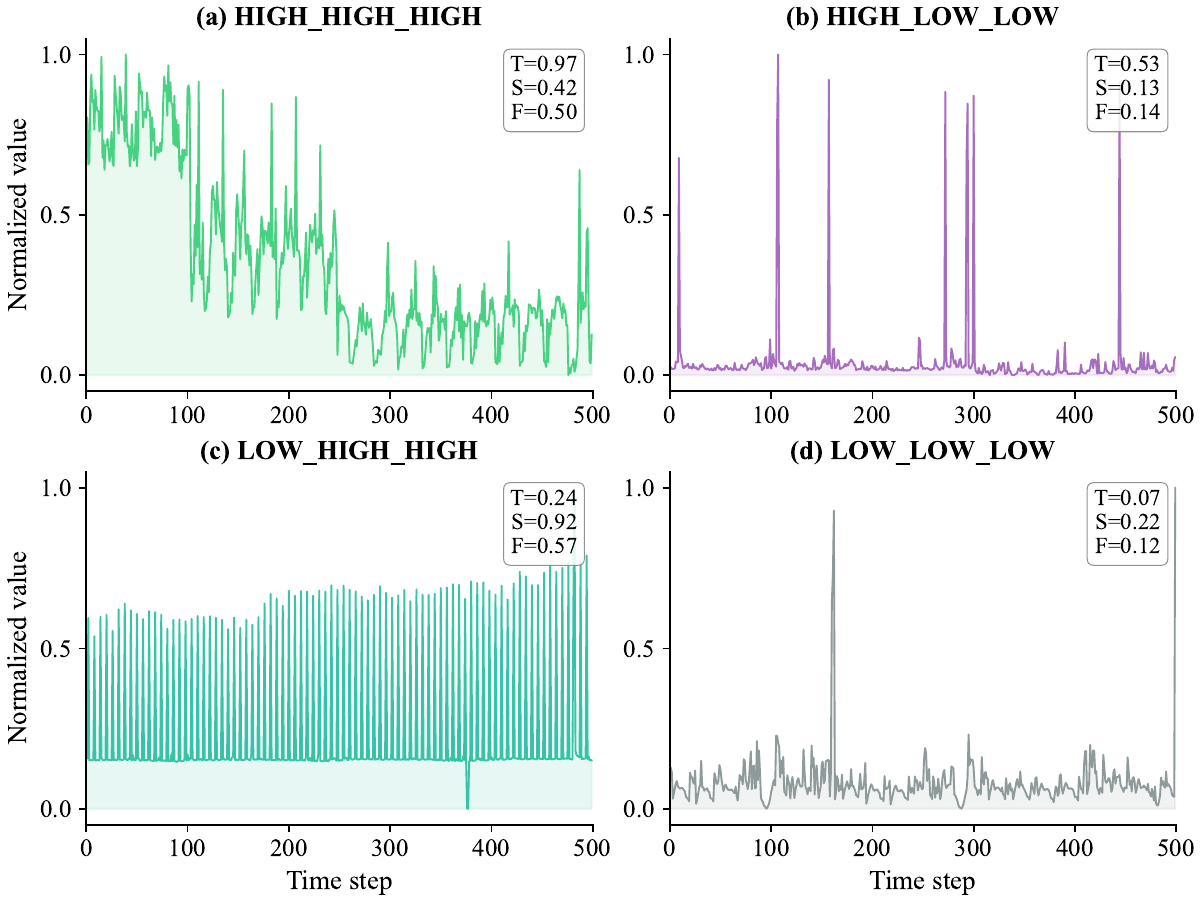}
\caption{Representative time series from four maximally contrasting \textsc{tsf} regimes (last 500 time steps, min--max normalized). Panel titles use the canonical \textsc{trend\_seasonality\_forecastability} notation; inset boxes report the exact diagnostic values. (a)~\textsc{high\_high\_high} displays a pronounced downward drift with moderate periodicity, whereas (b)~\textsc{high\_low\_low} is dominated by irregular spikes. (c)~\textsc{low\_high\_high} shows regular bar-like seasonality, and (d)~\textsc{low\_low\_low} exhibits unstructured noise.}
\label{fig:regime_examples}
\end{figure}

\Cref{fig:regime_examples} shows a representative time series from four maximally contrasting regimes.
Each regime exhibits a visually distinct waveform:
\begin{itemize}[nosep,leftmargin=*]
\item \textsc{high\_high\_high}: a pronounced downward drift with moderate periodicity, the easiest regime to forecast.
\item \textsc{high\_low\_low}: irregular spikes with no periodicity, the pathological regime where all models struggle.
\item \textsc{low\_high\_high}: regular, bar-like seasonality with no overall drift.
\item \textsc{low\_low\_low}: irregular noise with no discernible structure, the hardest regime.
\end{itemize}
The remaining four regimes interpolate between these extremes (\eg \textsc{high\_high\_low} shows spiky seasonal behaviour that is harder to predict despite its periodicity; \textsc{low\_high\_low} exhibits noisy seasonal fluctuations; \textsc{high\_low\_high} contains sparse spikes on a low base with an underlying trend; \textsc{low\_low\_high} is a quiet signal with occasional bursts).
These visual contrasts illustrate why forecastability is the strongest predictor of model difficulty (\cref{sec:analysis_forecastability}): low-forecastability regimes are consistently dominated by irregular spikes or noise.

\subsection{TSF Computation for External Benchmarks}
\label{app:tsf_external}
To enable cross-benchmark comparisons, we compute TSF diagnostics on the Timer and GIFT-Eval benchmarks using the same methodology as for \textsc{Quito}, with the following adaptations.

\emph{Timer benchmark.}
Each multivariate Timer dataset (ETT, ECL, Traffic, Weather, PEMS) is decomposed into independent univariate series, one per variate column. 
For each univariate series, the seasonal period $p$ is set according to the dataset's native resolution: $p{=}24$ for hourly series (ETTh, ECL, Traffic), $p{=}96$ for 15-minute series (ETTm), $p{=}144$ for 10-minute series (Weather), and $p{=}288$ for 5-minute series (PEMS).
Regime labels are then assigned with the same 0.4 threshold used for \textsc{Quito}, yielding eight TSF regime cells per series.

\emph{GIFT-Eval benchmark.}
GIFT-Eval~\citep{aksu2024gifteval} comprises 55 datasets spanning diverse domains and frequencies.
For multivariate series, only the first variate is retained, converting the task to univariate evaluation.
The seasonal period is auto-detected from the dataset frequency ($p{=}24$ for hourly, $p{=}7$ for daily, $p{=}52$ for weekly, $p{=}12$ for monthly, $p{=}4$ for quarterly).
The same 0.4 threshold and eight-cell regime taxonomy apply.

In both cases, the \texttt{evaluate\_series} function from our quality toolkit computes forecastability via Welch's spectral entropy, and trend and seasonality strength via STL decomposition, exactly as described in \cref{app:tsf_diagnostics} for \textsc{Quito}.
This ensures that regime labels are fully comparable across all three benchmarks.

\subsection{Cross-Provenance TSF Coverage Analysis}
\label{app:tsf_provenance}

A natural concern with single-provenance data is that the TSF regime distribution might reflect organization-specific workload characteristics.
To test this, we quantitatively compare the continuous TSF distributions of \textsc{Quito} (single provenance, 9 business verticals, 1{,}290 items) against the Timer benchmark (11 public datasets from 5 distinct domains and 4 countries: power transformers~\citep{zhou2021informer}, residential electricity, road traffic, weather stations~\citep{wu2021autoformer}, and highway sensors~\citep{chen2001freeway}), which represents the most widely used multi-provenance evaluation suite in time series forecasting.

\begin{figure}[h]
    \centering
    \begin{subfigure}[t]{0.32\linewidth}
        \centering
        \includegraphics[width=\linewidth]{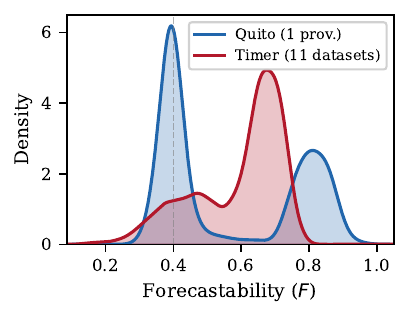}
        \caption{Forecastability ($F$)}
        \label{fig:tsf_provenance_kde_a}
    \end{subfigure}
    \hfill
    \begin{subfigure}[t]{0.32\linewidth}
        \centering
        \includegraphics[width=\linewidth]{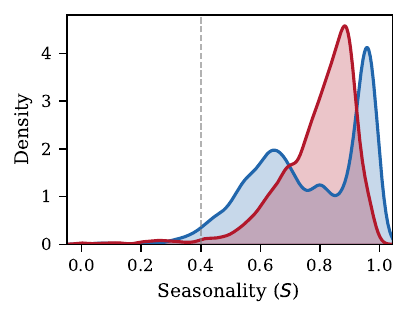}
        \caption{Seasonality ($S$)}
        \label{fig:tsf_provenance_kde_b}
    \end{subfigure}
    \hfill
    \begin{subfigure}[t]{0.32\linewidth}
        \centering
        \includegraphics[width=\linewidth]{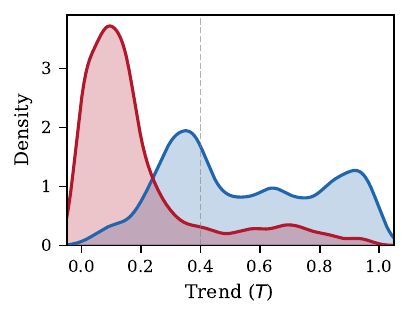}
        \caption{Trend ($T$)}
        \label{fig:tsf_provenance_kde_c}
    \end{subfigure}
    \vspace{2pt}
    \caption{Kernel density estimates of each TSF diagnostic for Quito (single provenance, blue) vs.\ Timer (multi-provenance, red).
    Dashed lines mark the $\tau{=}0.4$ binarization threshold.
    Quito's distributions are broader on all three axes, with particularly greater spread in forecastability and trend, confirming that single-provenance data is not restricted to a narrow TSF sub-region.}
    \label{fig:tsf_provenance_kde}
\end{figure}

\vspace{2pt}\noindent\textbf{TSF diversity metrics.}
For each TSF axis (trend $T$, seasonality $S$, forecastability $F$), we compare: (i) standard deviation as a measure of spread; (ii) Jensen--Shannon divergence (JSD) between the two distributions; and (iii) kernel density overlap.
\Cref{tab:tsf_provenance} summarizes the results; \cref{fig:tsf_provenance_kde} visualizes the distributions.

\begin{table}[h]
\centering
\small
\begin{sc}
\begin{tabular}{lccccc}
\toprule
Diagnostic & \multicolumn{2}{c}{Std dev} & Quito/Timer & JSD & KDE overlap \\
& Quito & Timer & (diversity ratio) & (bits$^2$) & \\
\midrule
$F$ & 0.205 & 0.131 & 1.57$\times$ & 0.603 & 0.40 \\
$S$ & 0.182 & 0.134 & 1.36$\times$ & 0.210 & 0.62 \\
$T$ & 0.260 & 0.210 & 1.24$\times$ & 0.437 & 0.35 \\
\midrule
\multicolumn{6}{l}{Regime entropy (bits; max = 3.00 for 8 cells)} \\
& 1.99 & 1.19 & 1.67$\times$ & \multicolumn{2}{c}{Quito: 66\% / Timer: 40\% of max} \\
\bottomrule
\end{tabular}
\end{sc}
\vspace{4pt}
\caption{TSF diversity comparison: Quito (single provenance, 9 verticals, 1{,}290 items) vs.\ Timer (11 public datasets, 5 domains, 2{,}950 series).
Quito achieves 1.24--1.57$\times$ higher standard deviation on every TSF axis and 1.67$\times$ higher regime entropy, demonstrating that single-provenance data can be \emph{more} diverse in TSF space than multi-provenance benchmarks.
JSD: Jensen--Shannon divergence (lower = more similar); KDE overlap: area under the minimum of two kernel density estimates (higher = more similar).}
\label{tab:tsf_provenance}
\end{table}

\vspace{2pt}\noindent\textbf{Key findings.}
\begin{itemize}[leftmargin=*]
    \item Quito achieves $1.24$--$1.57\times$ higher standard deviation than Timer on every TSF axis (\cref{tab:tsf_provenance}).
    The regime entropy of Quito's TSF distribution (1.99 bits, 66\% of maximum) is $1.67\times$ that of Timer (1.19 bits, 40\%), indicating far more balanced coverage.
    Timer concentrates 75.6\% of its series in a single regime (\textsc{low\_high\_high}), despite drawing from 11 datasets across 5 domains.

    \item The low KDE overlap (0.35--0.62) reflects distributional differences in \emph{location}, not deficiency in coverage.
    Timer series cluster around high seasonality and low trend (reflecting the dominance of highway-sensor and electricity-consumption data, which exhibit strong diurnal patterns with minimal drift), whereas Quito spans a wider range of trend and forecastability values, reflecting the heterogeneity of its nine business verticals (\cref{app:vertical_dist}).

    \item The cross-benchmark consistency reported in \cref{app:cross_benchmark_consistency} (Spearman $\rho{=}0.865$) provides an independent functional validation: model rankings obtained on Quito's single-provenance data transfer to Timer's multi-provenance data, confirming that the underlying difficulty drivers, not the data provenance, determine model performance.
\end{itemize}

\vspace{2pt}\noindent\textbf{Relation to public CloudOps traces.}
Published characterisations of public cloud traces (Azure VM workloads~\citep{cortez2017resource}, Google Borg clusters~\citep{tirmazi2020borg}, and Alibaba microservice traces~\citep{luo2021characterizing}) report the same qualitative diversity patterns observed in \textsc{Quito}: strong diurnal seasonality in user-facing services, weak or absent seasonality in batch pipelines, and forecastability ranging from near-deterministic (steady-state infrastructure) to near-random (bursty event-driven workloads).
Because TSF diagnostics measure these intrinsic statistical properties rather than application labels, the regime taxonomy is inherently provenance-agnostic; our quantitative comparison with Timer confirms this empirically.
We note that a direct TSF analysis of Azure and Google traces would further strengthen this evidence and leave it as promising future work.

\subsection{Why TSF Regimes over Domain-Based Splits?}
\label{app:tsf_justification}

Traditional benchmarks group series by application domain (\eg traffic, electricity, weather), yet these labels reflect data provenance rather than forecasting difficulty: two ``traffic'' series can differ far more in predictability than a ``traffic'' and an ``electricity'' series that share similar statistical structure.
We argue that categorizing series by their intrinsic trend, seasonality, and forecastability (TSF) regime is both more principled and more useful, and our experiments provide three lines of evidence.

\begin{itemize}[leftmargin=*]
    \item Forecastability dominates domain as a difficulty predictor.
    Across all ten models, high-forecastability series achieve $3.64\times$ lower MAE than low-forecastability series, regardless of their application domain (\cref{tab:forecastability_analysis}).
    In contrast, domain labels show no consistent relationship with error magnitude: traffic series, for instance, span the full difficulty range from the easiest to the hardest regime cells (\cref{fig:quito_grid}).

    \item TSF regimes yield actionable model-selection guidance.
    Because trend, seasonality, and forecastability can be computed from any new dataset, practitioners can directly consult regime-specific rankings to choose an appropriate model.
    For example, deep learning models hold 18--38\% MAE advantages in low-seasonality regimes (\cref{tab:group_analysis}), while foundation models dominate in 6 of 8 high-seasonality or high-forecastability groups.
    Domain labels offer no comparable mapping between data characteristics and model strengths.

    \item TSF-based rankings generalize across benchmarks.
    Despite QuitoBench and the Timer benchmark having entirely different provenance and distributional profiles, model rankings on the two benchmarks exhibit a Spearman correlation of $\rho{=}0.865$ ($p{<}0.01$; \cref{tab:timer_consistency,fig:quito_timer_consistency}).
    This confirms that TSF-based evaluation captures fundamental forecasting behavior rather than dataset-specific artifacts, and that findings transfer to real-world, domain-diverse distributions.
\end{itemize}

\subsection{QuitoBench Test Set Details}
\label{app:quitobench_testset}

\vspace{2pt}\noindent\textbf{Train, valid and test split.}
We use a temporal split with a global test cutoff at 2023-07-28 00:00:00.
Data before this point is divided into train/valid (80/20 ratio); data from this point onward forms the test set.
For each series: $\text{valid\_size} = \lfloor \text{train\_valid\_size} \times 0.2 \rfloor$, 
$\text{train\_size} = \text{train\_valid\_size} - \text{valid\_size}$, and 
$\text{test\_size} = \text{total\_length} - \text{train\_valid\_size}$, 
where $\text{train\_valid\_size}$ is the index at the global test cutoff.
This ensures zero temporal leakage and chronological ordering (Train $\rightarrow$ Valid $\rightarrow$ Test).

\section{Experiment Details}
\label[appendix]{app:exp}

\subsection{Implementation Details}\label{app:imp}

All experiments are implemented in PyTorch~\citep{paszke2019pytorch} and managed through a YAML-based configuration system that specifies data paths, model architecture, training hyperparameters, and evaluation settings in a single file per run.
Data are stored in Apache Parquet format and loaded via PyArrow; each series is reconstructed by filtering on \texttt{item\_id} and sorting by \texttt{date\_time}.

For the five deep learning models (CrossFormer, DLinear, iTransformer, PatchTST, TSMixer), we use our own implementations following the original papers.
Training uses the Adam or AdamW optimizer with a cosine learning-rate schedule and MSE as the loss function (see \cref{app:training_details} for the full three-stage pipeline).
Multi-GPU runs use PyTorch Distributed Data Parallel (DDP) with the NCCL backend.

Foundation models are loaded from their publicly released checkpoints: Chronos-2~\citep{ansari2025Chronos} via the Amazon \texttt{Chronos-forecasting} package, TimesFM-2.5~\citep{das2023TimesFM} via Google's official release, and TiRex~\citep{25tirex} via its public repository.
No gradient updates are performed during foundation model evaluation.
Statistical baselines (Exponential Smoothing and Seasonal Na\"ive) are computed with \texttt{statsmodels}.

TSF diagnostics (trend strength, seasonality strength, and forecastability) are computed using STL decomposition (\texttt{statsmodels}) and Welch's spectral entropy (\texttt{scipy}); see \cref{app:tsf_diagnostics} for formulations.
The evaluation framework and project site are built upon the codebase of FAMMA~\citep{xue2024famma}.
All evaluation, analysis, and figure-generation code is included in the public release to ensure full reproducibility.

\subsection{Training and Testing Details}\label{app:training_details}

\vspace{2pt}\noindent\textbf{Deep learning model pipeline.}
Each deep learning model is trained and evaluated through a three-stage pipeline, implemented using the YAML-based configuration system in the \textsc{Quito} codebase.

\begin{enumerate}[leftmargin=*, label=\textbf{Stage \arabic*.}]

\item Hyperparameter tuning.
For each task configuration (context length $L$, forecast horizon $H$, and forecasting mode), we first conduct a hyperparameter search to identify the best model architecture and training parameters.
The search space covers architecture parameters (such as hidden dimension ($d_{\text{model}}$), number of attention heads ($n_{\text{heads}}$), feedforward width ($d_{\text{ff}}$), dropout rate, patch length, and stride).
Candidate configurations are trained for a small number of epochs, and the best configuration is selected based on MSE on the validation split.

\item Fine-tuning.
With the best hyperparameters fixed, the model is trained from scratch on the training split for up to 100 epochs using the Adam or AdamW optimizer with a cosine learning-rate schedule.
Checkpoints are saved every epoch and the best checkpoint (lowest validation MSE) is retained via early stopping.
All training uses MSE as the loss function.

\item Evaluation.
The best checkpoint is loaded and evaluated on the test split via the evaluation runner (\texttt{runner: eval}).
Both MAE and MSE are computed per series.
To account for random initialization, once the hyperparameters are fixed, each model is trained and evaluated under $N=3$ different random seeds; the reported MAE is the mean across seeds.

\end{enumerate}

\vspace{2pt}\noindent\textbf{Foundation model inference.}
Foundation models (Chronos-2, TimesFM-2.5, and TiRex) are applied in a \emph{zero-shot} manner: their publicly released pre-trained weights are loaded directly without any task-specific fine-tuning on \textsc{QuitoBench} data.
For each task configuration, the forecast horizon $H$ is passed as a prediction-length parameter and the context window of length $L$ is used as input; no gradient updates are performed.
This ensures that the reported foundation model results reflect pure generalization from pre-training, with no information leakage from the benchmark splits.

\vspace{2pt}\noindent\textbf{Scaling experiment details.}\label{app:finetuning_details}
The data- and model-scaling experiments in Analysis~I trained TimesFM-2.5 alongside CrossFormer from scratch.
Below we detail the choice of foundation model, the training protocol, and the compute cost.

\emph{Why TimesFM-2.5.}
Among the three foundation models in our benchmark, TimesFM-2.5 is the only one whose architecture supports straightforward fine-tuning with a standard regression loss.
TimesFM-2.5~\citep{das2023TimesFM} is a decoder-only transformer that maps a continuous-valued input patch sequence directly to a continuous-valued forecast via a regression head.
Its forward pass is fully differentiable end-to-end, so fine-tuning with MSE loss proceeds exactly as for any task-specific deep learning model: one simply unfreezes the pre-trained weights and continues training on in-domain data.

Chronos~\citep{ansari2024chronos}, by contrast, is built on a T5 encoder--decoder backbone and uses a \emph{tokenisation-based} representation: continuous time-series values are first scaled, then quantized into 4\,096 discrete bins by a learned tokenizer, and the model is trained with cross-entropy loss over these token IDs.
Fine-tuning such an architecture introduces several complications:
(i)~the bin boundaries of the tokenizer are calibrated during pre-training on a specific data distribution; applying them to out-of-distribution series can cause significant quantization mismatch;
(ii)~the cross-entropy training objective is fundamentally different from the MSE objective used for all other models in our benchmark, making the comparison less controlled;
and (iii)~the \texttt{Chronos-forecasting} library, as of our experimental deadline, does not expose a documented fine-tuning API: accessing the underlying T5 model for gradient updates requires non-trivial workarounds through the \texttt{ChronosPipeline} abstraction.

TiRex~\citep{25tirex} employs a retrieval-augmented architecture that conditions forecasts on retrieved exemplar series at inference time.
Its publicly released codebase does not support fine-tuning, and adapting its retrieval mechanism to a new corpus would constitute a substantial engineering effort beyond the scope of this study.

Selecting TimesFM-2.5 (the largest foundation model in our suite at 200\,M parameters) therefore provides the strongest and fairest test of whether domain-specific fine-tuning can close the gap with task-specific deep learning.

\emph{Training protocol.}
TimesFM-2.5 is trained from scratch using the AdamW optimizer with learning rate $1\times10^{-5}$, cosine schedule, and MSE loss, the same loss and optimizer family used for all deep learning models.
CrossFormer is trained from scratch with its best hyperparameters selected on the validation split (Stage~1 above).
Both models are trained for up to 50 epochs with early stopping based on validation MSE, using batch size 256.

\emph{Compute cost.}
All scaling experiments are conducted on a eight NVIDIA A100 80\,GB GPU.


\vspace{2pt}\noindent\textbf{Rolling window evaluation.}\label{app:rolling_windows}
As introduced in \cref{sec:exp_setup}, \textsc{QuitoBench} adopts dense rolling windows with unit stride for test-set evaluation, in contrast to the sparse non-overlapping scheme used by most existing benchmarks.
Here we provide the full formulation and quantitative comparison.

\vspace{2pt}\noindent\textbf{Formulation.}
Each series is partitioned 70\%/20\%/10\% into train, validation, and test portions.
The test data block for a given context length $L$ spans $T_{\text{test}}+L$ consecutive time steps, where $T_{\text{test}}$ is the length of the 10\% test segment ($T_{\text{test}}\approx 1{,}536$ steps for hourly series and $T_{\text{test}}\approx 590$ steps for 10-minute series).
Sliding a window of total length $L+H$ one step at a time gives
\[
W(H) \;=\; T_{\text{test}} - H + 1
\]
evaluation windows per series per $(L,H)$ pair.
The per-series MAE is the mean over all $W(H)$ windows.
Notably, $W(H)$ is independent of $L$: changing the context length shifts which history is visible, but the number of prediction targets remains the same.

\vspace{2pt}\noindent\textbf{Comparison with \textsc{gift-eval}.}
\textsc{gift-eval} generates windows with stride $H$ (non-overlapping), with the number of windows set to $\lceil 0.1\,T_{\min}/H \rceil$ (where $T_{\min}$ is the shortest series in the dataset), hard-capped at $\mathtt{MAX\_WINDOW}=20$.
In practice this yields at most a small number of MAE samples per (series, $H$) pair (for long-horizon tasks a single series may have only one or two evaluation windows), making headline metrics sensitive to the specific evaluation endpoints chosen.
The dense scheme of \textsc{QuitoBench} provides hundreds to over a thousand windows per series (see \cref{tab:rolling_windows}), producing far more stable per-series error estimates and enabling reliable regime-level analysis across the 1,290 benchmark series.

\begin{figure*}[h]
\centering
\includegraphics[width=\textwidth]{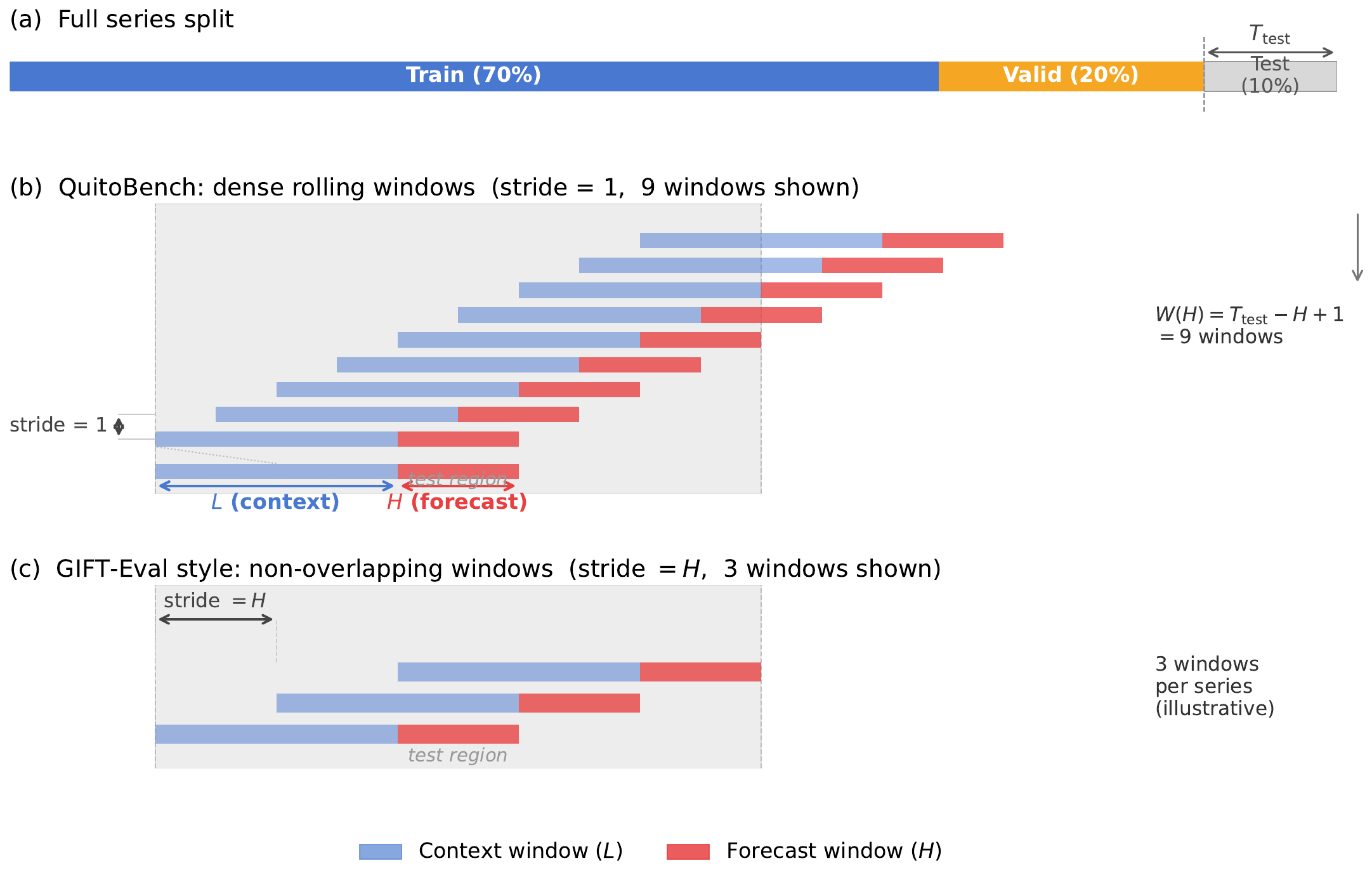}
\caption{Illustration of the rolling window evaluation strategy.
(a) Each series is split 70\%/20\%/10\% into train, validation, and test portions.
(b) \textsc{QuitoBench} slides a context+forecast window one step at a time across the test region, generating $W(H)=T_{\text{test}}-H+1$ windows per series per $(L,H)$ configuration.
(c) In contrast, \textsc{gift-eval}-style evaluation uses non-overlapping windows with stride $H$ (here shown with 3 windows for illustration), yielding far fewer evaluation samples per series.}
\label{fig:rolling_windows}
\end{figure*}

\Cref{tab:rolling_windows} lists $W(H)$ for each frequency and horizon, and \cref{fig:rolling_windows} illustrates the contrast between the two evaluation strategies.

\begin{table}[h]
    \centering
    \small
    \begin{sc}
    \begin{tabular}{lrrr|r}
        \toprule
        Freq & $H=48$ & $H=288$ & $H=512$ & Total (3 $L$ values, 3 $H$ values) \\
        \midrule
        Hour ($T_{\text{test}}\approx1{,}536$)    & 1,489 & 1,249 & 1,025 & 11,289 \\
        10-min ($T_{\text{test}}\approx590$)      &   543 &   303 &    79 &  2,775 \\
        \bottomrule
    \end{tabular}
    \end{sc}
    \vspace{4pt}
    \caption{Number of rolling evaluation windows per series for each forecast horizon $H$ and granularity.
    The rightmost column sums $W(H)$ over all three context lengths $L\in\{96,576,1024\}$ and all three horizons.}
    \label{tab:rolling_windows}
\end{table}

Across all 517 hourly and 773 10-minute test series, two forecasting modes, and all $(L,H)$ configurations, \textsc{QuitoBench} generates approximately $1.6\times10^7$ individual predictions per model (one to two orders of magnitude more than \textsc{gift-eval}'s non-overlapping scheme), providing far more stable performance estimates at the series level.

\subsection{Model Details and Evaluation Metrics}
\label[appendix]{app:metrics}

\vspace{2pt}\noindent\textbf{Model sizes.}
\Cref{tab:model_sizes} lists the parameter counts for all benchmarked models.
Deep learning models are task-specific and their size varies with the task configuration (context length, forecast horizon, number of variates); the ranges below correspond to the full sweep of configurations used in this work.
Foundation models are fixed-size pre-trained networks whose weights are not updated during evaluation.

\begin{table}[h]
\centering
\small
\begin{sc}
\begin{tabular}{llrr}
\toprule
Model & Category & Size range & Mean size \\
\midrule
Crossformer  & Deep learning  & 0.5M--3M   & $\sim$1M   \\
DLinear      & Deep learning  & 10K--1M    & $\sim$300K \\
iTransformer & Deep learning  & 0.5M--5M   & $\sim$2M   \\
PatchTST     & Deep learning  & 1M--15M    & $\sim$5M   \\
TSMixer      & Deep learning  & 100K--3M   & $\sim$1M   \\
\midrule
TiRex        & Foundation     & 30M        & 30M  \\
Chronos-2    & Foundation     & 100M       & 100M \\
TimesFM-2.5  & Foundation     & 200M       & 200M \\
\bottomrule
\end{tabular}
\end{sc}
\vspace{4pt}
\caption{Parameter counts for benchmarked models. Deep learning model sizes vary across task configurations; foundation model sizes are fixed.}
\label{tab:model_sizes}
\end{table}

\vspace{2pt}\noindent\textbf{Ranking methodology.}
To aggregate fairly across series with heterogeneous scales, we use a rank-first-then-average approach.
For each test series and task configuration, the ten models are ranked 1 (best) to 10 (worst) by MAE; ties receive average ranks.
The reported \emph{mean rank} for a model is the average of its per-series, per-configuration ranks over the entire evaluation set (or over a subset of \textsc{tsf} regime cells for macro-averaged results).
This ensures that each series contributes equally regardless of its absolute error scale, and that no single outlier series dominates the aggregate.

\begin{table}[h]
    \centering
    \small
    \begin{sc}
    \begin{tabular}{llrrl}
        \toprule
        Model & Category & MV MAE & UV MAE & Preferred mode \\
        \midrule
        CrossFormer  & Deep learning  & 0.282 & 0.275 & UV ($+$2.5\%) \\
        PatchTST     & Deep learning  & 0.299 & 0.298 & UV ($+$0.4\%) \\
        TSMixer      & Deep learning  & 0.313 & 0.309 & UV ($+$1.3\%) \\
        iTransformer & Deep learning  & 0.299 & 0.302 & MV ($+$0.7\%) \\
        DLinear      & Deep learning  & 0.368 & 0.371 & MV ($+$0.7\%) \\
        \midrule
        Chronos-2    & Foundation     & 0.310 & 0.317 & MV ($+$2.3\%) \\
        TimesFM-2.5  & Foundation     & 0.319 & 0.319 & Neutral \\
        TiRex        & Foundation     & 0.322 & 0.322 & Neutral \\
        \bottomrule
    \end{tabular}
    \end{sc}
    \vspace{4pt}
    \caption{Mean MAE by MV and UV mode. ``Preferred mode'' shows the better option and relative improvement; differences below 0.1\% are marked Neutral.}
    \label{tab:mode_analysis}
\end{table}

\subsection{TSF Regime Analysis: Forecastability, Specialization, and Pathological Regimes}
\label{app:tsf_regime_analysis}

\vspace{2pt}\noindent\textbf{Forecastability as the dominant difficulty driver.}
While trend and seasonality strength are commonly cited as predictors of forecast difficulty, our analysis reveals that \emph{forecastability} (the inherent predictability of a time series based on its signal-to-noise ratio, stability, and regularity) is the dominant factor.

\Cref{tab:forecastability_analysis} shows mean MAE across all models for each \textsc{tsf} regime, ranked by difficulty.
The easiest regime (\textsc{high\_high\_high}: high trend, high seasonality, high forecastability) averages MAE 0.205, while the hardest (\textsc{high\_low\_low}: high trend, low seasonality, low forecastability) averages 0.749, a dramatic 3.64$\times$ difference.
Strikingly, even series with low trend and low seasonality but high forecastability (\textsc{low\_low\_high}) achieve MAE 0.220, only 7\% harder than the easiest regime.
Conversely, series with high trend and seasonality but low forecastability (\textsc{high\_high\_low}) are 2.32$\times$ harder than the easiest, despite having strong structural patterns.
This confirms that forecastability, not structural complexity per se, drives predictive difficulty.

\begin{table}[h]
\centering
\small
\begin{sc}
\begin{tabular}{llcccl}
\toprule
Rank & TSF Regime & Trend & Season & Forecast & Mean MAE \\
\midrule
1 (Easiest) & \textsc{high\_high\_high} & HIGH & HIGH & HIGH & 0.205 \\
2 & \textsc{low\_low\_high} & LOW & LOW & HIGH & 0.220 \\
3 & \textsc{low\_high\_high} & LOW & HIGH & HIGH & 0.299 \\
4 & \textsc{low\_high\_low} & LOW & HIGH & LOW & 0.359 \\
5 & \textsc{high\_low\_high} & HIGH & LOW & HIGH & 0.376 \\
6 & \textsc{low\_low\_low} & LOW & LOW & LOW & 0.456 \\
7 & \textsc{high\_high\_low} & HIGH & HIGH & LOW & 0.478 \\
8 (Hardest) & \textsc{high\_low\_low} & HIGH & LOW & LOW & \textbf{0.749} \\
\bottomrule
\end{tabular}
\end{sc}
\vspace{4pt}
\caption{TSF regime difficulty ranking by mean MAE across all models. \textsc{high\_low\_low} is 3.64$\times$ harder than the easiest regime, demonstrating forecastability's dominant effect.}
\label{tab:forecastability_analysis}
\end{table}

\vspace{2pt}\noindent\textbf{Model sensitivity to forecastability.}
Model sensitivity to forecastability varies significantly by architecture family, as shown in \cref{tab:model_forecastability_sensitivity}.
Deep learning models show the highest sensitivity: PatchTST, iTransformer, TSMixer, and CrossFormer all exhibit approximately 2.2--2.3$\times$ performance degradation when moving from high to low forecastability series.
Foundation models (Chronos-2, TimesFM-2.5, TiRex) are notably more robust, with only 1.7--1.8$\times$ degradation, suggesting their pre-training on diverse corpora provides better noise tolerance.
Traditional baselines (SNaive, ES) fall in between at 1.6$\times$, while DLinear shows moderate sensitivity at 2.0$\times$.
This pattern suggests that for low-forecastability series (common in volatile cloud monitoring scenarios), foundation models offer superior robustness despite their larger parameter count.

\begin{table}[h]
    \centering
    \small
    \begin{sc}
    \begin{tabular}{lccc}
        \toprule
        Model & HIGH Forecast MAE & LOW Forecast MAE & Ratio \\
        \midrule
        PatchTST & 0.185 & 0.420 & 2.28$\times$ \\
        iTransformer & 0.186 & 0.422 & 2.26$\times$ \\
        TSMixer & 0.194 & 0.436 & 2.25$\times$ \\
        CrossFormer & 0.174 & 0.390 & 2.24$\times$ \\
        DLinear & 0.245 & 0.501 & 2.04$\times$ \\
        \midrule
        SNaive & 0.517 & 0.843 & 1.63$\times$ \\
        ES & 0.550 & 0.850 & 1.55$\times$ \\
        \midrule
        TimesFM-2.5 & 0.235 & 0.409 & 1.74$\times$ \\
        TiRex & 0.237 & 0.413 & 1.74$\times$ \\
        Chronos-2 & 0.230 & 0.403 & 1.75$\times$ \\
        \bottomrule
    \end{tabular}
    \end{sc}
    \vspace{4pt}
    \caption{Model sensitivity to forecastability. Deep learning models (top) show higher sensitivity (2.0--2.3$\times$) than foundation models (bottom, 1.7--1.8$\times$), indicating greater robustness of pre-trained representations for unpredictable series.}
\label{tab:model_forecastability_sensitivity}
\end{table}

\vspace{2pt}\noindent\textbf{The pathological \textsc{high\_low\_low} regime.}
The \textsc{high\_low\_low} regime (characterized by high trend, low seasonality, and low forecastability) represents a uniquely challenging forecasting scenario.
With mean MAE 0.749, it is 56.7\% harder than the second-most-difficult regime (\textsc{high\_high\_low}: 0.478) and 3.64$\times$ harder than the easiest.
This pathology arises from a problematic combination: high trend creates directional drift requiring extrapolation, low seasonality removes periodic structure that could anchor predictions, and low forecastability adds noise that obscures true patterns.

\Cref{tab:high_low_low_performance} shows model performance rankings within this challenging regime.
CrossFormer achieves the best relative performance (MAE 0.600), though this is still 2.91$\times$ worse than its performance on the easiest regime.
Foundation models cluster tightly behind (Chronos-2 0.628, TimesFM-2.5 0.633, TiRex 0.652), with only 3.8\% separating second from fourth place.
Deep learning models show wider variance: iTransformer, PatchTST, and TSMixer range from 0.656 to 0.691 MAE, while DLinear degrades to 0.805 (+34.2\% vs.\ CrossFormer).
Statistical baselines fail catastrophically (ES 1.061, SNaive 1.091).
The tight clustering of foundation models suggests their pre-training provides a performance floor, while CrossFormer's cross-dimension attention may better distinguish signal from noise in limited-structure environments.

\begin{table}[h]
    \centering
    \small
    \begin{sc}
    \begin{tabular}{clcc}
        \toprule
        Rank & Model & MAE & Gap from Best \\
        \midrule
        1 & CrossFormer & 0.600 & --- \\
        2 & Chronos-2 & 0.628 & +4.6\% \\
        3 & TimesFM-2.5 & 0.633 & +5.5\% \\
        4 & TiRex & 0.652 & +8.6\% \\
        5 & iTransformer & 0.656 & +9.3\% \\
        6 & PatchTST & 0.669 & +11.5\% \\
        7 & TSMixer & 0.691 & +15.2\% \\
        8 & DLinear & 0.805 & +34.3\% \\
        9 & ES & 1.061 & +77.0\% \\
        10 & SNaive & 1.091 & +81.9\% \\
        \bottomrule
    \end{tabular}
    \end{sc}
    \vspace{4pt}
    \caption{Model performance on the pathological \textsc{high\_low\_low} regime. Even the best model (CrossFormer) performs 3$\times$ worse than on easy series, and statistical baselines fail catastrophically.}
    \label{tab:high_low_low_performance}
\end{table}

Practically, \textsc{high\_low\_low} series require specialized handling: trend decomposition, denoising preprocessing, ensemble approaches with uncertainty quantification, and, critically, realistic error expectations.

\subsection{TSF Threshold Sensitivity Analysis}
\label{app:threshold_sensitivity}

\begin{figure}[h]
    \centering
    \begin{subfigure}[t]{0.32\linewidth}
        \centering
        \includegraphics[width=\linewidth]{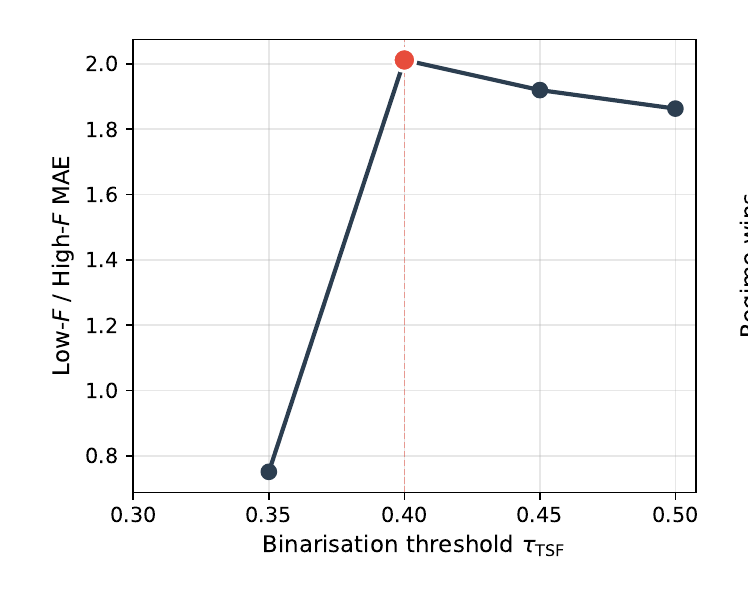}
        \caption{Forecastability gap}
        \label{fig:thresh_sens_a}
    \end{subfigure}
    \hfill
    \begin{subfigure}[t]{0.32\linewidth}
        \centering
        \includegraphics[width=\linewidth]{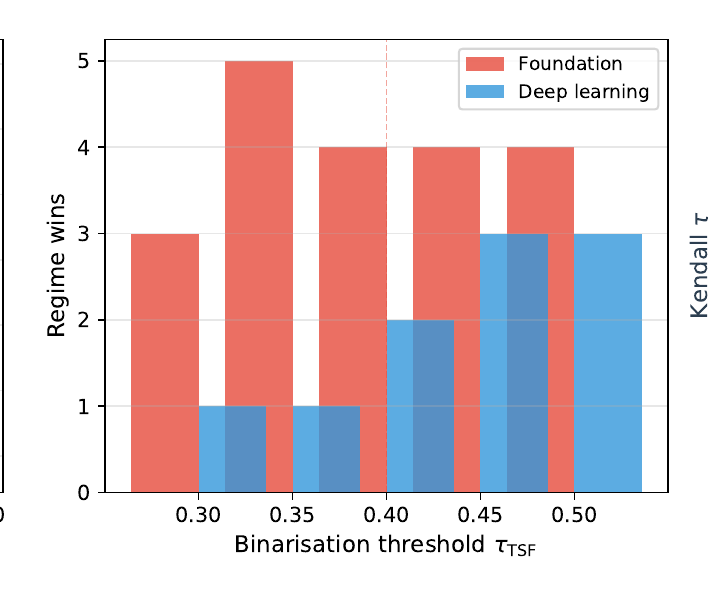}
        \caption{Family regime wins}
        \label{fig:thresh_sens_b}
    \end{subfigure}
    \hfill
    \begin{subfigure}[t]{0.32\linewidth}
        \centering
        \includegraphics[width=\linewidth]{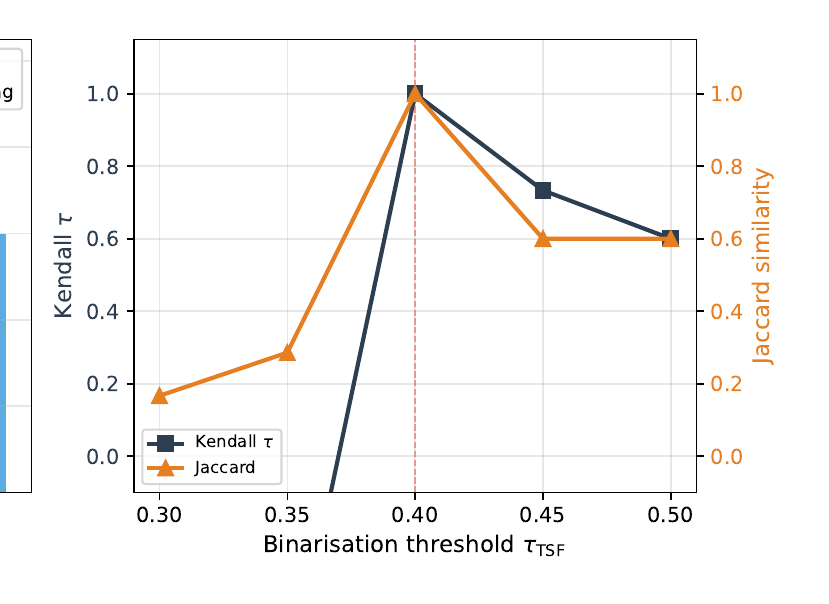}
        \caption{Ranking \& winner stability}
        \label{fig:thresh_sens_c}
    \end{subfigure}
    \caption{
        TSF threshold sensitivity.
        \subref{fig:thresh_sens_a}~The forecastability gap (low-$F$/high-$F$ MAE) is large and stable for $\tau_{\mathrm{TSF}}\ge0.40$.
        \subref{fig:thresh_sens_b}~Foundation models consistently win more regimes than deep learning across all thresholds.
        \subref{fig:thresh_sens_c}~Regime difficulty ranking (Kendall~$\tau$) and FM-winning-regime overlap (Jaccard) remain high for $\tau_{\mathrm{TSF}}\in[0.40,0.50]$.
    }
    \label{fig:threshold_sensitivity}
\end{figure}

A potential concern with the TSF regime framework is that the binarization threshold $\tau_{\mathrm{TSF}}=0.4$ is an analyst choice.
To establish robustness, we sweep $\tau_{\mathrm{TSF}}\in\{0.30,0.35,0.40,0.45,0.50\}$ and re-evaluate every claim that depends on the regime construction.
For each threshold, we re-classify all 1{,}290 test items using the continuous per-channel-averaged TSF diagnostics (the same methodology described in \cref{app:tsf_diagnostics}) and recompute (i) the forecastability gap (ratio of low-$F$ to high-$F$ MAE), (ii) the identity of the easiest and hardest regimes, (iii) the number of regimes won by foundation models versus deep learning, (iv) the Kendall~$\tau$ between the regime difficulty ranking and the reference ranking at $\tau_{\mathrm{TSF}}{=}0.40$, and (v) the Jaccard similarity of the set of foundation-model-winning regimes.
\Cref{tab:threshold_sensitivity} summarizes the results; \cref{fig:threshold_sensitivity} visualizes key trends.

\vspace{2pt}\noindent\textbf{Key findings.}
\begin{itemize}[leftmargin=*]
    \item Forecastability dominates difficulty at every threshold:
    for $\tau_{\mathrm{TSF}}\in[0.40,0.50]$, the low-$F$/high-$F$ MAE ratio ranges from $1.86\times$ to $2.01\times$.
    At $\tau_{\mathrm{TSF}}<0.40$ the split becomes degenerate (nearly all items are high-$F$), which itself validates $0.4$ as a lower bound for meaningful separation.

    \item CrossFormer is the best overall model at every threshold: this finding is completely independent of the regime construction.

    \item Foundation models win at least as many regimes as deep learning across all thresholds ($3$--$5$ FM wins vs.\ $1$--$3$ DL wins).

    \item For $\tau_{\mathrm{TSF}}\in[0.40,0.50]$, the regime difficulty ranking is stable: Kendall~$\tau\ge0.60$ and Jaccard similarity of FM-winning regimes $\ge0.60$ relative to the reference.
    The high-forecastability regimes are consistently the easiest and low-forecastability regimes the hardest, confirming that the dominant role of forecastability is not an artifact of any single threshold choice.

    \item Extreme thresholds ($\tau_{\mathrm{TSF}}\le0.35$) produce highly skewed cell occupancy (\eg $>80\%$ of items in a single cell at $\tau{=}0.30$), rendering regime-level analysis uninformative.
    This further justifies the choice of $0.4$, which yields the most balanced occupied-cell distribution while remaining consistent with the GIFT-Eval and Timer benchmarks.
\end{itemize}

\begin{table}[h]
    \centering
    \small
    \begin{sc}
    \begin{tabular}{ccccccc}
        \toprule
        $\tau_{\mathrm{TSF}}$ & High-$F$ MAE & Low-$F$ MAE & $F$-gap & FM / DL wins & Kendall~$\tau$ & Jaccard \\
        \midrule
        0.30 & ---  & --- & ---        & 3/4 vs.\ 1/4 & --- & 0.17 \\
        0.35 & 0.317 & 0.238 & 0.75$\times$ & 5/6 vs.\ 1/6 & $-$0.67 & 0.29 \\
        0.40\rlap{\,\textdagger} & 0.211 & 0.424 & 2.01$\times$ & 4/6 vs.\ 2/6 & 1.00 & 1.00 \\
        0.45 & 0.209 & 0.400 & 1.92$\times$ & 4/7 vs.\ 3/7 & 0.73 & 0.60 \\
        0.50 & 0.210 & 0.391 & 1.86$\times$ & 4/7 vs.\ 3/7 & 0.60 & 0.60 \\
        \bottomrule
    \end{tabular}
    \end{sc}
    \vspace{4pt}
    \caption{TSF threshold sensitivity analysis.
    $F$-gap: ratio of low-$F$ to high-$F$ mean MAE; FM/DL wins: occupied regimes won by each model family; Kendall~$\tau$ and Jaccard are computed against the reference threshold $\tau_{\mathrm{TSF}}{=}0.40$ (\textdagger).
    At $\tau{=}0.30$ nearly all items are classified as high-$F$, so the $F$-gap and ranking metrics are undefined (---).
    }
    \label{tab:threshold_sensitivity}
\end{table}

\subsection{Parameter Efficiency Frontier Plots}
\label{app:efficiency_plots}

\Cref{tab:param_roi_full} provides the full parameter efficiency comparison between foundation models and deep learning models, including context-conditional and per-parameter metrics.

\begin{table}[h]
    \centering
    \small
    \begin{sc}
    \begin{tabular}{lrrr}
        \toprule
        Metric & Foundation & Deep learning & ROI \\
        \midrule
        Avg.\ params (M)           & 110    & 1.9    & $59\times$ fewer \\
        Best MAE $\downarrow$      & 0.3138 & 0.2789 & deep learning $-11\%$ \\
        Mean MAE $\downarrow$      & 0.3185 & 0.3117 & deep learning $-2\%$ \\
        \midrule
        MAE at $L{=}96$            & 0.4551 & 0.3432 & deep learning $-25\%$ \\
        MAE at $L{\ge}576$         & 0.2502 & 0.2960 & foundation model $-15\%$ \\
        \midrule
        MAE / M\,params            & 0.0029 & 0.1676 & $57.9\times$ \\
        Rank / M\,params           & 0.0361 & 2.6505 & $73.3\times$ \\
        \bottomrule
    \end{tabular}
    \end{sc}
    \vspace{4pt}
    \caption{Full parameter efficiency comparison: foundation models (avg.\ 110\,M params) vs.\ deep learning models (avg.\ 1.9\,M params), covering overall performance, context-conditional performance, and per-parameter efficiency.}
    \label{tab:param_roi_full}
\end{table}

\begin{figure}[h]
    \centering
    \includegraphics[width=\linewidth]{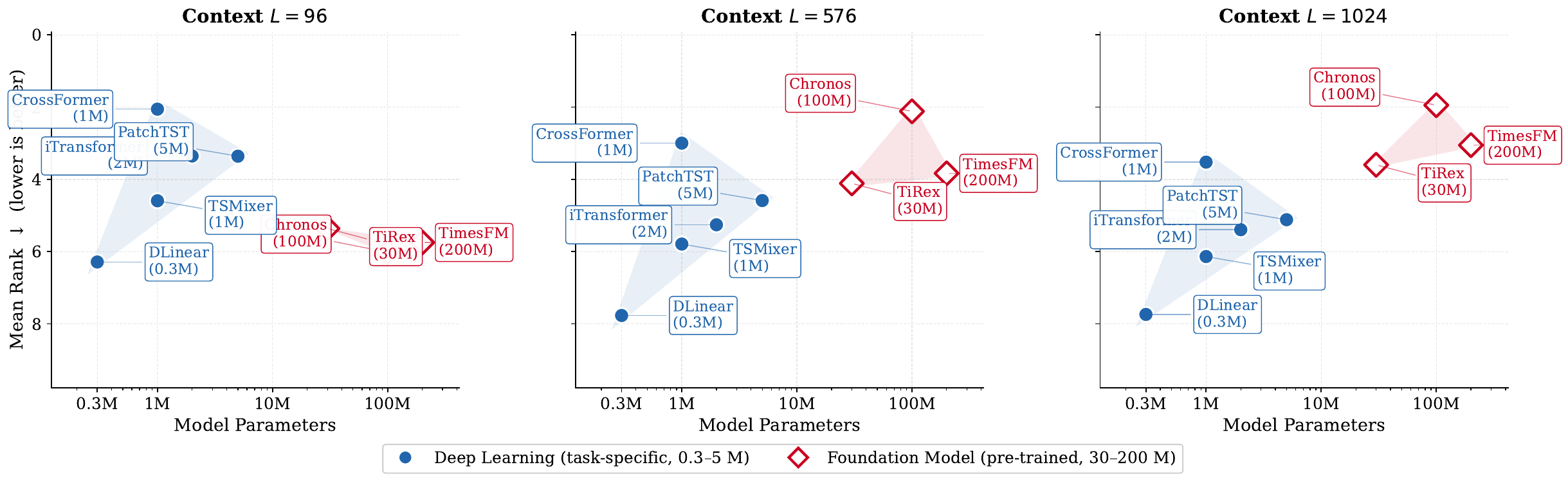}
    \vspace{2pt}
    \caption{Efficiency frontier stratified by context length $L$. At $L=96$ the deep learning cluster dominates; at $L\ge576$ foundation models (Chronos-2) rise to the top, illustrating the context-conditional nature of the efficiency advantage.}
    \label{fig:efficiency_context}
\end{figure}

\Cref{fig:efficiency_context} decomposes the efficiency frontier by context length $L$: at $L = 96$, the deep learning cluster sits clearly above foundation models in rank (24.6\% MAE advantage), and CrossFormer wins all six task configurations at this context.
At $L \ge 576$ the picture reverses: Chronos-2 (100\,M) rises to the top, with foundation models leading by 15--22\%, because their large capacity, honed on diverse pretraining corpora, enables 43--50\% MAE improvement from extended history, whereas deep learning models gain only 7--12\%.
The practical implication is therefore nuanced: for short-context or resource-constrained deployments, task-specific deep learning models offer the best accuracy-per-parameter trade-off; when long historical context is available, the additional capacity of Chronos-2 begins to pay off.

\subsection{Statistical Tests}
\label{app:stat_tests}

This section details the statistical tests referenced in the main text.
All tests are performed on the 232,200 evaluation instances (1,290 unique evaluation items $\times$ 10 models $\times$ 18 configurations).

\vspace{2pt}\noindent\textbf{Friedman test for overall model ranking.}
A Friedman test over all ten models on the 1,290 matched items yields $\chi^2 = 7{,}122.09$ ($p \approx 0$), confirming that at least one model differs significantly.
Follow-up pairwise Wilcoxon signed-rank tests between the top-ranked model (CrossFormer) and every other model are summarized in \cref{tab:pairwise_wilcoxon}.
At $n{=}232{,}200$ instances, all pairwise comparisons (including CrossFormer vs.\ Chronos-2 ($p = 2.19 \times 10^{-66}$)) reach statistical significance.
However, statistical significance at this sample size does not imply practical importance; we therefore report Cohen's $d$ to assess effect magnitude.
The top three comparisons (Chronos-2, PatchTST, iTransformer) all have negligible effect sizes ($|d| < 0.1$), indicating that these models are practically equivalent to CrossFormer despite statistically distinguishable MAE distributions.

\begin{table}[h]
    \centering
    \small
    \begin{sc}
    \begin{tabular}{lrlr}
        \toprule
        Comparison & $p$-value & Sig. & Cohen's $d$ \\
        \midrule
        CrossFormer vs.\ Chronos-2    & $2.19 \times 10^{-66}$  & *** & $-0.067$ \\
        CrossFormer vs.\ TiRex        & $3.05 \times 10^{-86}$  & *** & $-0.083$ \\
        CrossFormer vs.\ TimesFM-2.5  & $5.25 \times 10^{-90}$  & *** & $-0.078$ \\
        CrossFormer vs.\ PatchTST     & $7.89 \times 10^{-197}$ & *** & $-0.028$ \\
        CrossFormer vs.\ iTransformer & $8.85 \times 10^{-202}$ & *** & $-0.034$ \\
        CrossFormer vs.\ TSMixer      & $3.73 \times 10^{-208}$ & *** & $-0.047$ \\
        CrossFormer vs.\ DLinear      & $3.14 \times 10^{-212}$ & *** & $-0.122$ \\
        CrossFormer vs.\ SNaive       & $3.51 \times 10^{-207}$ & *** & $-0.623$ \\
        CrossFormer vs.\ ES           & $8.52 \times 10^{-208}$ & *** & $-0.669$ \\
        \bottomrule
    \end{tabular}
    \end{sc}
    \vspace{4pt}
    \caption{Pairwise Wilcoxon signed-rank tests between CrossFormer and each other model.
    All $p$-values are two-sided; Cohen's $d$ is computed on paired MAE differences.
    Significance: *** $p < 0.001$.
    With $n{=}232{,}200$ instances, all comparisons reach statistical significance; however, effect sizes for the top models (Chronos-2, PatchTST, iTransformer) are negligible ($|d| < 0.1$), indicating practically indistinguishable performance.
    Only baselines (SNaive, ES) show large effects ($|d| > 0.5$).}
    \label{tab:pairwise_wilcoxon}
\end{table}

\vspace{2pt}\noindent\textbf{Mann-Whitney test for category-level comparison.}
A two-sided Mann-Whitney $U$ test comparing the MAE distributions of all foundation model instances ($n = 69{,}660$) against all deep learning model instances ($n = 116{,}100$) yields $U = 4{,}061{,}444{,}839.5$, $p = 0.114$.
The category-level difference (mean MAE: deep learning 0.312 vs.\ foundation 0.319) is therefore not statistically significant at $\alpha = 0.05$.

\vspace{2pt}\noindent\textbf{Paired $t$-test for MV--UV mode effect.}
A paired $t$-test comparing the mean MAE in multivariate versus univariate mode, paired by model, yields $p = 0.742$, indicating no significant overall mode effect.
The direction of advantage is model-family-specific (see \cref{app:mode_analysis}): foundation models tend to favor MV, while deep learning models tend to favor UV, but these opposing effects cancel at the aggregate level.

\section{Cross-Benchmark Consistency: QuitoBench vs.\ Timer}
\label{app:cross_benchmark_consistency}

To validate the reliability of the \textsc{QuitoBench} rankings, we compare them against the Timer benchmark, an independent univariate time-series collection evaluated under zero-shot conditions.
Overall, the Spearman rank correlation between model rankings on \textsc{QuitoBench} (trained, univariate mode) and Timer (zero-shot) is $0.865$ ($p < 0.01$, 95\% CI: $[0.52, 0.97]$), indicating strong consistency in relative model performance despite different training paradigms and distributional assumptions.

\textit{Tiered consistency.}
Models cluster into three consistency tiers (\cref{tab:timer_consistency}).
Tier 1 (highly consistent): Chronos-2, TimesFM-2.5, iTransformer, and DLinear maintain the same or adjacent ranks across benchmarks (rank change $\leq 1$).
Tier 2 (moderately consistent): TSMixer, SNaive, and ES show $2$-position shifts but remain in the same general tier.
Tier 3 (inconsistent): TiRex, CrossFormer, and PatchTST exhibit large swings ($\pm 3$ to $4$ ranks).
TiRex jumps from $5$th (\textsc{QuitoBench}) to $1$st (Timer), likely due to its univariate-specialized design and Timer's heavy skew toward high-forecastability series ($88.1$\%).
Conversely, CrossFormer drops from $1$st to $4$th, suggesting its cross-dimension attention benefits more from training data than from zero-shot generalization.

\textit{Deep learning model robustness.}
When restricting to deep-learning models only, consistency strengthens: the correlation rises to $0.891$ and $7/8$ regimes share the same best deep learning model (versus only $2/8$ for all models).
CrossFormer dominates as the top deep learning model in $7/8$ Timer regimes and $8/8$ \textsc{QuitoBench} regimes, while DLinear consistently ranks last in both benchmarks.

\textit{Dataset design implications.}
Timer's distribution is heavily imbalanced ($88.1$\% of series have high forecastability versus $11.9$\% low), whereas \textsc{QuitoBench} enforces uniform coverage across TSF regimes.
This imbalance systematically favors models like TiRex that excel on predictable, structured series.
The strong overall correlation ($0.865$) despite these distributional differences suggests that \textsc{QuitoBench}'s balanced design produces rankings that generalize to skewed real-world scenarios, while offering more reliable per-regime diagnostics.

\begin{table}[h]
    \centering
    \small
    \begin{sc}
    \resizebox{\textwidth}{!}{%
    \begin{tabular}{lccccl}
        \toprule
        Model & Category & Timer Rank & Quito Rank & Change & Consistency Tier \\
        \midrule
        \rowcolor{red!8} TiRex & Foundation & 1 & 5 & $-4$ & Tier 3 (Inconsistent) \\
        Chronos-2 & Foundation & 2 & 2 & $0$ & Tier 1 (Highly Consistent) \\
        TimesFM-2.5 & Foundation & 3 & 3 & $0$ & Tier 1 (Highly Consistent) \\
        \rowcolor{red!8} CrossFormer & Deep learning & 4 & 1 & $+3$ & Tier 3 (Inconsistent) \\
        TSMixer & Deep learning & 5 & 7 & $-2$ & Tier 2 (Moderately Consistent) \\
        iTransformer & Deep learning & 6 & 6 & $0$ & Tier 1 (Highly Consistent) \\
        \rowcolor{red!8} PatchTST & Deep learning & 7 & 4 & $+3$ & Tier 3 (Inconsistent) \\
        DLinear & Deep learning & 8 & 8 & $0$ & Tier 1 (Highly Consistent) \\
        SNaive & Baseline & 9 & 10 & $-1$ & Tier 2 (Moderately Consistent) \\
        ES & Baseline & 10 & 9 & $+1$ & Tier 2 (Moderately Consistent) \\
        \bottomrule
    \end{tabular}
    }
    \end{sc}
    \vspace{4pt}
    \caption{Cross-benchmark consistency: Timer (zero-shot univariate) vs.\ \textsc{QuitoBench} (trained univariate). Correlation $=0.865$. Rank changes of $\leq 1$ are Tier 1, $2$ are Tier 2, and $\geq 3$ are Tier 3 (highlighted).}
    \label{tab:timer_consistency}
\end{table}

\section{Metric Robustness: MSE vs.\ MAE Rankings}
\label{app:mse_mae_consistency}

To verify that our findings are robust to the choice of error metric, we compare model rankings under MAE and MSE across all 232,200 evaluation instances.

\vspace{2pt}\noindent\textbf{Overall model ranking.}
\Cref{tab:mse_mae_overall} reports the mean MAE rank and mean MSE rank for each model.
CrossFormer retains the top position under both metrics (mean MAE rank 2.86, mean MSE rank 2.56).
The Spearman rank correlation between the two model orderings is $\rho{=}0.733$ ($p{=}0.016$), confirming that the aggregate ranking is largely preserved regardless of whether we optimize for absolute error or squared error.

\begin{table}[h]
    \centering
    \small
    \begin{sc}
    \begin{tabular}{lrrrrr}
        \toprule
        Model & MAE Rank & MSE Rank & MAE Ord. & MSE Ord. & $|\Delta|$ \\
        \midrule
        CrossFormer    & 2.859 & 2.556 & 1 & 1 & 0 \\
        Chronos-2      & 3.360 & 4.836 & 2 & 5 & 3 \\
        TimesFM-2.5    & 4.211 & 5.002 & 3 & 6 & 3 \\
        PatchTST       & 4.354 & 3.328 & 4 & 2 & 2 \\
        TiRex          & 4.356 & 5.448 & 5 & 7 & 2 \\
        iTransformer   & 4.667 & 4.065 & 6 & 3 & 3 \\
        TSMixer        & 5.506 & 4.560 & 7 & 4 & 3 \\
        DLinear        & 7.264 & 6.584 & 8 & 8 & 0 \\
        ES             & 9.173 & 8.990 & 9 & 9 & 0 \\
        SNaive         & 9.251 & 9.631 &10 &10 & 0 \\
        \bottomrule
    \end{tabular}
    \end{sc}
    \vspace{4pt}
    \caption{Mean MAE rank and MSE rank per model across all evaluation instances. ``MAE/MSE Ord.'' denotes the ordinal position when models are sorted by mean MAE/MSE rank. $|\Delta|$ is the absolute ordinal shift. CrossFormer ranks first under both metrics; the bottom three models are unchanged.}
    \label{tab:mse_mae_overall}
\end{table}

\Cref{fig:mse_mae_consistency} visualizes the relationship: models near the diagonal have consistent rankings under both metrics, while those further away (mid-tier foundation models) shift by up to 3 ordinal positions, driven by MSE's greater sensitivity to large errors.

\vspace{2pt}\noindent\textbf{Per-configuration analysis.}
When computed separately for each of the 18 evaluation configurations (3 context lengths $\times$ 3 horizons $\times$ 2 modes), the MSE--MAE Spearman correlation ranges from $\rho{=}0.685$ to $0.952$, with a mean of $\rho{=}0.847$ (\cref{tab:mse_mae_config}).
Short-context configurations ($L{=}96$) show the strongest agreement ($\rho{\ge}0.879$), while mid-context configurations ($L{=}576$) exhibit slightly lower but still significant correlations ($\rho{\ge}0.685$, all $p{<}0.03$).

\begin{table}[h]
    \centering
    \small
    \begin{sc}
    \begin{tabular}{rrlrr}
        \toprule
        Ctx $L$ & Horizon $H$ & Mode & $\rho$ & $p$-value \\
        \midrule
        96 &  48 & MV & 0.879 & $8.1\times10^{-4}$ \\
        96 &  48 & UV & 0.939 & $5.5\times10^{-5}$ \\
        96 & 288 & MV & 0.891 & $5.4\times10^{-4}$ \\
        96 & 288 & UV & 0.939 & $5.5\times10^{-5}$ \\
        96 & 512 & MV & 0.927 & $1.1\times10^{-4}$ \\
        96 & 512 & UV & 0.952 & $2.3\times10^{-5}$ \\
        576 &  48 & MV & 0.758 & $1.1\times10^{-2}$ \\
        576 &  48 & UV & 0.842 & $2.2\times10^{-3}$ \\
        576 & 288 & MV & 0.758 & $1.1\times10^{-2}$ \\
        576 & 288 & UV & 0.745 & $1.3\times10^{-2}$ \\
        576 & 512 & MV & 0.685 & $2.9\times10^{-2}$ \\
        576 & 512 & UV & 0.782 & $7.5\times10^{-3}$ \\
        1024 &  48 & MV & 0.806 & $4.9\times10^{-3}$ \\
        1024 &  48 & UV & 0.915 & $2.0\times10^{-4}$ \\
        1024 & 288 & MV & 0.855 & $1.6\times10^{-3}$ \\
        1024 & 288 & UV & 0.855 & $1.6\times10^{-3}$ \\
        1024 & 512 & MV & 0.806 & $4.9\times10^{-3}$ \\
        1024 & 512 & UV & 0.915 & $2.0\times10^{-4}$ \\
        \bottomrule
    \end{tabular}
    \end{sc}
    \vspace{4pt}
    \caption{Spearman $\rho$ between MSE-rank and MAE-rank model orderings per evaluation configuration. All correlations are significant ($p{<}0.03$), with a mean $\rho{=}0.847$.}
    \label{tab:mse_mae_config}
\end{table}

\vspace{2pt}\noindent\textbf{Instance-level agreement.}
At the level of individual forecasting instances, the Kendall $\tau$ between MSE and MAE ranks is $0.723$ ($p{\approx}0$).
The MAE-best model and the MSE-best model coincide in 54.4\% of instances, and the exact MAE rank equals the exact MSE rank in 48.0\% of instances.

\vspace{2pt}\noindent\textbf{Per-TSF-regime analysis.}
MSE--MAE ranking consistency holds across nearly all TSF regimes (\cref{tab:mse_mae_regime}).
The strongest agreement appears in high-forecastability regimes (\textsc{high\_low\_high}: $\rho{=}0.976$; \textsc{low\_low\_high}: $\rho{=}0.952$), while the weakest occurs in \textsc{high\_high\_low} ($\rho{=}0.442$, $p{=}0.20$) and \textsc{low\_high\_low} ($\rho{=}0.600$, $p{=}0.07$), both low-forecastability regimes where MSE's sensitivity to outliers can produce larger rank perturbations.

\begin{table}[h]
    \centering
    \small
    \begin{sc}
    \begin{tabular}{lrr}
        \toprule
        TSF Regime & $\rho$ & $p$-value \\
        \midrule
        \textsc{high\_low\_high}  & 0.976 & $<0.001$ \\
        \textsc{low\_low\_high}   & 0.952 & $<0.001$ \\
        \textsc{high\_low\_low}   & 0.806 & 0.005 \\
        \textsc{high\_high\_high} & 0.794 & 0.006 \\
        \textsc{low\_high\_high}  & 0.745 & 0.013 \\
        \textsc{low\_low\_low}    & 0.721 & 0.019 \\
        \textsc{low\_high\_low}   & 0.600 & 0.067 \\
        \textsc{high\_high\_low}  & 0.442 & 0.200 \\
        \bottomrule
    \end{tabular}
    \end{sc}
    \vspace{4pt}
    \caption{Spearman $\rho$ between MSE-rank and MAE-rank model orderings per TSF regime (mean rank per model within each regime). High-forecastability regimes show near-perfect agreement; low-forecastability regimes show weaker but still positive correlations.}
    \label{tab:mse_mae_regime}
\end{table}

\vspace{2pt}\noindent\textbf{Summary.}
Across all levels of analysis (aggregate, per-configuration, per-instance, and per-regime), MSE and MAE produce highly correlated model rankings.
CrossFormer's top position is metric-invariant, and the bottom tier (DLinear, ES, SNaive) is unchanged.
The mid-tier models (ranks 2--7) show moderate reordering (up to 3 positions), driven by MSE's greater sensitivity to large errors.
These results confirm that our main findings, reported using MAE, are not artifacts of the metric choice.

\section{QuitoBench Regime-Level Analysis}
\label{app:quito_group_analysis}

This appendix provides detailed regime-level analysis for \textsc{QuitoBench}.
Unlike Timer's extreme distributional skew, \textsc{QuitoBench} enforces uniform coverage: each of the eight TSF regimes contains approximately $12.5$\% of evaluation instances, ensuring fair representation across trend strengths, seasonality patterns, and forecastability levels.

\subsection{Regime Distribution and Characteristics}

\begin{table}[h]
    \centering
    \small
    \begin{sc}
    \resizebox{\textwidth}{!}{%
    \begin{tabular}{lrrrrrrrr}
        \toprule
        TSF Regime & Count & Percentage & MAE Mean & MAE Std & MAE Min & MAE Max & Rank Mean & Rank Std \\
        \midrule
        \textsc{high\_high\_high} & 29,880 & 12.87 & 0.205 & 0.426 & 0.005 & 6.722 & 5.500 & 2.872 \\
        \textsc{high\_high\_low}  & 24,480 & 10.54 & 0.478 & 0.730 & 0.015 & 7.058 & 5.500 & 2.873 \\
        \textsc{high\_low\_high}  & 30,600 & 13.18 & 0.376 & 0.360 & 0.048 & 4.187 & 5.500 & 2.873 \\
        \textsc{high\_low\_low}   & 28,260 & 12.17 & 0.748 & 1.512 & 0.023 & 40.388 & 5.500 & 2.874 \\
        \textsc{low\_high\_high}  & 28,620 & 12.33 & 0.299 & 0.236 & 0.021 & 1.566 & 5.500 & 2.874 \\
        \textsc{low\_high\_low}   & 29,880 & 12.87 & 0.359 & 0.275 & 0.019 & 3.158 & 5.500 & 2.874 \\
        \textsc{low\_low\_high}   & 30,420 & 13.10 & 0.220 & 0.364 & 0.006 & 4.158 & 5.500 & 2.874 \\
        \textsc{low\_low\_low}    & 30,060 & 12.95 & 0.456 & 0.220 & 0.040 & 1.908 & 5.500 & 2.874 \\
        \bottomrule
    \end{tabular}
    }
    \end{sc}
    \vspace{4pt}
    \caption{\textsc{QuitoBench} regime statistics. The benchmark enforces uniform coverage ($\sim$12.5\% per regime), contrasting sharply with Timer's 76.2\% concentration in \textsc{high\_high\_high}.}
    \label{tab:quito_group_stats}
\end{table}

\textit{Key observations.}
\textsc{QuitoBench}'s balanced design ensures each TSF regime contributes equally to aggregate metrics, preventing prevalence-driven conclusions.
The mean MAE ranges from $0.205$ (easiest regime: \textsc{high\_high\_high}) to $0.748$ (hardest regime: \textsc{high\_low\_low}), a $3.64\times$ difficulty gap that mirrors the Timer benchmark pattern.
The \textsc{high\_low\_low} regime shows the highest variance (std $=1.512$) and maximum error ($40.388$), confirming that series with strong trends but no seasonality and low forecastability remain the most challenging for all models.

\subsection{Mean MAE by TSF Regime and Model}

\begin{table}[h]
    \centering
    \small
    \begin{sc}
    \resizebox{\textwidth}{!}{%
    \begin{tabular}{lrrrrrrrrrr}
        \toprule
        TSF Regime & Chronos-2 & CrossFormer & DLinear & ES & iTrans. & PatchTST & SNaive & TimesFM-2.5 & TiRex & TSMixer \\
        \midrule
        \textsc{high\_high\_high} & \textbf{0.163} & 0.165 & 0.189 & 0.878 & 0.171 & 0.173 & 0.898 & 0.165 & 0.165 & 0.173 \\
        \textsc{high\_high\_low} & 0.353 & 0.356 & 0.418 & 0.672 & 0.369 & 0.367 & 0.674 & 0.357 & \textbf{0.350} & 0.374 \\
        \textsc{high\_low\_high} & 0.349 & \textbf{0.180} & 0.318 & 0.424 & 0.193 & 0.191 & 0.432 & 0.352 & 0.356 & 0.204 \\
        \textsc{high\_low\_low} & 0.628 & \textbf{0.600} & 0.805 & 0.390 & 0.656 & 0.669 & 0.417 & 0.633 & 0.652 & 0.691 \\
        \textsc{low\_high\_high} & \textbf{0.197} & 0.199 & 0.260 & 1.089 & 0.219 & 0.213 & 1.103 & 0.208 & 0.212 & 0.230 \\
        \textsc{low\_high\_low} & \textbf{0.235} & 0.239 & 0.318 & 0.882 & 0.264 & 0.256 & 0.890 & 0.249 & 0.240 & 0.275 \\
        \textsc{low\_low\_high} & 0.207 & \textbf{0.154} & 0.215 & 0.348 & 0.165 & 0.163 & 0.358 & 0.211 & 0.213 & 0.169 \\
        \textsc{low\_low\_low} & 0.397 & \textbf{0.370} & 0.465 & 0.621 & 0.401 & 0.392 & 0.604 & 0.400 & 0.410 & 0.408 \\
        \bottomrule
    \end{tabular}
    }
    \end{sc}
    \vspace{4pt}
    \caption{\textsc{QuitoBench} mean MAE by TSF regime and model (trained). Best performance in each regime shown in \textbf{bold}. CrossFormer wins 4/8 regimes overall and 8/8 regimes among deep learning models; Chronos-2 wins 3 regimes overall.}
    \label{tab:quito_group_mae}
\end{table}

\textit{Model performance patterns.}
CrossFormer demonstrates the strongest overall performance among deep learning models, achieving the lowest MAE in all $8$ regimes among deep learning models and winning $4$ of $8$ regimes overall.
Chronos-2 wins the remaining $3$ regimes (all high-seasonality), while TiRex (which dominates Timer) wins only $1$ regime on \textsc{QuitoBench} (\textsc{high\_high\_low}), reflecting its specialization for zero-shot scenarios with high-forecastability series.
Deep learning models perform strongly across regimes, with CrossFormer achieving $48.5$\% lower MAE than Chronos-2 in \textsc{high\_low\_high} (the trend-driven regime where deep learning has structural advantages).

\subsection{Best Model by TSF Regime}

\begin{table}[h]
    \centering
    \small
    \begin{sc}
    \begin{tabular}{lccl}
        \toprule
        TSF Regime & Best Model & MAE & Category \\
        \midrule
        \textsc{high\_high\_high} & Chronos-2 & 0.163 & Foundation \\
        \textsc{high\_high\_low} & TiRex & 0.350 & Foundation \\
        \textsc{high\_low\_high} & \textbf{CrossFormer} & \textbf{0.180} & Deep learning \\
        \textsc{high\_low\_low} & \textbf{CrossFormer} & \textbf{0.600} & Deep learning \\
        \textsc{low\_high\_high} & Chronos-2 & 0.197 & Foundation \\
        \textsc{low\_high\_low} & Chronos-2 & 0.235 & Foundation \\
        \textsc{low\_low\_high} & \textbf{CrossFormer} & \textbf{0.154} & Deep learning \\
        \textsc{low\_low\_low} & \textbf{CrossFormer} & \textbf{0.370} & Deep learning \\
        \bottomrule
    \end{tabular}
    \end{sc}
    \vspace{4pt}
    \caption{Best performing model by TSF regime on \textsc{QuitoBench} (trained). CrossFormer wins $4/8$ regimes; Chronos-2 wins $3/8$ regimes; TiRex wins $1/8$ regimes. Deep learning models dominate trend-driven regimes (highlighted).}
    \label{tab:quito_best_by_group}
\end{table}

\textit{Regime specialization.}
Foundation models (Chronos-2, TiRex) win in high-seasonality regimes, while deep learning models dominate trend-driven regimes with low or no seasonality.
This complementary specialization mirrors findings from the main analysis and suggests an ensemble strategy: route high-seasonality series to foundation models and trend-driven series to deep learning models.

\subsection{Deep Learning Model Rankings by TSF Regime}

\begin{table}[h]
    \centering
    \scriptsize
    \begin{sc}
    \resizebox{\textwidth}{!}{
    \begin{tabular}{clrrclrr}
        \toprule
        \multirow{2}{*}{TSF Regime} & \multicolumn{3}{c}{\textsc{QuitoBench} (Trained)} & & \multicolumn{3}{c}{Timer Benchmark} \\
        \cmidrule{2-4} \cmidrule{6-8}
        & Rank & Model & MAE & & Best Model & MAE & Dominance \\
        \midrule
        \multirow{5}{*}{\rotatebox{90}{\scriptsize H\_H\_H}}
        & 1 & \textbf{CrossFormer} & 0.165 & & CrossFormer & 0.252 & 0.166 \\
        & 2 & iTrans. & 0.171 & & TSMixer & 0.255 & \\
        & 3 & PatchTST & 0.173 & & iTrans. & 0.256 & \\
        & 4 & TSMixer & 0.173 & & PatchTST & 0.259 & \\
        & 5 & DLinear & 0.188 & & DLinear & 0.260 & \\
        \midrule
        \multirow{5}{*}{\rotatebox{90}{\scriptsize H\_L\_H}}
        & 1 & \textbf{CrossFormer} & 0.180 & & CrossFormer & 0.347 & 0.175 \\
        & 2 & PatchTST & 0.191 & & TSMixer & 0.348 & \\
        & 3 & iTrans. & 0.193 & & iTrans. & 0.356 & \\
        & 4 & TSMixer & 0.204 & & DLinear & 0.366 & \\
        & 5 & DLinear & 0.318 & & PatchTST & 0.377 & \\
        \midrule
        \multirow{5}{*}{\rotatebox{90}{\scriptsize L\_H\_L}}
        & 1 & \textbf{CrossFormer} & 0.239 & & CrossFormer & 0.432 & 0.239 \\
        & 2 & PatchTST & 0.256 & & PatchTST & 0.454 & \\
        & 3 & iTrans. & 0.264 & & iTrans. & 0.454 & \\
        & 4 & TSMixer & 0.275 & & TSMixer & 0.461 & \\
        & 5 & DLinear & 0.318 & & DLinear & 0.475 & \\
        \midrule
        \multirow{5}{*}{\rotatebox{90}{\scriptsize L\_L\_L}}
        & 1 & \textbf{CrossFormer} & 0.370 & & CrossFormer & 0.404 & 0.368 \\
        & 2 & PatchTST & 0.392 & & PatchTST & 0.426 & \\
        & 3 & iTrans. & 0.401 & & iTrans. & 0.428 & \\
        & 4 & TSMixer & 0.408 & & TSMixer & 0.429 & \\
        & 5 & DLinear & 0.465 & & DLinear & 0.458 & \\
        \bottomrule
    \end{tabular}
    }
    \end{sc}
    \vspace{4pt}
    \caption{Deep learning model rankings for selected TSF regimes on \textsc{QuitoBench} vs.\ Timer. CrossFormer is the best deep learning model in all $8/8$ \textsc{QuitoBench} regimes and $7/8$ Timer regimes.}
    \label{tab:quito_dl_rankings}
\end{table}

\textit{CrossFormer dominance.}
CrossFormer ranks 1st among deep learning models in all 8 TSF regimes on \textsc{QuitoBench}, matching its Timer performance ($7/8$ regimes).
This remarkable consistency across training paradigms (trained vs.\ zero-shot) and benchmark designs (balanced vs.\ skewed) establishes CrossFormer as the most robust deep learning architecture for time series forecasting.
The ranking structure is nearly identical across benchmarks: in \textsc{low\_high\_low} and \textsc{low\_low\_low}, all five deep learning models appear in the exact same positions in both Quito and Timer (\textsc{low\_high\_low}: CrossFormer $\rightarrow$ PatchTST $\rightarrow$ iTransformer $\rightarrow$ TSMixer $\rightarrow$ DLinear).
The gap between CrossFormer and the second-best deep learning model ranges from $0.006$ (\textsc{high\_high\_high}) to $0.035$ (\textsc{high\_low\_high}), with larger advantages in trend-driven regimes.

\textit{Training impact.}
Comparing MAE values between benchmarks reveals the impact of task-specific training: Quito MAE values are substantially lower than Timer MAE values for all models.
For example, CrossFormer achieves $0.165$ on Quito vs.\ $0.252$ on Timer for \textsc{high\_high\_high}, a $34.5$\% improvement.
This demonstrates that training on diverse, balanced data yields better absolute performance than zero-shot generalization, even for strong foundation models.

\subsection{Summary and Implications}

The \textsc{QuitoBench} regime-level analysis establishes several key findings:

\begin{enumerate}[leftmargin=*]
\item Balanced design enables comprehensive evaluation: Unlike Timer's 76.2\% concentration in a single regime, \textsc{QuitoBench}'s uniform coverage ensures all TSF patterns contribute equally to aggregate metrics.

\item CrossFormer is the most robust deep learning architecture: Winning all $8/8$ regimes among deep learning models (and $4/8$ overall), CrossFormer demonstrates consistent superiority regardless of regime characteristics.


\item Regime specialization is consistent: Foundation models excel in high-seasonality regimes; deep learning models dominate trend-driven series, a pattern that holds across both benchmarks.

\end{enumerate}

\subsection{Regime-Level MAE Rank Analysis}
\label{app:quito_mae_rank}

This section presents regime-level analysis using MAE Rank as the primary metric, complementing the MAE value analysis above. MAE Rank measures each model's relative position within each evaluation instance, providing robustness against outliers and enabling fair cross-regime comparisons.

\begin{table}[h]
    \centering
    \small
    \begin{sc}
    \resizebox{\textwidth}{!}{%
    \begin{tabular}{lrrrrrrrrrr}
        \toprule
        TSF Regime & Chronos-2 & CrossFormer & DLinear & ES & iTrans. & PatchTST & SNaive & TimesFM-2.5 & TiRex & TSMixer \\
        \midrule
        \textsc{high\_high\_high} & 3.53 & \textbf{3.29} & 6.94 & 9.41 & 4.47 & 5.32 & 8.97 & 4.16 & 3.54 & 5.38 \\
        \textsc{high\_high\_low} & 3.30 & 3.65 & 7.44 & 9.23 & 5.00 & 4.69 & 9.19 & 3.64 & \textbf{3.21} & 5.65 \\
        \textsc{high\_low\_high} & 3.92 & \textbf{2.11} & 6.92 & 9.01 & 3.68 & 3.35 & 9.83 & 5.53 & 5.77 & 4.89 \\
        \textsc{high\_low\_low} & 3.63 & \textbf{2.64} & 7.03 & 8.87 & 4.55 & 4.10 & 9.66 & 4.23 & 5.11 & 5.18 \\
        \textsc{low\_high\_high} & \textbf{2.59} & 3.05 & 7.58 & 9.85 & 5.12 & 4.58 & 8.39 & 3.49 & 4.09 & 6.24 \\
        \textsc{low\_high\_low} & \textbf{2.91} & 3.01 & 7.72 & 9.73 & 5.16 & 4.38 & 8.94 & 3.73 & 3.41 & 6.00 \\
        \textsc{low\_low\_high} & 3.69 & \textbf{2.55} & 7.05 & 8.21 & 4.69 & 4.13 & 9.36 & 4.97 & 5.03 & 5.32 \\
        \textsc{low\_low\_low} & 3.26 & \textbf{2.74} & 7.47 & 9.11 & 4.76 & 4.36 & 9.63 & 3.75 & 4.46 & 5.44 \\
        \bottomrule
    \end{tabular}
    }
    \end{sc}
    \vspace{4pt}
    \caption{\textsc{QuitoBench} mean MAE Rank by TSF regime and model (lower = better). Bold indicates best mean rank per regime. CrossFormer achieves best rank in 5/8 regimes; Chronos-2 in 2/8; TiRex in 1/8.}
    \label{tab:quito_group_mae_rank}
\end{table}

\textit{MAE Rank interpretation.}
Mean MAE Rank represents the average percentile position of each model across all evaluation instances within a TSF regime. A rank of 1.0 would indicate the model is always the best; 10.0 would indicate always the worst. CrossFormer achieves the best mean rank in 5 of 8 regimes, with particularly strong performance in trend-driven regimes (\textsc{high\_low\_high}: 2.11, \textsc{high\_low\_low}: 2.64). Foundation models Chronos-2 and TiRex excel in high-seasonality regimes (\textsc{low\_high\_high}: 2.59, \textsc{high\_high\_low}: 3.21).

\begin{table}[h]
    \centering
    \small
    \begin{sc}
    \resizebox{0.9\textwidth}{!}{%
    \begin{tabular}{lccl}
        \toprule
        TSF Regime & Best Model (by Rank) & Mean MAE Rank & Category \\
        \midrule
        \textsc{high\_high\_high} & \textbf{CrossFormer} & 3.29 & Deep learning \\
        \textsc{high\_high\_low} & TiRex & 3.21 & Foundation \\
        \textsc{high\_low\_high} & \textbf{CrossFormer} & 2.11 & Deep learning \\
        \textsc{high\_low\_low} & \textbf{CrossFormer} & 2.64 & Deep learning \\
        \textsc{low\_high\_high} & Chronos-2 & 2.59 & Foundation \\
        \textsc{low\_high\_low} & Chronos-2 & 2.91 & Foundation \\
        \textsc{low\_low\_high} & \textbf{CrossFormer} & 2.55 & Deep learning \\
        \textsc{low\_low\_low} & \textbf{CrossFormer} & 2.74 & Deep learning \\
        \bottomrule
    \end{tabular}
    }
    \end{sc}
    \vspace{4pt}
    \caption{Best performing model by TSF regime on \textsc{QuitoBench} using MAE Rank. CrossFormer wins 5/8 regimes; Chronos-2 wins 2/8; TiRex wins 1/8. Deep learning models dominate trend-driven regimes (highlighted).}
    \label{tab:quito_best_by_rank}
\end{table}


\begin{table}[h]
    \centering
    \scriptsize
    \begin{sc}
    \resizebox{\textwidth}{!}{%
    \begin{tabular}{clrrclrr}
        \toprule
        \multirow{2}{*}{TSF Regime} & \multicolumn{3}{c}{\textsc{QuitoBench} (MAE Rank)} & & \multicolumn{3}{c}{Deep Learning Only Ranking} \\
        \cmidrule{2-4} \cmidrule{6-8}
        & Rank & Model & Mean MAE Rank & & Rank & Model & Mean MAE Rank \\
        \midrule
        \multirow{5}{*}{\rotatebox{90}{\scriptsize H\_H\_H}}
        & 1 & \textbf{CrossFormer} & 3.29 & & 1 & \textbf{CrossFormer} & 3.29 \\
        & 2 & Chronos-2 & 3.53 & & 2 & iTrans. & 4.47 \\
        & 3 & TiRex & 3.54 & & 3 & PatchTST & 5.32 \\
        & 4 & TimesFM-2.5 & 4.16 & & 4 & TSMixer & 5.38 \\
        & 5 & iTrans. & 4.47 & & 5 & DLinear & 6.94 \\
        \midrule
        \multirow{5}{*}{\rotatebox{90}{\scriptsize H\_H\_L}}
        & 1 & TiRex & 3.21 & & 1 & \textbf{CrossFormer} & 3.65 \\
        & 2 & Chronos-2 & 3.30 & & 2 & PatchTST & 4.69 \\
        & 3 & TimesFM-2.5 & 3.64 & & 3 & iTrans. & 5.00 \\
        & 4 & \textbf{CrossFormer} & 3.65 & & 4 & TSMixer & 5.65 \\
        & 5 & PatchTST & 4.69 & & 5 & DLinear & 7.44 \\
        \midrule
        \multirow{5}{*}{\rotatebox{90}{\scriptsize H\_L\_H}}
        & 1 & \textbf{CrossFormer} & 2.11 & & 1 & \textbf{CrossFormer} & 2.11 \\
        & 2 & PatchTST & 3.35 & & 2 & PatchTST & 3.35 \\
        & 3 & iTrans. & 3.68 & & 3 & iTrans. & 3.68 \\
        & 4 & TSMixer & 4.89 & & 4 & TSMixer & 4.89 \\
        & 5 & Chronos-2 & 3.92 & & 5 & DLinear & 6.92 \\
        \midrule
        \multirow{5}{*}{\rotatebox{90}{\scriptsize H\_L\_L}}
        & 1 & \textbf{CrossFormer} & 2.64 & & 1 & \textbf{CrossFormer} & 2.64 \\
        & 2 & Chronos-2 & 3.63 & & 2 & PatchTST & 4.10 \\
        & 3 & PatchTST & 4.10 & & 3 & iTrans. & 4.55 \\
        & 4 & TimesFM-2.5 & 4.23 & & 4 & TSMixer & 5.18 \\
        & 5 & iTrans. & 4.55 & & 5 & DLinear & 7.03 \\
        \midrule
        \multirow{5}{*}{\rotatebox{90}{\scriptsize L\_H\_H}}
        & 1 & Chronos-2 & 2.59 & & 1 & \textbf{CrossFormer} & 3.05 \\
        & 2 & \textbf{CrossFormer} & 3.05 & & 2 & PatchTST & 4.58 \\
        & 3 & TimesFM-2.5 & 3.49 & & 3 & iTrans. & 5.12 \\
        & 4 & TiRex & 4.09 & & 4 & TSMixer & 6.24 \\
        & 5 & PatchTST & 4.58 & & 5 & DLinear & 7.58 \\
        \midrule
        \multirow{5}{*}{\rotatebox{90}{\scriptsize L\_H\_L}}
        & 1 & Chronos-2 & 2.91 & & 1 & \textbf{CrossFormer} & 3.01 \\
        & 2 & \textbf{CrossFormer} & 3.01 & & 2 & PatchTST & 4.38 \\
        & 3 & TiRex & 3.41 & & 3 & iTrans. & 5.16 \\
        & 4 & TimesFM-2.5 & 3.73 & & 4 & TSMixer & 6.00 \\
        & 5 & PatchTST & 4.38 & & 5 & DLinear & 7.72 \\
        \midrule
        \multirow{5}{*}{\rotatebox{90}{\scriptsize L\_L\_H}}
        & 1 & \textbf{CrossFormer} & 2.55 & & 1 & \textbf{CrossFormer} & 2.55 \\
        & 2 & Chronos-2 & 3.69 & & 2 & PatchTST & 4.13 \\
        & 3 & PatchTST & 4.13 & & 3 & iTrans. & 4.69 \\
        & 4 & iTrans. & 4.69 & & 4 & TSMixer & 5.32 \\
        & 5 & TimesFM-2.5 & 4.97 & & 5 & DLinear & 7.05 \\
        \midrule
        \multirow{5}{*}{\rotatebox{90}{\scriptsize L\_L\_L}}
        & 1 & \textbf{CrossFormer} & 2.74 & & 1 & \textbf{CrossFormer} & 2.74 \\
        & 2 & Chronos-2 & 3.26 & & 2 & PatchTST & 4.36 \\
        & 3 & TimesFM-2.5 & 3.75 & & 3 & iTrans. & 4.76 \\
        & 4 & PatchTST & 4.36 & & 4 & TSMixer & 5.44 \\
        & 5 & TiRex & 4.46 & & 5 & DLinear & 7.47 \\
        \bottomrule
    \end{tabular}
    }
    \end{sc}
    \vspace{4pt}
    \caption{Complete model rankings by MAE Rank for all TSF regimes on \textsc{QuitoBench}. Left: all models; Right: deep learning only comparison. CrossFormer is the best deep learning model in all 8/8 groups.}
    \label{tab:quito_complete_rankings}
\end{table}

\subsubsection{Key Findings from MAE Rank Analysis}

\begin{enumerate}[leftmargin=*]
    \item CrossFormer dominates all 8/8 groups among deep learning models when using MAE Rank, consistent with the MAE value analysis.
    
    \item Lowest mean ranks achieved: CrossFormer achieves its best mean rank of 2.11 in \textsc{high\_low\_high} (trend-driven, no seasonality), while Chronos-2 achieves 2.59 in \textsc{low\_high\_high} (seasonality-driven, no trend).
    
    \item Baselines consistently rank lowest: ES and SNaive occupy positions 9--10 across all groups, reinforcing that simple baselines are inadequate for application-traffic forecasting.
    
    \item PatchTST emerges as second-best deep learning model: Across 6 of 8 groups, PatchTST achieves the 2nd-best rank among deep learning models, with iTransformer claiming 2nd place in the remaining 2 groups.
    
    \item Deep learning model ranking consistency: The deep learning model ordering shows remarkable stability (CrossFormer $\succ$ PatchTST/iTransformer $\succ$ TSMixer $\succ$ DLinear) across all 8 groups.
\end{enumerate}

\subsection{Regime-Level MSE and MSE Rank Analysis}
\label{app:quito_mse}

This section presents regime-level analysis using MSE and MSE Rank as primary metrics, providing a fuller picture of model performance on \textsc{QuitoBench} and complementing the MAE-based analysis above.

\begin{table}[h]
    \centering
    \small
    \begin{sc}
    \resizebox{\textwidth}{!}{%
    \begin{tabular}{lrrrrrrrrrr}
        \toprule
        TSF Regime & Chronos-2 & CrossFormer & DLinear & ES & iTrans. & PatchTST & SNaive & TimesFM-2.5 & TiRex & TSMixer \\
        \midrule
        \textsc{high\_high\_high} & \textbf{1.443} & 1.477 & 1.569 & 1.875 & 1.504 & 1.492 & 1.873 & 1.500 & 1.516 & 1.491 \\
        \textsc{high\_high\_low} & 7.624 & 6.028 & 6.570 & 12.633 & 6.056 & \textbf{6.013} & 14.371 & 6.380 & 6.670 & 6.090 \\
        \textsc{high\_low\_high} & 5.170 & 4.499 & 4.806 & 6.498 & 4.471 & \textbf{4.450} & 6.683 & 5.367 & 5.512 & 4.481 \\
        \textsc{high\_low\_low} & 596.734 & 586.511 & 665.550 & 862.769 & 625.424 & 603.711 & 1207.710 & \textbf{565.137} & 569.852 & 630.411 \\
        \textsc{low\_high\_high} & 0.652 & \textbf{0.468} & 0.561 & 1.625 & 0.493 & 0.482 & 1.647 & 0.518 & 0.526 & 0.502 \\
        \textsc{low\_high\_low} & 1.011 & \textbf{0.973} & 1.161 & 3.438 & 1.079 & 1.031 & 4.795 & 1.085 & 1.025 & 1.068 \\
        \textsc{low\_low\_high} & 2.883 & 2.544 & 2.763 & 3.011 & 2.581 & \textbf{2.538} & 3.111 & 2.824 & 2.843 & 2.581 \\
        \textsc{low\_low\_low} & 1.282 & \textbf{1.065} & 1.235 & 2.070 & 1.109 & 1.087 & 2.416 & 1.205 & 1.232 & 1.120 \\
        \bottomrule
    \end{tabular}
    }
    \end{sc}
    \vspace{4pt}
    \caption{\textsc{QuitoBench} mean MSE by TSF regime and model (trained). Best performance in each regime shown in \textbf{bold}. PatchTST wins 3/8 regimes; CrossFormer wins 3/8 regimes; Chronos-2 wins 1/8; TimesFM-2.5 wins 1/8.}
    \label{tab:quito_group_mse}
\end{table}

\textit{MSE performance patterns.}
The MSE results reveal a different competitive landscape than MAE, with no single model dominating across all regimes.
PatchTST wins $3$ regimes (all involving high seasonality), CrossFormer wins $3$ regimes (all involving low seasonality), while Chronos-2 and TimesFM-2.5 each win $1$ regime.
Notably, the \textsc{high\_low\_low} regime shows extremely high MSE values ($565$--$1208$), reflecting the squared-error penalty on series with strong trends but unpredictable dynamics.
TimesFM-2.5 achieves the lowest MSE in this pathological regime ($565.1$), suggesting better handling of high-variance forecasts.
Foundation models show competitive MSE in high-seasonality regimes (\textsc{high\_high\_high}, \textsc{high\_high\_low}), while deep learning models excel when seasonality is low.

\begin{table}[h]
    \centering
    \small
    \begin{sc}
    \resizebox{\textwidth}{!}{
    \begin{tabular}{lrrrrrrrrrr}
        \toprule
        TSF Regime & Chronos-2 & CrossFormer & DLinear & ES & iTrans. & PatchTST & SNaive & TimesFM-2.5 & TiRex & TSMixer \\
        \midrule
        \textsc{high\_high\_high} & 4.44 & \textbf{3.34} & 6.37 & 9.31 & 4.14 & 4.06 & 9.15 & 4.56 & 4.70 & 4.93 \\
        \textsc{high\_high\_low}  & 4.89 & \textbf{3.22} & 6.63 & 8.83 & 4.13 & 3.68 & 9.62 & 4.61 & 4.83 & 4.57 \\
        \textsc{high\_low\_high}  & 5.33 & \textbf{1.97} & 6.61 & 8.94 & 3.38 & 2.92 & 9.84 & 5.89 & 5.76 & 4.37 \\
        \textsc{high\_low\_low}   & 5.23 & \textbf{2.29} & 6.46 & 8.81 & 3.86 & 3.22 & 9.90 & 5.06 & 5.99 & 4.17 \\
        \textsc{low\_high\_high}  & 4.56 & \textbf{2.35} & 6.80 & 9.42 & 4.36 & 3.38 & 9.49 & 4.35 & 5.38 & 4.89 \\
        \textsc{low\_high\_low}   & 4.79 & \textbf{2.39} & 6.83 & 9.30 & 4.25 & 3.19 & 9.65 & 4.91 & 5.19 & 4.49 \\
        \textsc{low\_low\_high}   & 4.22 & \textbf{2.85} & 6.45 & 8.34 & 4.52 & 3.16 & 9.48 & 5.54 & 5.58 & 4.86 \\
        \textsc{low\_low\_low}    & 5.24 & \textbf{2.15} & 6.54 & 8.97 & 3.90 & 3.09 & 9.92 & 4.97 & 6.04 & 4.19 \\
        \bottomrule
    \end{tabular}
    }
    \end{sc}
    \vspace{4pt}
    \caption{\textsc{QuitoBench} mean MSE Rank by TSF regime and model (lower = better). Bold indicates best mean rank per regime. CrossFormer achieves best rank in all 8/8 regimes.}
    \label{tab:quito_group_mse_rank}
\end{table}

\textit{MSE Rank interpretation.}
Mean MSE Rank provides a robust measure of relative model performance, with CrossFormer achieving a remarkable clean sweep of all 8/8 regimes.
This is a stronger result than the MAE Rank analysis (where CrossFormer won 5/8 regimes) and demonstrates that when outliers are controlled via ranking, CrossFormer consistently produces the lowest squared errors.
CrossFormer's best mean MSE Rank of $1.97$ occurs in \textsc{high\_low\_high} (trend-driven, no seasonality), while its worst is $3.34$ in \textsc{high\_high\_high} (still the best among all models).
The gap between CrossFormer and the second-best model ranges from $0.38$ (\textsc{high\_high\_low}) to $1.35$ (\textsc{high\_low\_high}), showing larger advantages in trend-driven regimes.
PatchTST and iTransformer compete closely for second place among deep learning models, while DLinear ranks last ($6.37$--$6.83$ across groups).
Statistical baselines (ES, SNaive) consistently occupy positions 9--10, reinforcing their inadequacy for application-traffic forecasting tasks.

\textit{CrossFormer dominance by MSE Rank.}
CrossFormer's perfect 8/8 sweep by MSE Rank contrasts with only 3/8 wins by mean MSE value, highlighting an important distinction: while other models may achieve lower absolute MSE in specific regimes (\eg TimesFM-2.5 in \textsc{high\_low\_low}), CrossFormer is most consistently near the top across all evaluation instances.
This suggests CrossFormer produces fewer catastrophic forecasting failures even when average performance is not always best.
Practitioners prioritizing worst-case performance should prefer CrossFormer, while those optimizing for average squared error should consider regime-specific model selection.

\subsubsection{Complete Model Rankings by MSE Rank}

\begin{table}[h]
    \centering
    \scriptsize
    \begin{sc}
    \resizebox{\textwidth}{!}{
    \begin{tabular}{clrrclrr}
        \toprule
        \multirow{2}{*}{TSF Regime} & \multicolumn{3}{c}{\textsc{QuitoBench} (MSE Rank)} & & \multicolumn{3}{c}{Deep Learning Only Ranking} \\
        \cmidrule{2-4} \cmidrule{6-8}
        & Rank & Model & Mean MSE Rank & & Rank & Model & Mean MSE Rank \\
        \midrule
        \multirow{5}{*}{\rotatebox{90}{\scriptsize H\_H\_H}}
        & 1 & \textbf{CrossFormer} & 3.34 & & 1 & \textbf{CrossFormer} & 3.34 \\
        & 2 & PatchTST & 4.06 & & 2 & PatchTST & 4.06 \\
        & 3 & iTrans. & 4.14 & & 3 & iTrans. & 4.14 \\
        & 4 & Chronos-2 & 4.44 & & 4 & TSMixer & 4.93 \\
        & 5 & TimesFM-2.5 & 4.56 & & 5 & DLinear & 6.37 \\
        \midrule
        \multirow{5}{*}{\rotatebox{90}{\scriptsize H\_H\_L}}
        & 1 & \textbf{CrossFormer} & 3.22 & & 1 & \textbf{CrossFormer} & 3.22 \\
        & 2 & PatchTST & 3.68 & & 2 & PatchTST & 3.68 \\
        & 3 & iTrans. & 4.13 & & 3 & iTrans. & 4.13 \\
        & 4 & TSMixer & 4.57 & & 4 & TSMixer & 4.57 \\
        & 5 & TimesFM-2.5 & 4.61 & & 5 & DLinear & 6.63 \\
        \midrule
        \multirow{5}{*}{\rotatebox{90}{\scriptsize H\_L\_H}}
        & 1 & \textbf{CrossFormer} & 1.97 & & 1 & \textbf{CrossFormer} & 1.97 \\
        & 2 & PatchTST & 2.92 & & 2 & PatchTST & 2.92 \\
        & 3 & iTrans. & 3.38 & & 3 & iTrans. & 3.38 \\
        & 4 & TSMixer & 4.37 & & 4 & TSMixer & 4.37 \\
        & 5 & Chronos-2 & 5.33 & & 5 & DLinear & 6.61 \\
        \midrule
        \multirow{5}{*}{\rotatebox{90}{\scriptsize H\_L\_L}}
        & 1 & \textbf{CrossFormer} & 2.29 & & 1 & \textbf{CrossFormer} & 2.29 \\
        & 2 & PatchTST & 3.22 & & 2 & PatchTST & 3.22 \\
        & 3 & iTrans. & 3.86 & & 3 & iTrans. & 3.86 \\
        & 4 & TSMixer & 4.17 & & 4 & TSMixer & 4.17 \\
        & 5 & TimesFM-2.5 & 5.06 & & 5 & DLinear & 6.46 \\
        \midrule
        \multirow{5}{*}{\rotatebox{90}{\scriptsize L\_H\_H}}
        & 1 & \textbf{CrossFormer} & 2.35 & & 1 & \textbf{CrossFormer} & 2.35 \\
        & 2 & PatchTST & 3.38 & & 2 & PatchTST & 3.38 \\
        & 3 & TimesFM-2.5 & 4.35 & & 3 & iTrans. & 4.36 \\
        & 4 & iTrans. & 4.36 & & 4 & TSMixer & 4.89 \\
        & 5 & Chronos-2 & 4.56 & & 5 & DLinear & 6.80 \\
        \midrule
        \multirow{5}{*}{\rotatebox{90}{\scriptsize L\_H\_L}}
        & 1 & \textbf{CrossFormer} & 2.39 & & 1 & \textbf{CrossFormer} & 2.39 \\
        & 2 & PatchTST & 3.19 & & 2 & PatchTST & 3.19 \\
        & 3 & iTrans. & 4.25 & & 3 & iTrans. & 4.25 \\
        & 4 & TSMixer & 4.49 & & 4 & TSMixer & 4.49 \\
        & 5 & Chronos-2 & 4.79 & & 5 & DLinear & 6.83 \\
        \midrule
        \multirow{5}{*}{\rotatebox{90}{\scriptsize L\_L\_H}}
        & 1 & \textbf{CrossFormer} & 2.85 & & 1 & \textbf{CrossFormer} & 2.85 \\
        & 2 & PatchTST & 3.16 & & 2 & PatchTST & 3.16 \\
        & 3 & Chronos-2 & 4.22 & & 3 & iTrans. & 4.52 \\
        & 4 & iTrans. & 4.52 & & 4 & TSMixer & 4.86 \\
        & 5 & TSMixer & 4.86 & & 5 & DLinear & 6.45 \\
        \midrule
        \multirow{5}{*}{\rotatebox{90}{\scriptsize L\_L\_L}}
        & 1 & \textbf{CrossFormer} & 2.15 & & 1 & \textbf{CrossFormer} & 2.15 \\
        & 2 & PatchTST & 3.09 & & 2 & PatchTST & 3.09 \\
        & 3 & iTrans. & 3.90 & & 3 & iTrans. & 3.90 \\
        & 4 & TSMixer & 4.19 & & 4 & TSMixer & 4.19 \\
        & 5 & TimesFM-2.5 & 4.97 & & 5 & DLinear & 6.54 \\
        \bottomrule
    \end{tabular}
    }
    \end{sc}
    \vspace{4pt}
    \caption{Complete model rankings by MSE Rank for all TSF regimes on \textsc{QuitoBench}. Left: all models; Right: deep learning only comparison. CrossFormer is the best deep learning model in \textbf{all 8/8 groups}.}
    \label{tab:quito_mse_complete_rankings}
\end{table}

\textit{Ranking consistency.}
The complete rankings reveal that CrossFormer dominates across all 8 TSF regimes with remarkable consistency.
PatchTST secures 2nd place in all 8 regimes among deep learning models, demonstrating strong performance as the runner-up architecture.
iTransformer consistently ranks 3rd among deep learning models, while TSMixer and DLinear occupy positions 4 and 5 respectively.
The ranking pattern (CrossFormer $\succ$ PatchTST $\succ$ iTransformer $\succ$ TSMixer $\succ$ DLinear) holds across all 8 regimes, showing remarkable stability in the MSE Rank metric.
Foundation models (Chronos-2, TimesFM-2.5, TiRex) appear in the top 5 for 5 of 8 regimes, but never claim the top spot.

\section{arXiv Benchmark Analysis}
\label{app:arxiv}

\vspace{2pt}\noindent\textbf{Data source and time period.}
We queried the arXiv API\footnote{\small \url{https://arxiv.org/help/api/}} to count papers submitted between January 1, 2020 and December 31, 2025. The arXiv preprint server provides open access to over 2 million scholarly articles and is widely used as a proxy for research activity in machine learning and related fields.

\vspace{2pt}\noindent\textbf{Domain definitions.}

We compared four research domains representing different data modalities:

\begin{itemize}[leftmargin=*]
    \item Time series: papers containing ``time series'', ``time-series'', ``timeseries'', or ``temporal forecasting'' in the title or abstract. Since there is no dedicated arXiv category for time series research, we relied on keyword matching.
    
    \item Computer vision: papers submitted to the \texttt{cs.CV} (Computer Vision and Pattern Recognition) category.
    
    \item NLP: papers submitted to the \texttt{cs.CL} (Computation and Language) category.
    
    \item Speech and audio: papers submitted to the \texttt{eess.AS} (Audio and Speech Processing) or \texttt{cs.SD} (Sound) categories.
\end{itemize}

\vspace{2pt}\noindent\textbf{Benchmark paper identification.}

To identify papers that introduce new benchmarks or datasets (rather than merely using existing ones), we applied a strict filtering criterion focusing on title keywords. A paper was classified as a ``benchmark paper'' if its title contained any of the following terms:

\begin{itemize}[leftmargin=*]
    \item \texttt{benchmark}, \texttt{-Bench}, or \texttt{Bench:}
    \item \texttt{new dataset}, \texttt{novel dataset}, \texttt{a dataset}, \texttt{dataset for}, or \texttt{datasets for}
    \item \texttt{introducing}, \texttt{we introduce}, \texttt{we present}, or \texttt{we release}
    \item \texttt{corpus for} or \texttt{new corpus}
\end{itemize}

This title-based approach provides high precision: papers introducing benchmarks typically name them prominently in the title (\eg ``ImageNet: A Large-Scale Hierarchical Image Database'' or ``GLUE: A Multi-Task Benchmark'').

\vspace{2pt}\noindent\textbf{Counting procedure.}

For each domain $d$ and year $y$, we computed:
\begin{enumerate}[leftmargin=*]
    \item $N_{d,y}^{\text{total}}$: Total number of papers matching the domain query.
    \item $N_{d,y}^{\text{bench}}$: Number of papers matching both the domain query and benchmark keywords.
\end{enumerate}

The benchmark share for domain $d$ in year $y$ is:
\[
    \text{BenchmarkShare}_{d,y} 
    = \frac{N_{d,y}^{\text{bench}}}{N_{d,y}^{\text{total}}} \times 100\%
\]

The aggregate benchmark share (2020--2025) is computed analogously using cumulative counts.

\vspace{2pt}\noindent\textbf{API query details.}

Queries were constructed using arXiv's search syntax. For example, the query for Time Series benchmark papers in 2024 was:

\begin{lstlisting}[caption={arXiv search query used to retrieve time series benchmark and dataset papers in 2024. The query matches time-series–related terms in titles or abstracts, dataset/benchmark-introducing phrases in titles.},label={lst:arxiv_queries}]
(ti:"time series" OR ti:"time-series" OR ti:timeseries 
 OR abs:"time series" OR abs:"time-series" OR abs:timeseries 
 OR ti:"temporal forecasting" OR abs:"temporal forecasting")
AND (ti:benchmark OR ti:"new dataset" OR ti:"novel dataset" 
     OR ti:"a dataset" OR ti:"dataset for" OR ti:"datasets for" 
     OR ti:"introducing" OR ti:"we introduce" OR ti:"we present" 
     OR ti:"we release" OR ti:"corpus for" OR ti:"new corpus" 
     OR ti:"-bench" OR ti:"Bench:" OR ti:"-dataset")
AND submittedDate:[202401010000 TO 202412312359]
\end{lstlisting}

The total result count was extracted from the \texttt{opensearch:totalResults} field in the API response.

\vspace{2pt}\noindent\textbf{Limitations.}

\begin{itemize}[leftmargin=*]
    \item Keyword sensitivity: Our keyword-based approach may miss benchmark papers with unconventional titles or include false positives. However, the consistent methodology across domains ensures fair comparison.
    
    \item arXiv coverage: Not all research is posted to arXiv. Coverage varies by field: NLP and ML have high arXiv adoption, while some application domains may be underrepresented.
    
    \item Time series category: Unlike Computer Vision (\texttt{cs.CV}) or NLP (\texttt{cs.CL}), time series research lacks a dedicated arXiv category, requiring keyword-based identification which may have different recall characteristics.
    
    \item Preprint vs.\ publication: arXiv submissions are preprints and may not reflect final publication venues or peer review outcomes.
\end{itemize}

\vspace{2pt}\noindent\textbf{Reproducibility.}

The analysis code is available at \url{https://github.com/alipay/quito}, which is partially derived from \url{https://github.com/SerendipityOneInc/look-bench}~\citep{gao2026lookbench}. The queries can be reproduced using the Python \texttt{arxiv} package or direct API calls to \url{https://export.arxiv.org/api/query}.

\end{document}